\documentclass{article} %
\usepackage[T1]{fontenc}

\newif\iffinalcopy
\finalcopytrue

\newif\ifisarxiv
\isarxivtrue

\newcommand{\textSwitch}[2]{%
  \ifisarxiv
    #1%
  \else
    #2%
  \fi
}

\usepackage{url}
\usepackage{catchfile}

\usepackage{amsmath}
\usepackage{amssymb}

\usepackage{subcaption}
\usepackage{booktabs}
\usepackage{graphicx}
\usepackage{svg}
\usepackage{pgf}

\usepackage{layout}
\usepackage{enumitem}

\usepackage{silence}
\usepackage{dblfloatfix}  %

\usepackage[accepted]{icml2023}

\usepackage{hyperref}
\hypersetup{
    colorlinks=true,
    linkcolor=blue,
    urlcolor=cyan,
}

\usepackage{printlen}
\uselengthunit{in}

\definecolor{darkred}{HTML}{990000}
\definecolor{darkgreen}{HTML}{38761d}
\usepackage{pifont}%
\newcommand{\cmark}{\ding{51}}%
\newcommand{\xmark}{\ding{55}}%
\newcommand{\yessymbol}{{\color{darkgreen} \cmark}}
\newcommand{\nosymbol}{{\color{darkred} \xmark}}

\newcommand{\ourmctslong}{Adversarial MCTS}
\newcommand{\ourmctsabbrev}{A-MCTS}

\newcommand{\ourmctssamplelong}{Adversarial MCTS: Sample}

\newcommand{\ourmctssampleabbrev}{A-MCTS-S}

\newcommand{\ourmctssamplesymmetrylong}{Adversarial MCTS: More Accurate Sampling}
\newcommand{\ourmctssamplesymmetryabbrev}{A-MCTS-S++}

\newcommand{\ourmctsperfectlong}{Adversarial MCTS: Recursive}

\newcommand{\ourmctsperfectabbrev}{A-MCTS-R}

\newcommand{\ourmctsmodellong}{Adversarial MCTS: Victim Model}
\newcommand{\ourmctsmodelabbrev}{A-MCTS-VM}

\def\signed #1{{\leavevmode\unskip\nobreak\hfil\penalty50\hskip2em
  \hbox{}\nobreak\hfil(#1)%
  \parfillskip=0pt \finalhyphendemerits=0 \endgraf}}

\newsavebox\mybox
\newenvironment{aquote}[1]
  {\savebox\mybox{#1}\begin{quote}}
  {\signed{\usebox\mybox}\end{quote}}

\iffinalcopy
\newcommand{\githuburl}{\href{https://github.com/AlignmentResearch/go_attack/}{GitHub}}
\newcommand{\demosite}{https://goattack.far.ai}
\newcommand{\demositecleanhref}{\href{\demosite}{\texttt{goattack.far.ai}}}
\newcommand{\gorcolablink}{\href{https://colab.research.google.com/drive/10qCu7jMaR5ry2IkEe-kfjhouZL8\_srbV\#sandboxMode=true}{Colab notebook link}}
\else
\newcommand{\githuburl}{---anonymized---}
\newcommand{\demosite}{https://go-attack-icml.netlify.app}
\newcommand{\demositecleanhref}{\href{\demosite}{\texttt{go-attack-icml.netlify.app}}}
\newcommand{\gorcolablink}{---anonymized---}
\fi

\numberwithin{figure}{section}
\numberwithin{table}{section}

\newcommand{\cpthirtynine}{\texttt{cp39}}
\newcommand{\cpsixtythree}{\texttt{cp63}}
\newcommand{\cponetwentyseven}{\texttt{cp127}}
\newcommand{\cpfivezerofive}{\texttt{Latest}}
\newcommand{\cponezerothree}{\texttt{Original}}
\newcommand{\hardened}[1]{#1$_\texttt{def}$}

\begin{document}

\twocolumn[
    \icmltitle{Adversarial Policies Beat Superhuman Go AIs}

    \icmlsetsymbol{equal}{*}
    
    \begin{icmlauthorlist}
        \icmlauthor{Tony T Wang}{equal,mit}
        \icmlauthor{Adam Gleave}{equal,berk,far}
        \icmlauthor{Tom Tseng}{far}
        \icmlauthor{Kellin Pelrine}{far,mila}
        \icmlauthor{Nora Belrose}{far}
        \icmlauthor{Joseph Miller}{far}
        \icmlauthor{Michael D Dennis}{berk}
        \icmlauthor{Yawen Duan}{berk}
        \icmlauthor{Viktor Pogrebniak}{}
        \icmlauthor{Sergey Levine}{berk}
        \icmlauthor{Stuart Russell}{berk}
    \end{icmlauthorlist}
    
    \icmlaffiliation{mit}{MIT}
    \icmlaffiliation{berk}{UC Berkeley}
    \icmlaffiliation{far}{FAR AI}
    \icmlaffiliation{mila}{McGill University; Mila}
    
    \icmlcorrespondingauthor{Tony T Wang}{twang6@mit.edu}
    \icmlcorrespondingauthor{Adam Gleave}{adam@far.ai}

    \icmlkeywords{Go, Attack}
    
    \vskip 0.3in
]

\printAffiliationsAndNotice{\icmlEqualContribution} %

\begin{abstract}

    We attack the state-of-the-art Go-playing AI system KataGo by training adversarial policies against it,
    achieving a >97\% win rate against KataGo running at superhuman settings.
    Our adversaries do not win by playing Go well. Instead, they trick KataGo into making serious blunders.
    Our attack transfers zero-shot to other superhuman Go-playing AIs, and is comprehensible to the extent that human experts can implement it without algorithmic assistance to consistently beat superhuman AIs.
    The core vulnerability uncovered by our attack persists even in KataGo agents adversarially trained to defend against our attack.
    Our results demonstrate that even superhuman AI systems may harbor surprising failure modes.
    Example games are available at \demositecleanhref.
\end{abstract}

\section{Introduction}

The average-case performance of AI systems has grown rapidly in recent years, from RL agents achieving superhuman performance in competitive games~\citep{silver2016,silver2018,openai2019dota} to generative models showing signs of general intelligence~\citep{openai2023gpt4,bubeck2023sparks}. However, designing AI systems with good \emph{worst-case} performance remains an open problem. One key question is whether average-case performance gains can be translated into worst-case robustness. If so, then efforts to increase average-case performance such as through scaling models would naturally lead to robustness. We find that even superhuman systems can fail catastrophically, suggesting that capabilities are not enough: a dedicated effort will be needed to make systems robust.

\begin{figure*}[t]
    \centering
    \begin{subfigure}{0.48\textwidth}
        \centering
        \includesvg[width=0.9\textwidth]{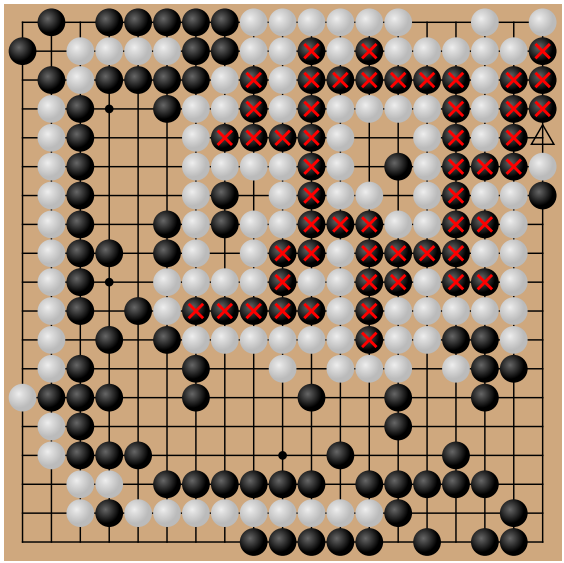}
	\caption{Our \textit{cyclic-adversary} wins as white by capturing a cyclic group (\textcolor{red}{$\mathbf{\times}$}) that the victim (\cpfivezerofive{}, $10$ million visits) leaves vulnerable. \href{\demosite/adversarial-policy-katago?row=3\#10mil_visits-board}{Explore the game}.}
    \label{fig:cp505vis1-board:hardened}
    \end{subfigure}
    \quad
     \begin{subfigure}{0.48\textwidth}
        \centering
        \includesvg[width=0.9\textwidth]{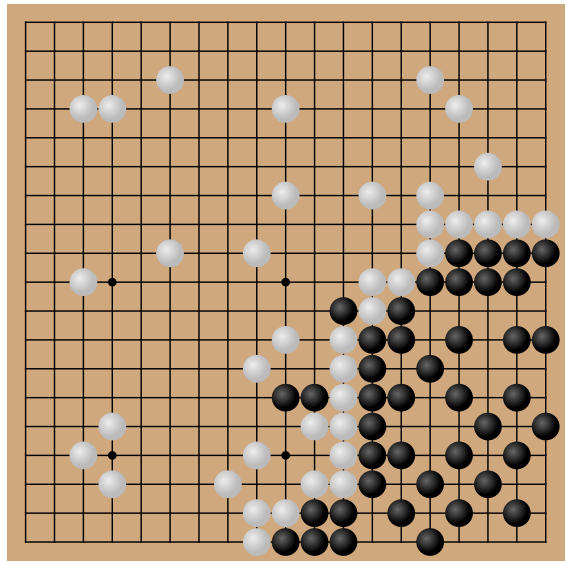}
	\caption{Our \textit{pass-adversary} wins as black by tricking the victim (\cpfivezerofive{}, no search) into passing prematurely, ending the game. \href{\demosite/pass-based-attack?row=1\#no_search-board}{Explore the game}.}
    \label{fig:cp505vis1-board:unhardened}
    \end{subfigure}
    \vspace{-1mm}
	\caption{
    Games between the strongest KataGo network at the time of conducting this research (which we refer to as \cpfivezerofive{}) and two different types of adversaries we trained.
	(a)
    Our \textit{cyclic-adversary} beats KataGo even when KataGo plays with far more search than is needed to be superhuman.
	The adversary lures the victim into letting a large group of cyclic victim stones (\textcolor{red}{$\mathbf{\times}$}) get captured by the adversary's next move ($\Delta$).
	Appendix~\ref{app:kellin-donut-analysis} has a detailed description of this adversary's behavior.
	(b)
	Our \textit{pass-adversary} beats no-search KataGo by tricking it into passing. The adversary then passes in turn, ending the game with the adversary winning under the Tromp-Taylor ruleset for computer Go~\citep{tromp:2014} that KataGo was trained and configured to use (see Appendix~\ref{app:rules}).
	The adversary gets points for its territory in the bottom-right corner (devoid of victim stones) whereas the victim does not get points for the territory in the top-left due to the presence of the adversary's stones.
    }
    \vspace{-4mm}
    \label{fig:cp505vis1-board}
\end{figure*}

In particular, we find vulnerabilities in KataGo~\citep{wu2019}, the strongest publicly available Go-playing AI system. We find these vulnerabilities by training adversarial policies to beat KataGo. Using less than $14\%$ of the compute used to train KataGo, we obtain adversarial policies that win >$99\%$ of the time against KataGo with no search, and >$97\%$ of the time against KataGo with enough search to be superhuman. Critically, our adversaries do not win by playing Go well.\footnote{Despite being able to beat KataGo,
our adversarial policies lose against even amateur Go players (Appendix~\ref{app:experiments:human-vs-adversary}).} Instead, they trick KataGo into making serious blunders that cause it to lose the game (Figure~\ref{fig:cp505vis1-board}).

Our adversaries transfer zero-shot to other superhuman Go-playing AIs, and the strategy they use can be replicated by human experts to consistently beat many different superhuman AIs (Section~\ref{sec:repro}). Moreover, the core vulnerability uncovered by our attack persists even in KataGo agents adversarially trained to defend against our attack, suggesting that the vulnerability is non-trivial to patch.

We chose to attack KataGo as we expected it to be unusually challenging to exploit, such that a successful attack suggests that a broad swathe of other systems will be vulnerable. In particular, KataGo's capabilities are superhuman by a large margin, whereas the state-of-the-art in broader domains like language modeling are still subhuman at many tasks. Moreover, Go is naturally an adversarial setting, such that average-case performance should be predictive of worst-case performance. 

Most prior work on robustness has focused on ML systems in isolation. However, techniques such as simulation of alternatives at inference time~\citep{yao2023tree} and self-reflection~\citep{bai2022constitutional} can improve system robustness. KataGo performs substantial simulation and self-reflection in the form of Monte-Carlo Tree Search~\citep{coulom2007efficient}, but our attack still wins more often than not even when KataGo searches 10 million nodes per move.

Our adversaries have no special powers: they can only place stones or pass, like a regular player.
We do however give our adversaries gray-box access to the victim network they are attacking (Section~\ref{sec:threat-model}).
In particular, we train our adversaries using an AlphaZero-style training process~\citep{silver2018}, similar to that of KataGo.
The key differences are that we collect games with the adversary playing against the victim, and that we use the victim network to select victim moves during the adversary's tree search.

Our paper makes three contributions.
First, we propose a novel attack method, hybridizing the attack of \citet{gleave2020} with AlphaZero-style training~\citep{silver2018}.
Second, we demonstrate the existence of two distinct adversarial policies against the state-of-the-art Go AI system, KataGo.
Finally, we provide a detailed empirical investigation into these adversarial policies.
Our open-source implementation is available at \githuburl{}.

\section{\textSwitch{Related work}{Related Work}}
\label{sec:related_work}

Our work is inspired by the presence of adversarial examples in a wide variety of models~\citep{szegedy2014}.
Notably, though not consistently superhuman~\citep{shankar2020}, many image classifiers reach and sometimes surpass human performance in a number of contexts~\citep{hophuoc2018,russakovsky2015,shankar2020,pham2021}.
Yet even these state-of-the-art image classifiers are vulnerable to adversarial examples~\citep{carlini2019,ren2020}.
This raises the question: could highly capable deep RL policies be similarly vulnerable? 

One might hope that the adversarial nature of self-play training would naturally lead to robustness.
This strategy works for image classifiers, where adversarial training is a somewhat effective if computationally expensive defense~\citep{madry2018,ren2020}.
This view is bolstered by idealized versions of self-play provably converging to a Nash equilibrium, which is unexploitable~\citep{brown1951,heinrich2015}.
However, our work finds that, in fact, even state-of-the-art and superhuman-level deep RL policies are still highly vulnerable to exploitation.

It is known that self-play may not converge in non-transitive games~\citep{balduzzi2019} like rock-paper-scissors, where A beats B and B beats C yet C beats A.
However, \citet{wojciech2020} argues real-world games like Go grow increasingly transitive as skill increases.
This would imply that while self-play may struggle with non-transitivity early in training, comparisons involving highly capable policies such as KataGo should be mostly transitive.
But we find significant non-transitivity: our adversaries exploit KataGo agents that beat human professionals, yet lose to most amateur Go players (Appendix~\ref{app:experiments:human-vs-adversary}).

\begin{figure*}[t]
    \centering
    \includegraphics[width=0.75\textwidth]{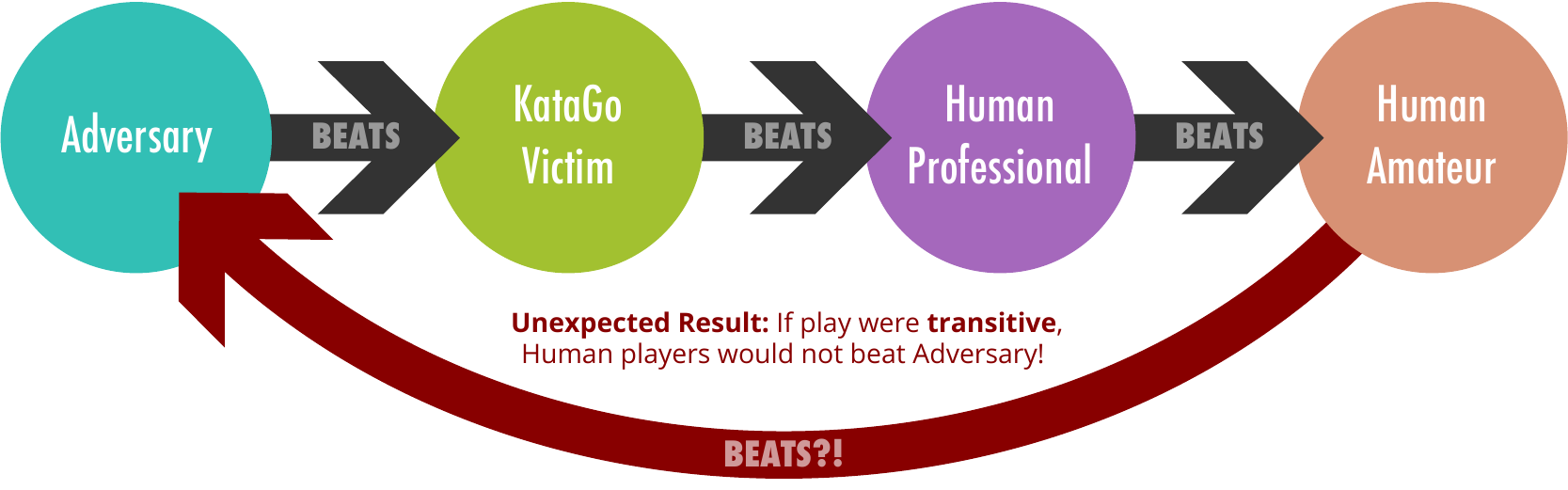}
    \caption[]{A human amateur beats our adversarial policy (Appendix~\ref{app:experiments:human-vs-adversary}) that beats KataGo. This non-transitivity shows the adversary is not a generally capable policy, and is just exploiting KataGo.}
    \label{fig:non-transitivity}
    \vspace{-4mm}
\end{figure*}

Most prior work attacking deep RL has focused on perturbing observations~\citep{huang2017,ilahi2022}.
Concurrent work by~\citet{lan2022} shows that KataGo with ${\leq}50$ visits can be induced to play poorly by adding two adversarially chosen moves to a board, even though these moves do not substantially change the win rate estimated by KataGo with 800 visits.
However, the perturbed input is unrealistic, as the move history seen by the KataGo network implies that it \emph{chose} to play a seemingly poor move on the previous turn.
Moreover, an attacker that can force the opponent to play a specific move has easier ways to win: it could simply make the opponent resign, or play a maximally bad move.
We instead follow the threat model introduced by \citet{gleave2020} of an adversarial \emph{agent} acting in a shared environment.

Prior work on such \emph{adversarial policies} has focused on attacking subhuman policies in simulated robotics environments~\citep{gleave2020,wu2021}.
In these environments, the adversary can often win just by causing the victim to make small changes to its actions.
By contrast, our work focuses on exploiting superhuman-level Go policies that have a discrete action space.
Despite the more challenging setting, we find these policies are not only vulnerable to attack, but also fail in surprising ways that are quite different from human-like mistakes.

Adversarial policies give a lower bound on the \emph{exploitability} of an agent: 
how much expected utility a best-response policy achieves above the minimax value of the game.
Exactly computing a policy's exploitability is feasible in some low-dimensional games~\citep{johanson2011}, but not in larger games such as Go with approximately $10^{172}$ possible states~\citep[Section~6.3.12]{allis1994}.
Prior work has lower bounded the exploitability in some poker variants using search~\citep{lisy2017}, but the method relies on domain-specific heuristics that are not applicable to Go.

In concurrent work \citet{timbers2022} developed the \emph{approximate best response} (ABR) method to estimate exploitability.
Whereas we exploit an open-source system KataGo, they exploit a proprietary replica of AlphaZero from~\citet{schmid2021}.
Both \citeauthor{timbers2022} and our attacks use AlphaZero-style training modified to use the \emph{opponent's} policy during search, with a curriculum over the victim's search budget.
However, our curriculum also varies the victim checkpoint.
Furthermore, we trained our \emph{cyclic-adversary} by first patching KataGo to protect against our initial \emph{pass-adversary}, then repeating the attack.

Our main contribution lies in our experimental results. \citeauthor{timbers2022} obtain a 90\% win rate against no-search AlphaZero and 65\% with 800 visits~\citep[Figure~3]{timbers2022}.
In Appendix~\ref{app:experiments:strength:alphazero} we estimate that their AlphaZero victim with 800 visits plays at least at the level of a top-200 professional and may be superhuman. But we show our attack beats victims playing with an unquestionably superhuman $10^7$ visits.
Furthermore, our experiments give an in-depth investigation of this vulnerability, and include insights on defense, transfer, both human and mechanistic interpretability, the role of search for both victim and adversary, the evolution of the attack over training, and more.

\section{Background}

\subsection{\textSwitch{Threat model}{Threat Model}}
\label{sec:threat-model}

Following \citet{gleave2020}, we consider the setting of a two-player zero-sum Markov game~\citep{shapley1953}.
Our threat model assumes the attacker plays as one of the agents, which we will call the \emph{adversary}, and seeks to win via standard play against some \emph{victim} agent.

The key capability we grant to the attacker is gray-box access to the victim agent.
That is, the attacker can evaluate the victim's neural network on arbitrary inputs.
However, the attacker does not have direct access to the network weights.
We furthermore assume the victim agent follows a fixed policy, corresponding to the common case of a pre-trained model deployed with static weights.
Gray-box access to a fixed victim naturally arises whenever the attacker can run a copy of the victim agent, e.g., when attacking a commercially available or open-source Go AI system.
However, we also weaken this assumption in some of our experiments, seeking to \emph{transfer} the attack to an unseen victim agent---an extreme case of a black-box attack.

We know the victim must have weak spots: optimal play is intractable in a game as complex as Go.
However, these vulnerabilities could be quite hard to find, especially using only gray-box access.
Exploits that are easy to discover will tend to have already been found by self-play training, resulting in the victim being immunized against them.

Consequently, our two primary success metrics are the \emph{win rate} of the adversarial policy against the victim and the adversary's \emph{training and inference time}.
We also track the mean score difference between the adversary and victim, but this is not explicitly optimized for by the attack.
Tracking training and inference time rules out the degenerate ``attack'' of simply training KataGo for longer than the victim, or letting it search deeper at inference.

In principle, it is possible that a more sample-efficient training regime could produce a stronger agent than KataGo in a fraction of the training time.
While this might be an important result, we would hesitate to classify it as an attack.
Rather, we are looking for the adversarial policy to demonstrate \emph{non-transitivity}, as this suggests the adversary is winning by exploiting a specific weakness in the opponent.
That is, as depicted in Figure~\ref{fig:non-transitivity}, the adversary beats the victim, the victim beats some baseline opponent, and that baseline opponent can in turn beat the adversary.

\subsection{KataGo}

We chose to attack KataGo as it is the strongest publicly available Go AI system at the time of writing.
KataGo won against ELF OpenGo~\citep{tian2019} and Leela Zero~\citep{pascutto2019} after training for only 513 V100 GPU days~\citep[section~5.1]{wu2019}.
ELF OpenGo is itself superhuman, having won all 20 games played against four top-30 professional players.
The latest networks of KataGo are even stronger than the original, having been trained for over 15,000 V100-equivalent GPU days (Appendix~\ref{app:compute-estimates:katago}). %
Indeed, even the policy network with \emph{no search} is competitive with top professionals (see Appendix~\ref{app:experiments:strength:katago-no-search}).
KataGo learns via self-play, using an AlphaZero-style training procedure~\citep{silver2018}.
The agent contains a neural network with a \emph{policy head}, outputting a probability distribution over the next move, and a \emph{value head}, estimating the win rate from the current state.
It then conducts Monte-Carlo Tree Search (MCTS) using these heads to select self-play moves, described in Appendix~\ref{app:mcts-review}.
KataGo trains its policy head to mimic the outcome of this tree search, and its value head to predict whether the agent wins the self-play game.
Each step of training is designed to act as a policy-improvement operator.

In contrast to AlphaZero, KataGo has several additional heads that predict auxiliary targets such as the opponent's next move and which player ``owns'' a square on the board.
The outputs of these heads are not used for actual game play, serving only to speed up training via the addition of auxiliary losses.
KataGo also introduces architectural improvements such as global pooling, training process improvements such as playout cap randomization, and hand-engineered input features such as a ladderable stones mask.

These modifications to KataGo improve its sample and compute efficiency by several orders of magnitude over prior work such as ELF OpenGo, and protect KataGo from some previously known vulnerabilities in neural-net-based Go AIs (Appendix~\ref{app:bot_weaknesses}).
For these reasons, we choose to build our attack on top of KataGo, adopting its various architecture and training improvements and hand-engineered features. In principle though, the same attack could be implemented on top of any AlphaZero-style training pipeline.

\section{\textSwitch{Attack methodology}{Attack Methodology}}
\label{sec:methodology}

Prior works, such as KataGo and AlphaZero, train on self-play games where an agent plays many games against itself.
We instead train on games between our adversary and a fixed victim agent,
and only train the adversary on data from the turns where it is the adversary's move, since we wish to train the adversary to exploit the victim, not mimic it.
We dub this procedure \emph{victim-play}.

\begin{figure}[t]
    \centering
    \includegraphics[width=\columnwidth]{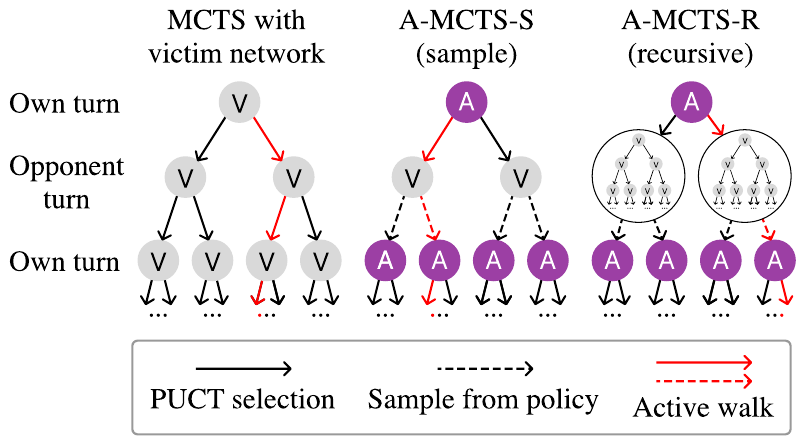}
    \vspace{-7mm}
    \caption{%
        MCTS (left) builds a search tree one node at a time.
        To add a node, it walks down the tree until a new leaf is reached (red arrows).
        At a node $x$, the next step of the walk is determined by a PUCT~\citep{rosinMultiarmedBanditsEpisode2011} algorithm (solid arrows) which takes into account neural network evaluations of each node in the subtree of $x$.
        \ourmctssampleabbrev{} (middle) walks down the tree by using a modified PUCT algorithm at adversary nodes, and sampling directly from the victim's policy network (dashed arrows) at victim nodes. \ourmctsperfectabbrev{} (right) performs a full simulation of the victim as opposed to sampling from the victim's policy net.
	    Search trees are depicted as binary for illustrative purposes only.
	    See Appendix~\ref{app:search-algorithms} for full details.
    }
    \label{fig:amcts-diagram}
    \vspace{-5mm}
\end{figure}

\textbf{Adversarial MCTS}. In regular self-play, an agent models its opponent's moves by sampling from its own policy network.
This makes sense, as in this case the policy \textit{is} playing itself.
But in victim-play, it would be a mistake to model the victim's behavior using the adversary's policy network.
We introduce \emph{\ourmctslong{}} (\ourmctsabbrev{}) to address this problem (Figure~\ref{fig:amcts-diagram}).

We experiment with three variants of \ourmctsabbrev{}.
The \emph{sample} variant (\ourmctssampleabbrev{}) models a computationally bounded version of the victim that plays moves directly from its policy head. \ourmctssamplesymmetryabbrev{} improves upon this by averaging the victim policy head over board symmetries to match the default behavior of KataGo. Finally, the \emph{recursive} variant (\ourmctsperfectabbrev{}) models the victim perfectly at the cost of increased computational complexity; the cost of adversary training and inference is increased by a factor equal to the victim's search budget. We use \ourmctsperfectabbrev{} to study the benefits of using a more accurate model of the victim.

\textbf{Initialization}. We randomly initialize the adversary's network.
We cannot initialize the adversary's weights to those of the victim as our threat model does not allow white-box access. A random initialization also encourages exploration to find weaknesses in the victim, rather than simply producing a stronger Go player.
However, a randomly initialized network will almost always lose against a highly capable network, leading to a challenging initial learning problem.
Fortunately, the adversary's network is able to learn something useful about the game even from games that are lost, due to KataGo's auxiliary loss functions.

\textbf{Curriculum}.
We use a curriculum that trains against successively stronger versions of the victim in order to help overcome the challenging random initialization.
We switch to a more challenging victim agent once the adversary's win rate exceeds a certain threshold.
We modulate victim strength in two ways.
First, we train against successively later checkpoints of the victim agent, as KataGo releases its entire training history.
Second, we increase the amount of search that the victim performs during victim-play.

\section{Evaluation} \label{sec:evaluation}

We evaluate our attack against KataGo~\citep{wu2019}, focusing on the \href{https://katagotraining.org/networks/}{\texttt{b40c256-s11840935168}} network, which was the strongest KataGo network at the time of our main experiments, and which we refer to as \cpfivezerofive{}.
In Section~\ref{sec:evaluation:no-search} we use \ourmctssampleabbrev{} with 600 adversary visits to train our \textit{pass-adversary}, achieving a 99.9\% win rate against \cpfivezerofive{} playing without search.
Even without search \cpfivezerofive{} is comparable to a top-100 European player (Appendix~\ref{app:experiments:strength:katago-no-search}). The pass-adversary beats \cpfivezerofive{} by tricking it into passing early and losing (Figure~\ref{fig:cp505vis1-board:unhardened}).

In Section~\ref{sec:evaluation:no-search-hardened}, we add a \emph{pass-alive defense} to the victim to defend against the pass-adversary. The defended victim \hardened{\cpfivezerofive{}} cannot lose via accidentally passing, and is about as strong as \cpfivezerofive{} (it beats \cpfivezerofive{} 456/1000 games when both agents use no tree search, and 461/1000 games when both use 2048 visits/move of search).

\begin{figure*}[ht]
    \centering
    \input{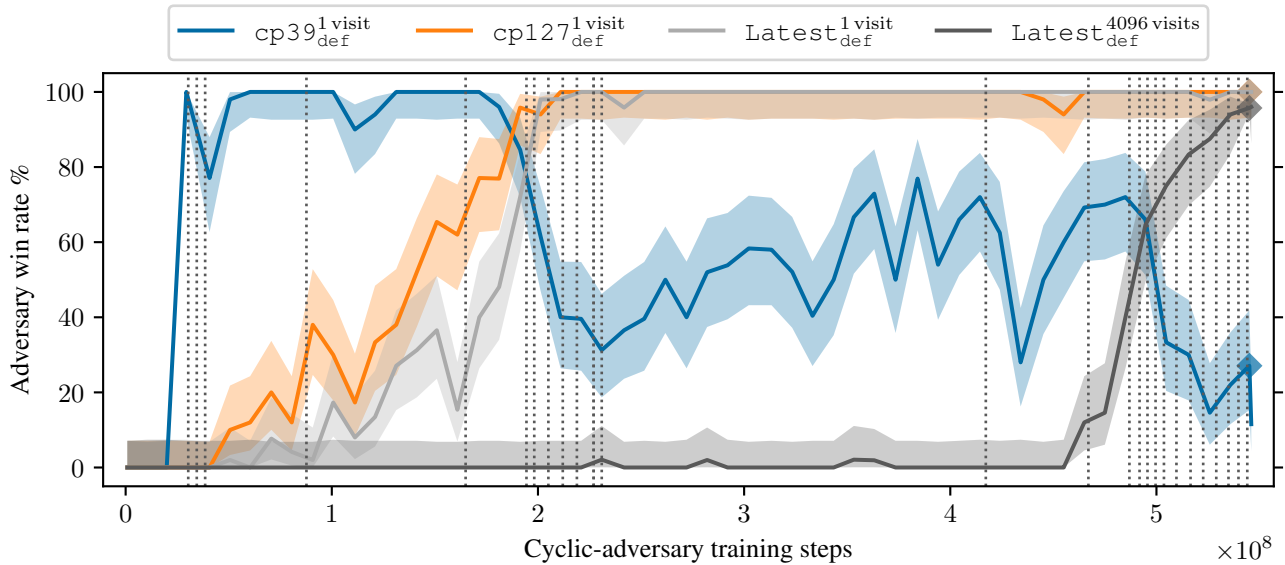}
    \vspace{-6mm}
    \caption{
    The win rate ($y$-axis) of the cyclic-adversary over time ($x$-axis) playing with 600 visits against four different victims. 
    The strongest cyclic-adversary checkpoint (marked $\blacklozenge$) wins 1048/1048 = 100\% games against \hardened{\cpfivezerofive{}} without search and 1007/1052 = 95.7\% games against \hardened{\cpfivezerofive{}} with 4096 visits.
    The shaded interval is a 95\% Clopper-Pearson interval over 50 games per checkpoint. 
    The cyclic-adversary is trained with a curriculum, starting from \hardened{\cpthirtynine{}} without search and ending at \hardened{\cpfivezerofive{}} with 131,072 visits. 
    Vertical dotted lines denote switches to stronger victim networks or to an increase in \hardened{\cpfivezerofive{}}'s search budget. 
    }
    \label{fig:evaluation:training-curve-cyclic}
    \vspace{-4mm}
\end{figure*}

Repeating the \ourmctssampleabbrev{} attack against \hardened{\cpfivezerofive{}} yields our \textit{cyclic-adversary}\footnotemark, which is qualitatively very different from the pass-adversary as it does not use the pass-trick (Figure~\ref{fig:cp505vis1-board:hardened}) that achieves a 100\% win rate over 1048 games against \hardened{\cpfivezerofive{}} playing without search.
\footnotetext{
    Unless otherwise specified, ``cyclic-adversary'' refers to the strongest checkpoint
    indicated in Figure~\ref{fig:evaluation:training-curve-cyclic}.
    Likewise ``pass-adversary'' refers to the strongest checkpoint in
    Figure~\ref{fig:evaluation:training-curve-no-search}.
}
The cyclic-adversary succeeds against victims playing with search as well (detailed in Section~\ref{sec:evaluation:with-search}), achieving a 95.7\% win rate against \hardened{\cpfivezerofive{}} with 4096 visits and a 72\% win rate against \cpfivezerofive{} with $10^7$ visits.\footnote{We verified it is not winning any games with the pass-trick.} In Appendix~\ref{app:experiments:strength:katago-search}, we estimate that \cpfivezerofive{} with 4096 visits is already much stronger than the best human Go players, and \cpfivezerofive{} with $10^7$ visits far surpasses them.

In the remaining results sections, we provide a deeper understanding of the cyclic adversary and vulnerability, looking at defense (Section~\ref{sec:evaluation:defense}), how the attack works and the victim fails (Section~\ref{sec:evaluation:understanding-the-attack}), and transfer (Section~\ref{sec:evaluation:transfer}).

\subsection{\textSwitch{Attacking a victim without search}{Attacking a Victim Without Search}}
\label{sec:evaluation:no-search}

Our pass-adversary playing with 600 visits achieves a 99.9\% win rate against \cpfivezerofive{} with no search.
Notably, our pass-adversary wins despite being trained for just 20.4 V100 GPU days, which is 0.13\% of \cpfivezerofive{}'s training budget (Appendix~\ref{app:compute-estimates}).
Importantly, the pass-adversary does not win by playing a stronger game of Go than \cpfivezerofive{}.
Instead, it follows a bizarre strategy illustrated in Figure~\ref{fig:cp505vis1-board:unhardened} that loses even against human amateurs (see Appendix~\ref{app:experiments:human-vs-adversary}).
The strategy tricks the KataGo policy head into passing prematurely at a move where the adversary has more points under Tromp-Taylor Rules (Appendix~\ref{app:rules}).

We trained our pass-adversary using \ourmctssampleabbrev{} and a curriculum, as described in Section~\ref{sec:methodology}.
Our curriculum starts from a checkpoint \cponetwentyseven{} around a quarter of the way through KataGo's training, and ends with the \cpfivezerofive{} checkpoint corresponding to the strongest KataGo network (see Appendix~\ref{app:hyperparameters:no-search} for details).
Appendix~\ref{app:unhardened-results} contains further evaluation and analysis of our pass-adversary.
Although this adversary was only trained on no-search victims, it transfers to very low search victims.
Using \ourmctsperfectabbrev{} the adversary achieves an 88\% win rate against \cpfivezerofive{} playing with 8 visits.
This win rate drops to 15\% when the adversary uses \ourmctssampleabbrev{}.

\subsection{\textSwitch{Attacking a defended victim}{Attacking a Defended Victim}}
\label{sec:evaluation:no-search-hardened}

We design a hard-coded defense for the victim against the attack found in Section~\ref{sec:evaluation:no-search}: prohibiting passing until it cannot change the game outcome. Concretely, we only allow the victim to pass when all its legal moves are in its own \emph{pass-alive territory}, a concept described in the official KataGo rules~\citep{wu2021katago} that extends the traditional Go notion of a pass-alive group (see Appendix~\ref{app:pass-alive} for full defense details). Given a victim \texttt{V}, we denote the victim with this defense applied \hardened{\texttt{V}} .
The defense completely thwarts the pass-adversary from 
Section~\ref{sec:evaluation:no-search}; the pass-adversary loses every game out of 1000 against \hardened{\cpfivezerofive{}}.

We repeat our \ourmctssampleabbrev{} attack against the defended victim, obtaining our cyclic-adversary.
The curriculum  (Appendix~\ref{app:hyperparameters:hardened-curriculum}) starts from an early checkpoint \hardened{\cpthirtynine{}} with no search and continues until \hardened{\cpfivezerofive{}}.
The curriculum then starts increasing the number of victim visits.

In Figure~\ref{fig:evaluation:training-curve-cyclic} we evaluate various cyclic-adversary checkpoints against the policy networks of \hardened{\cpthirtynine{}}, \hardened{\cponetwentyseven{}}, and \hardened{\cpfivezerofive{}}. We see that an attack that works against \hardened{\cpfivezerofive{}} transfers well to \hardened{\cponetwentyseven{}} but not to \hardened{\cpthirtynine{}}, and an attack against \hardened{\cpthirtynine{}} early in training did not transfer well to \hardened{\cponetwentyseven{}} or \hardened{\cpfivezerofive{}}. These results suggest that different checkpoints have unique vulnerabilities. We analyze the evolution of our cyclic-adversary's strategy in Appendix~\ref{app:training-games-analysis}.

Our best cyclic-adversary checkpoint playing with 600 visits against \hardened{\cpfivezerofive{}} playing with no search achieves a $100.0\%$ win rate over 1048 games. It also still works against \cpfivezerofive{} with the defense disabled, achieving a 100.0\% win rate over 1000 games. The cyclic-adversary is trained for 2223.2 V100 GPU days, which is roughly 14.0\% of the compute used for training \cpfivezerofive{} (Appendix~\ref{app:compute-estimates}). The cyclic-adversary still loses against human amateurs (Appendix~\ref{app:experiments:human-vs-adversary}).

\subsection{\textSwitch{Attacking a victim with search}{Attacking a Victim With Search}}
\label{sec:evaluation:with-search}

\begin{figure*}[b]
    \vspace{-4mm}
    \centering
    \begin{subfigure}{0.48\textwidth}%
        \centering
        \input{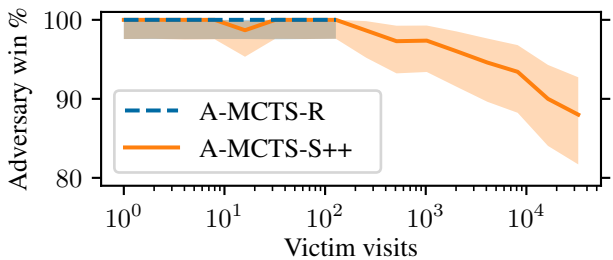}
        \caption{
        Win rate of cyclic-adversary (y-axis) playing with 600 visits/move vs. \hardened{\cpfivezerofive{}} with varying amounts of search ($x$-axis).
        Victims with more search are harder to exploit.}
        \label{fig:evaluation:search:victim-visits-hardened}
    \end{subfigure}
    \quad
    \begin{subfigure}{0.48\textwidth}%
        \centering
        \input{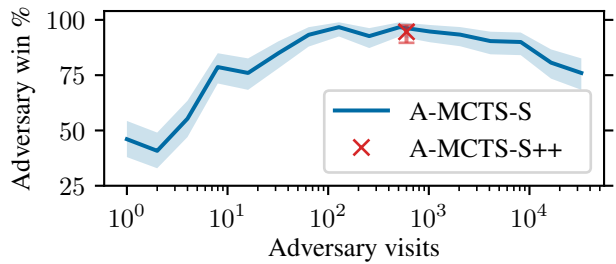}
        \caption{
        Win rate of cyclic-adversary (y-axis) playing with varying visits ($x$-axis). The victim \hardened{\cpfivezerofive{}} plays with a fixed 4096 visits/move. Win rates are best with 128--600 adversary visits.}
        \label{fig:evaluation:search:adv-visits-hardened}
    \end{subfigure}
    \vspace{-2mm}
    \caption{%
    We evaluate the cyclic-adversary's win rate against \hardened{\cpfivezerofive{}} with varying amounts of search (\textbf{left}: victim, \textbf{right}: adversary). Shaded regions and error bars denote 95\% Clopper-Pearson confidence intervals over \char`\~150 games.
    }
    \label{fig:evaluation:search-hardened}
    \vspace{2mm}
\end{figure*}

\begin{figure*}[b]
    \centering
    \begin{subfigure}{0.48\textwidth}%
        \centering
        \input{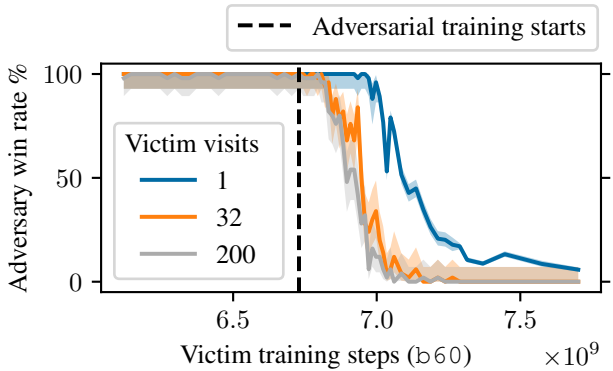}
        \caption{%
            Win rate of our cyclic-adversary$^\text{600 visits}$
            vs. 60-block KataGo nets
            from KataGo's ongoing distributed training run.
        }
        \label{fig:defense:b60-adv-train-curve}
    \end{subfigure}
    \quad
    \begin{subfigure}{0.48\textwidth}%
        \centering
        \input{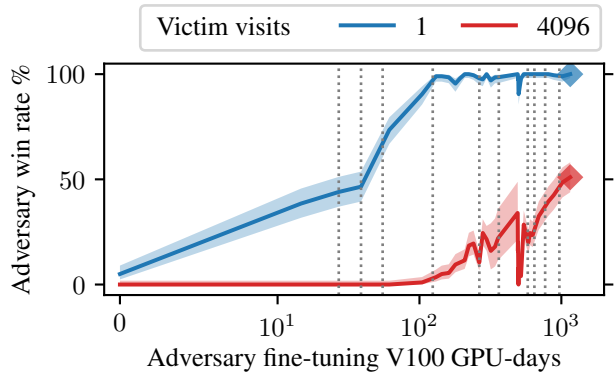}
        \caption{%
            Win rate of our cyclic-adversary$^\text{600 visits}$
            as it is fine-tuned against
            a recent adversarially trained KataGo net, \texttt{b60-s7702m}.
        }
        \label{fig:defense:adv-finetuning}
    \end{subfigure}
    \vspace{-2mm}
    \caption{%
        Adversarial training gradually makes KataGo immune to our cyclic-adversary (\textbf{left}).
        However, fine-tuning our cylic-adversary enables it to defeat KataGo once again (\textbf{right}).
        See Appendix~\ref{app:adv-training} for more detailed versions of the above plots.
    }
    \label{fig:defense}
\end{figure*}

We evaluate the ability of our cyclic-adversary to exploit victims playing \emph{with} search and find that it still achieves high win rates
by tricking its victims into making severe mistakes a human would avoid (see Appendix~\ref{app:kellin-donut-analysis}).

The cyclic-adversary achieves a win rate of 95.7\% (over 1052 games) against \hardened{\cpfivezerofive{}} with 4096 visits.
The adversary also achieves a 97.3\% win rate (over 1000 games) against an undefended \cpfivezerofive{} with 4096 visits, verifying that our adversary is not exploiting anomalous behavior introduced by the pass-alive defense.

We also tested our cyclic-adversary against \cpfivezerofive{} with substantially higher victim visits. The adversary (using \ourmctssampleabbrev{} with 600 visits/move)
achieved an 82\% win rate over 50 games against \cpfivezerofive{} with $10^6$ visits/move, and a 72\% win rate over 50 games against \cpfivezerofive{} with $10^7$ visits/move, using 10 and 1024 search threads respectively (see Appendix~\ref{app:hyperparameters}).
This demonstrates that search is not a practical defense against the attack: $10^7$ visits is already prohibitive in many applications, taking over one hour per move to evaluate even on high-end consumer hardware~\citep{insis2022}.
Indeed, \citet{tian2019} used two orders of magnitude less search than this even in tournament games against human professionals. 

That said, the adversary win rate does decrease with more victim search. This is shown in Figure~\ref{fig:evaluation:search:victim-visits-hardened}, and is even more apparent against a weaker adversary 
(Figure~\ref{fig:app:s497mil-amcts-r-vs-victim-visits}). The victim also judges decisive positions more accurately with more search (Appendix~\ref{app:role-of-search}). We conclude that search is a valid tool for improving robustness, but will not produce fully robust agents on its own.

We examine adversary search in Figure~\ref{fig:evaluation:search:adv-visits-hardened}. For a fixed victim search budget, the adversary does best at 128--600 visits, and \ourmctssamplesymmetryabbrev{} performs no better than the computationally cheaper \ourmctssampleabbrev{}.
Intriguingly, increasing adversary visits beyond 600 does not help and may even hurt performance, suggesting the adversary's strategy does not benefit from greater look-ahead.

In experiments with an earlier checkpoint of the cyclic-adversary, we saw \ourmctsperfectabbrev{} outperform \ourmctssampleabbrev{} (the latter of which incorrectly models the victim as having no search; see Figure~\ref{fig:app:s497mil-amcts-r-vs-victim-visits}). With our current version of the cyclic-adversary, \ourmctssampleabbrev{} does so well that \ourmctsperfectabbrev{} cannot provide any improvement up to 128 victim visit (Figure~\ref{fig:evaluation:search:victim-visits-hardened}). The downside of \ourmctsperfectabbrev{} is that it drastically increases the amount of compute, to the point that it is impractical in this context to evaluate \ourmctsperfectabbrev{} at higher visit counts. However, we do find indications that \ourmctsperfectabbrev{} helps in high-victim-visit regimes, with the benefits being visible even with very limited recursive victim simulation. We include an initial analysis of this phenomenon in Appendix~\ref{app:recursive}.

\subsection{Defense}
\label{sec:evaluation:defense}
In mid-December 2022, KataGo's official distributed training run was modified so that 0.08\% of its self-play games start from positions where the cyclic-exploit is in the process of being carried out. This mild form of adversarial training is designed to improve KataGo's understanding of cyclic positions while preserving KataGo's strength in normal games.
The performance of our cyclic-adversary$^\text{600 visits}$ dropped steadily after this was introduced (Figure~\ref{fig:defense:b60-adv-train-curve}), reaching a low of 0 / 50 won games against the $\texttt{b60-s7702m}^\text{32 visits}$ KataGo agent\footnote{\texttt{b60-s7702m} refers to the \texttt{b60c320-s7701878528} network released on \href{https://katagotraining.org/networks/}{May 17th, 2023}. This was the most recent 60-block network available at the time of conducting our research.} and 119 / 2050 won games against $\texttt{b60-7702m}^\text{1 visit}$.

However, after fine-tuning the cyclic-adversary for an additional
1154.9 V100 GPU days
against adversarially trained networks, it recovers its exploitation abilities, achieving a 47\% win rate over 400 games against $\texttt{b60-s7702m}^\text{4096 visits}$ and a 17.5\% win rate over 40 games against $\texttt{b60-s7702m}^\text{100,000 visits}$. These wins still rely on the cyclic-exploit, although carried out in a slightly different way. See Appendix~\ref{app:adv-training} for a sample game and details on KataGo's defense and our cyclic-adversary's fine-tuning.

In summary, while a small amount of training on adversarial positions is enough to robustly defend against a fixed adversary, such a defense does not generalize and can be broken again by fine-tuning the fixed adversary. However, such fine-tuning requires more compute per unit of win rate compared to attacking non-adversarially trained networks (compare Figure~\ref{fig:defense:adv-finetuning} with Figure~\ref{fig:compute:win-rate-vs-gpu-days-cyclic}), so it is plausible that with much more adversarial training, KataGo could become computationally infeasible to exploit. Computing more precise scaling laws for this type of adversarial training is a fruitful direction for future work.

\vspace{-0.6mm}
\subsection{\textSwitch{Understanding the cyclic-adversary}{Understanding the Cyclic-Adversary}}
\label{sec:evaluation:understanding-the-attack}

\vspace{-0.2mm}
Qualitatively, the cyclic-adversary we trained in Section~\ref{sec:evaluation:no-search-hardened} wins by first coaxing KataGo into creating a large group of stones in a circular pattern, and then exploiting a weakness in KataGo's network which allows the adversary to capture the group. This causes the score to shift decisively and unrecoverably in the adversary's favor.

\vspace{-0.15mm}
We test several hard-coded baseline attacks in Appendix~\ref{app:baseline-attacks}.
We find that none of the attacks work well against \hardened{\cpfivezerofive{}}, although the \emph{Edge} baseline playing as white wins almost half of the time against the undefended \cpfivezerofive{}.
This provides evidence that \hardened{\cpfivezerofive{}} is more robust than \cpfivezerofive{}, and that the cyclic-adversary has learned a relatively sophisticated exploit.

\vspace{-0.15mm}
To better understand the attack, we examined the win rate predictions produced by both the adversary's and the victim's value networks at each turn of a game. 
Typically the victim predicts that it will win with over 99\% confidence
for most of the game, then suddenly realizes it will lose with high probability, often just \emph{one move} before its cyclic group is captured.
This trend is depicted in Figure~\ref{fig:win rate:average}: in games the adversary wins, the victim's prediction loss is elevated throughout the majority of the game, only dipping close to the self-play baseline around 40-50 moves from the end of the game.
In some games, we observe that the victim's win rate prediction oscillates wildly before finally converging on certainty that it will lose (Figure~\ref{fig:win rate:example}). 
This is in stark contrast to the adversary's own predictions, which change much more slowly and are less confident.

\begin{figure}[ht]
    \centering
    \begin{minipage}{\columnwidth}
        \input{figs/katagovisualizer/winrate-avg.pgf}
        \vspace{-6mm}
        \caption{%
            Binary cross-entropy loss of \cpfivezerofive{}$^\text{4096 visits}$'s prediction of the game result
            over the course of the games played against the cyclic-adversary$^\text{600 visits}$.
            The green and purple curves are averaged over games won by the adversary and victim respectively.
            The blue curve is averaged over self-play games and serves as a baseline.
            Shaded regions denote $\pm1$ SD.
        }
        \label{fig:win rate:average}
    \end{minipage}
    \\[5mm]
    \begin{minipage}{\columnwidth}
        \input{figs/katagovisualizer/winrate-sample.pgf}
        \vspace{-7mm}
        \caption{%
            Probability of victory as predicted by the cyclic-adversary and \cpfivezerofive{}
            for a portion of a randomly selected game.
            Note the sudden changes in win rate prediction between moves 248 and 272 during a ko fight.
            \href{\demosite/game-analysis\#win-rate}{Explore the game}.
        }
        \label{fig:win rate:example}
    \end{minipage}
    \vspace{-5mm}
\end{figure}

\vspace{-0.15mm}
\emph{Why} does the victim misjudge these cyclic positions so severely? 
To understand this, we studied the differences in the activations of the victim between cyclic and minimally perturbed non-cyclic positions.
We found that a few channels at layer 26 show a clear divergence between cyclic and non-cyclic positions, as illustrated in Figure~\ref{fig:activations-L25v26}, whereas earlier layers showed no comparable trend.
Moreover, we found that the difference in activations between \cpfivezerofive{} and the adversarially trained \texttt{cp580} shows a similar pattern, suggesting that adversarial training preferentially changes the behavior of the network on cyclic positions at these channels.
These results provide a clear area for further investigation that could lead to a more detailed mechanistic understanding of this and similar vulnerabilities. Further analysis is available in Appendix~\ref{app:activations}.

\begin{figure}[ht]
    \centering
    \begin{subfigure}[t]{0.22\textwidth}
        \centering
        \includesvg[inkscapelatex=false, width=0.98\textwidth]{figs/boardstates/activations/connection_test_realgame2_A-2col.svg}
        \captionsetup{justification=centering}
        \caption{Position \textbf{B} (or \textbf{W} if the color of the {\color{red} $\times$} is flipped).}
    \end{subfigure}
    \raisebox{1.5cm}{
        \begin{subfigure}[t]{0.23\textwidth}
            \centering
            \begin{subfigure}[t]{0.48\textwidth}
                \centering
                \includesvg[inkscapelatex=false, width=\textwidth]{figs/boardstates/activations/synthesis/cp505_realgame2_A_vs_b40_1286_realgame2_A-L25-small.svg}
                \caption{}
            \end{subfigure}
            \begin{subfigure}[t]{0.48\textwidth}
                \centering
                \includesvg[inkscapelatex=false, width=\textwidth]{figs/boardstates/activations/synthesis/cp505_realgame2_A_vs_cp505_realgame2_B-L25-small.svg}
                \caption{}
            \end{subfigure}
            \captionsetup{justification=centering}
            \caption*{
                Layer\textbf{ 25}\\[1mm]
                b. \cpfivezerofive{}(\textbf{B}) -- \texttt{cp580}(\textbf{B})\\[1mm]
                c. \cpfivezerofive{}(\textbf{B}) -- \cpfivezerofive{}(\textbf{W})
            }
        \end{subfigure}
    }
    \\[2mm]
    \begin{subfigure}[t]{0.22\textwidth}
        \centering
        \includesvg[inkscapelatex=false, width=0.98\textwidth]{figs/boardstates/activations/synthesis/cp505_realgame2_A_vs_b40_1286_realgame2_A-L26-small.svg}
        \captionsetup{justification=centering}
        \caption{Layer\textbf{ 26}\\
                \cpfivezerofive{}(\textbf{B}) -- \texttt{cp580}(\textbf{B})}
    \end{subfigure}
    \quad
    \begin{subfigure}[t]{0.22\textwidth}
        \centering
        \includesvg[inkscapelatex=false, width=0.98\textwidth]{figs/boardstates/activations/synthesis/cp505_realgame2_A_vs_cp505_realgame2_B-L26-small.svg}
        \captionsetup{justification=centering}
        \caption{Layer\textbf{ 26}\\
                \cpfivezerofive{}(\textbf{B}) -- \cpfivezerofive{}(\textbf{W})}
    \end{subfigure}
    \vspace{0.5mm}
    \caption{%
        Comparison of activations of \cpfivezerofive{}
        and \texttt{cp580} (a checkpoint adversarially trained to defend against the cyclic adversary)
        on a cyclic (\textbf{B}) and non-cyclic position (\textbf{W})
        which differ by a single stone (a).
        We plot differences in activations (b-e);
         brighter colors indicate larger differences.
        In layer 25 (b,c) the activations are fairly similar.
        In layer 26 (d,e) there are strong differences localized to a few channels.
        Adversarial training (d) changes these channel activations
	in a similar manner to breaking the cycle \textbf{B} $\to$ \textbf{W} (e), suggesting these channels are linked to the cyclic vulnerability.
    }
    \label{fig:activations-L25v26}
    \vspace{-4.5mm}
\end{figure}

\subsection{Transfer}
\label{sec:evaluation:transfer}

In Appendix~\ref{app:transfer:algorithmic} we evaluate our cyclic-adversary (trained only on KataGo) in zero-shot transfer against two different superhuman Go agents, Leela Zero and ELF OpenGo.
This setting is especially challenging because \ourmctsabbrev{} models the victim as being KataGo and will be continually surprised by the moves taken by the Leela or ELF opponent.
Nonetheless, the adversary wins 6.1\% of games against Leela and 3.5\% of games against ELF.

In Appendix~\ref{app:transfer:human} one of our authors, a Go expert, was able to learn from our adversary's game records to implement this attack without any algorithmic assistance.
Playing in standard human conditions on the online Go server KGS they obtained a greater than 90\% win rate against a top ranked KataGo bot that is unaffiliated with the authors.
The author even won giving the bot 9 handicap stones, an enormous advantage: a human professional with this handicap would have a virtually 100\% win rate against any opponent, whether human or AI.
They also beat KataGo and Leela Zero playing with 100k visits each, which is normally far beyond human capabilities. Other humans have since used cyclic attacks to beat a variety of other top Go AIs (Section~\ref{sec:repro}).

These results confirm that the cyclic vulnerability is present in a range of bots under a variety of configurations.
They also further highlight the significance and interpretability of the exploit our algorithmic adversary finds.
The adversary is not finding, for instance, just a special sequence of moves,  but a strategy that a human can learn and act on. In addition, in both algorithmic and human transfer, the attacker does not have access to the victim model's weights, policy network output, or even a large number of games to learn from. This increases the threat level and suggests, for example, that one could learn an attack on an open-source system and then transfer it to a closed-source model.

\vspace{-0.5mm}
\section{\textSwitch{Limitations and future work}{Limitations and Future Work}}

We demonstrate that even superhuman agents can be vulnerable to adversarial policies.
However,
it is possible Go-playing AI systems are unusually vulnerable.
Thus a promising direction for future work is to evaluate our attack against strong AI systems in other games and settings.

It is also natural to ask how we can \emph{defend} against adversarial policies.
A first attempt was made by the KataGo team after we published an earlier version of this work, but we show in Section~\ref{sec:evaluation:defense} that this defense is as of yet inadequate.
Fortunately, there are a number of other promising multi-agent RL techniques.
One such technique is counterfactual regret minimization~\citep[CFR]{zinkevich2007}, which can beat professional human poker players~\citep{brown2017}.
CFR has difficulty scaling to high-dimensional state spaces, but regularization methods~\citep{perolat2021} can scale to games such as Stratego with a game tree $10^{175}$ times larger than Go~\citep{perolat2022}.
Alternatively, methods using populations of agents such as policy-space response oracles~\citep{lanctot2017}, AlphaStar's Nash league~\citep{vinyals2019} or population-based training \citep{czempin2022} may be more robust than self-play, at the cost of greater computation.

Finally, we found it harder to exploit agents that use search, with our attacks achieving a lower win rate and requiring more computational resources. An interesting direction for future work is to look for more effective and compute-efficient methods for attacking agents that use large amounts of search, such as learning a computationally efficient model of the victim (Appendix~\ref{app:search:our-mcts-model}).

\vspace{-0.5mm}
\section{Conclusion}

We trained adversarial policies that exploit superhuman Go AIs.
Notably, our adversaries do not win by playing a strong game of Go.
Instead, they exploit blind spots in their victims. This result suggests that even highly capable agents can harbor serious vulnerabilities.

KataGo was published in 2019 and has since been used by many Go enthusiasts and professional players as a playing partner and analysis engine~\citep{wu2019}.
Despite this public scrutiny, to the best of our knowledge the vulnerabilities discussed in this paper were never previously exploited.
This suggests that learning-based attacks like the ones developed in this paper may be an important tool for uncovering hard-to-find vulnerabilities in AI systems.

Our results underscore that improvements in capabilities do not always translate into adequate robustness.
Failures in Go AI systems are entertaining, but similar failures in safety-critical systems like automated financial trading or autonomous vehicles could have dire consequences.
We believe that the ML research community should invest in improving robust training and adversarial defense techniques in order to produce models with the high levels of reliability needed for safety-critical systems.

\subsection{\textSwitch{Reproducibility statement}{Reproducibility Statement}}
\label{sec:repro}
We take the following steps to promote and ensure reproducibility:
\begin{itemize}
    \item Our code is available at \githuburl{}. The code is containerized and includes instructions for running it.
    \item We make many game records available through our \href{\demosite/}{website}. We will make more game records available to researchers upon request and have already provided game records to David Wu, the creator and primary developer of KataGo, for use in KataGo's training process.
    \item We set up a bot running the most recent checkpoint of our cyclic-adversary on the KGS Go server, under the username \href{https://www.gokgs.com/graphPage.jsp?user=Adversary0}{\texttt{Adversary0}}. This bot was available for the public to play for a period of a month. See Appendix~\ref{app:experiments:human-vs-adversary} for more details.
\end{itemize}

A number of our key results have already been reproduced:
\begin{itemize}
    \item
    The vulnerability to the passing attack has been independently confirmed by David Wu.
    
    \item
    The vulnerability to the cyclic attack has been independently confirmed by David Wu, as well as many others in the computer Go community.
    
    \item
    The cyclic vulnerability and the adversary's ability to use it has been replicated through normal bot play against the KGS bot we made available, as has the result that novice human play beats the adversary.

    \item
    Human ability to use the cyclic attack has been independently reproduced against
    \href{https://online-go.com/game/53421988}{KataGo}, %
    as well as in transfer settings against
    \href{https://online-go.com/game/51321265}{ELF OpenGo},
    \href{https://h5.foxwq.com/txwqshare/index.html?chessid=1676910620010001365&boardsize=19}{FineArt},
    \href{https://online-go.com/game/51356405}{Leela Zero},
    and \href{https://online-go.com/game/51375020}{Sai}.
\end{itemize}

\vfill\break
\subsection{Acknowledgments}
Thanks to David Wu and the Computer Go Community Discord for sharing their knowledge of computer Go with us and for their helpful advice on how to work with KataGo,
to Adri\`a Garriga-Alonso for feedback and assistance setting up activation analysis and infrastructure,
to Lawrence Chan, Euan McLean, and Niki Howe for their feedback on earlier drafts of the paper,
to ChengCheng Tan and Alyse Spiehler for assistance preparing illustrations,
to David Fontaine for help with debugging KataGo deadlocks, %
to Matthew Harwit for help with Chinese communication and feedback especially on Go analysis,
to Daniel Filan for Go game analysis and feedback on project direction,
and to Nir Shavit for his support of and high level feedback on the project. %

Tony Wang was supported by funding from the Eric and Wendy Schmidt Center at the Broad Institute of MIT and Harvard. %

\subsection{\textSwitch{Author contributions}{Author Contributions}}
\iffinalcopy
Tony Wang invented and implemented the \ourmctssampleabbrev{} algorithm,
made several other code contributions, and ran and analyzed many of the experiments.
Adam Gleave managed the project, wrote the core of the paper, suggested the curriculum approach, helped manage the cluster experiments were run on, and implemented some minor features.
Tom Tseng implemented and ran transfer experiments, trained and ran experiments with the pass-hardening defense enabled, and ran many of the evaluations.
Kellin Pelrine (our resident Go expert) provided analysis of our adversary's strategy, search vs. robustness, activation analysis, and manually reproduced the cyclic-attack against different Go AIs.
Nora Belrose implemented and ran the experiments for baseline adversarial policies, and our pass-hardening defense.
Joseph Miller developed the website showcasing the games, and an experimental dashboard for internal use.
Michael Dennis developed an adversarial board state for KataGo that inspired us to pursue this project, and contributed a variety of high-level ideas and guidance such as adaptations to MCTS.
Yawen Duan ran some of the initial experiments and investigated the adversarial board state.
Viktor Pogrebniak implemented the curriculum functionality and improved the KataGo configuration system.
Sergey Levine and Stuart Russell provided guidance and general feedback.
\else
Removed for double blind submission.
\fi

\clearpage
\section*{References}
\addcontentsline{toc}{section}{References}
\begingroup
    \renewcommand{\section}[2]{}%
    \bibliography{refs}
    \bibliographystyle{icml2023}
\endgroup

\appendix
\onecolumn

\section{\textSwitch{Rules of Go used for evaluation}{Rules of Go Used for Evaluation}}
\label{app:rules}

We evaluate all games with Tromp-Taylor rules~\citep{tromp:2014}, after clearing opposite-color stones within pass-alive groups computed by Benson's algorithm~\citep{benson1976}.
Games end after both players pass consecutively, or once all points on the board belong to a pass-alive group or pass-alive territory (defined in Appendix~\ref{app:pass-alive}).
KataGo was configured to play using these rules in all our matches against it.
Indeed, these rules simply consist of KataGo's version of Tromp-Taylor rules with \texttt{SelfPlayOpts} enabled~\citep{wu2021katago}.
We use a fixed Komi of 6.5.

We chose these \emph{modified Tromp-Taylor} rules because they are simple, and KataGo was trained on (variants) of these rules so should be strongest playing with them.
Although the exact rules used were randomized during KataGo's training, modified Tromp-Taylor made up a plurality of the training data.
That is, modified Tromp-Taylor is at least as likely as any other configuration seen during training, and is more common than some other options.\footnote{In private communication, the author of KataGo estimated that modified Tromp-Taylor made up a ``a few \%'' of the training data, ``growing to more like 10\% or as much as 20\%'' depending on differences such as ``self-capture and ko rules that shouldn't matter for what you're investigating, but aren't fully the same rules as Tromp-Taylor''.}

In particular, KataGo training randomized between area vs.\ territory scoring as well as ko, suicide, taxation and button rules from the options described in \citet{wu2021katago}.
These configuration settings are provided as input to the neural network~\citep[Table~4]{wu2019}, so the network should learn to play appropriately under a range of rule sets.
Additionally, during training komi was sampled randomly from a normal distribution with mean 7 and standard deviation 1~\citep[Appendix~D]{wu2019}.

\subsection{\textSwitch{Difference from typical human play}{Difference from Typical Human Play}}
Although KataGo supports a variety of rules, all of them involve automatically scoring the board at the end of the game.
By contrast, when a match between humans end, the players typically confer and agree which stones are dead, removing them from the board prior to scoring.
If no agreement can be reached then either the players continue playing the game until the situation is clarified, or a referee arbitrates the outcome of the game.

KataGo has a variety of optional features to help it play well under human scoring rules.
For example, KataGo includes an auxiliary prediction head for whether stones are dead or alive.
This enables it to propose which stones it believes are dead when playing on online Go servers.
Additionally, it includes hard-coded features that can be enabled to make it play in a more human-like way, such as \texttt{friendlyPassOk} to promote passing when heuristics suggest the game is nearly over.

These features have led some to speculate that the (undefended) victim passes prematurely in games such as those in Figure~\ref{fig:cp505vis1-board:unhardened} because it has learned or is configured to play in a more human-like way.
\emph{Prima facie}, this view seems credible: a human player certainly might pass in a similar situation to our victim, viewing the game as already won under human rules.
Although tempting, this explanation is not correct: the optional features described above were disabled in our evaluation.
Therefore KataGo loses under the rules it was both trained and configured to use.

In fact, the majority of our evaluation used the \texttt{match} command to run KataGo vs.\ KataGo agents which naturally does not support these human-like game play features.
We did use the \texttt{gtp} command, implementing the Go Text Protocol (GTP), for a minority of our experiments, such as when evaluating KataGo against other AI systems or human players and when evaluating our adversary against KataGo with $10^7$ visits.
In those experiments, we configured \texttt{gtp} to follow the same Tromp-Taylor rules described above, with any human-like extensions disabled.

\clearpage
\section{\textSwitch{Search algorithms}{Search Algorithms}}
\label{app:search-algorithms}

\subsection{\textSwitch{A review of Monte-Carlo tree search (MCTS)}{A Review of Monte-Carlo Tree Search (MCTS)}}
\label{app:mcts-review}
In this section, we review the basic Monte-Carlo Tree Search (MCTS) algorithm as used in AlphaGo-style agents~\citep{silver2016}.
This formulation is heavily inspired by the description of MCTS given in \citet{wu2019}.

MCTS is an algorithm for growing a game tree one node at a time.
It starts from a tree $T_0$ with a single root node $x_0$.
It then goes through $N$ \emph{playouts},
where every playout adds a leaf node to the tree.
We will use $T_i$ to denote the game tree after $i$ playouts,
and will use $x_i$ to denote the node that was added to $T_{i - 1}$ to get $T_i$.
After MCTS finishes, we have a tree $T_N$ with $N + 1$ nodes.
We then use simple statistics of $T_N$
to derive a sampling distribution for the next move.

\subsubsection{\textSwitch{MCTS playouts}{MCTS Playouts}}
MCTS playouts are governed by two learned functions:
\begin{enumerate}
    \item[a.]
    A value function estimator
    $\hat{V}: \mathcal{T} \times \mathcal{X} \to \mathbb{R}$,
    which returns a real number $\hat{V}_T(x)$ given a tree $T$ and a node $x$ in $T$ (where $\mathcal{T}$ is the set of all trees, and $\mathcal{X}$ is the set of all nodes).
    The value function estimator is meant to estimate
    how good it is to be at $x$
    from the perspective of the player to move at the root of the tree.
    
    \item[b.]
    A policy estimator
    $\hat{\pi}: \mathcal{T} \times \mathcal{X} \to \mathcal{P}(\mathcal{X})$,
    which returns a probability distribution over possible next states
    $\hat{\pi}_T(x)$
    given a tree $T$ and a node $x$ in $T$.
    The policy estimator is meant to approximate the
    result of playing the optimal policy from $x$
    (from the perspective of the player to move at $x$).
\end{enumerate}
For both KataGo and AlphaGo,
the value function estimator and policy estimator
are defined by two deep neural network heads with a shared backbone.
The reason that $\hat{V}$ and $\hat{\pi}$ also take a tree $T$ as an argument
is because the estimators factor in the sequence of moves leading up to a node in the tree.

A playout is performed by taking a walk in the current game tree $T$. 
The walk goes down the tree until
it attempts to walk to a node $x'$
that either doesn't exist in the tree
or is a terminal node.\footnote{
A ``terminal'' node is one where the game is finished,
whether by the turn limit being reached,
one player resigning,
or by two players passing consecutively.
}
At this point the playout ends
and $x'$ is added as a new node to the tree
(we allow duplicate terminal nodes in the tree).

Walks start at the root of the tree.
Let $x$ be where we are currently in the walk.
The child $c$ we walk to
(which may not exist in the tree)
is given by
\begin{align}
    &\text{walk}^\text{MCTS}_{T}(x) \notag \\
    &=
    \begin{cases}
        \underset{c}{\mathrm{argmax}}\quad
        \bar{V}_T(c)
        + \alpha \cdot \hat{\pi}_T(x)[c]
                 \cdot \frac{\sqrt{S_T(x) - 1}}{1 + S_T(c)}
        & \text{if root player to move at $x$},  \\
        \underset{c}{\mathrm{argmin}}\quad\,
        \bar{V}_T(c)
        - \alpha \cdot \hat{\pi}_T(x)[c]
                 \cdot \frac{\sqrt{S_T(x) - 1}}{1 + S_T(c)}
        & \text{if opponent player to move at $x$},
    \end{cases}
    \label{eqn:app:mcts-playout-dist}
\end{align}
where the argmin and argmax are taken over all children
reachable in a single legal move from $x$.
There are some new pieces of notation in Eq~\ref{eqn:app:mcts-playout-dist}.
Here is what they mean:
\begin{enumerate}
    \item$\bar{V}_T: \mathcal{X} \to \mathbb{R}$ takes a node $x$ and returns the average value of $\hat{V}_T$ across all the nodes in the subtree of $T$ rooted at $x$ (which includes $x$).
    In the special case that $x$ is a terminal node,
    $\bar{V}_T(x)$ is the result of the finished game as given by the game-simulator.
    When $x$ does not exist in $T$,
    we instead use the more complicated formula\footnote{
    Which is used in KataGo and LeelaZero but not AlphaGo~\citep{wu2019}.
    }
    \begin{equation*}
        \bar{V}_T(x) = 
        \bar{V}_T(\mathrm{par}_T(x))
        - \beta \cdot \sqrt{
        \sum_{x' \,\in\, \mathrm{children}_T(\mathrm{par}_T(x))}
        \hat{\pi}_T(\mathrm{par}_T(x))[x']
        }\;\,,
    \end{equation*}
    where $\mathrm{par}_T(x)$ is the parent of $x$ in $T$
    and $\beta$ is a constant that controls
    how much we de-prioritize exploration
    after we have already done some exploration.
    
    \item $\alpha \geq 0$ is a constant to trade off between exploration and exploitation.
    
    \item $S_T: \mathcal{X} \to \mathbb{Z}_{\geq 0}$
    takes a node $x$ and
    returns the size of the subtree of $T$ rooted at $x$.
    Duplicate terminal nodes are counted multiple times.
    If $x$ is not in $T$, then $S_T(x) = 0$.
\end{enumerate}

In Eq~\ref{eqn:app:mcts-playout-dist}, one can interpret the first term as biasing the search towards exploitation, and the second term as biasing the search towards exploration. The form of the second term is inspired by UCB algorithms.

\subsubsection{\textSwitch{MCTS final move selection}{MCTS Final Move Selection}}
The final move to be selected by MCTS is
sampled from a distribution proportional to
\begin{equation}
    S_{T_N}(c)^{1 / \tau},
\end{equation}
where $c$ in this case is a child of the root node.
The temperature parameter $\tau$ 
trades off between exploration and exploitation.\footnote{
See
\href{https://github.com/lightvector/KataGo/blob/21b4efef6bf8c9dd72bb79abd3703281b2878fd1/cpp/search/search.h\#L265-L267}{\texttt{search.h::getChosenMoveLoc}}
and
\href{https://github.com/lightvector/KataGo/blob/d8d0cd76cf73df08af3d7061a639488ae9494419/cpp/search/searchresults.cpp\#L420-L438}{\texttt{searchresults.cpp::getChosenMoveLoc}} to see how KataGo does this.}

\subsubsection{\textSwitch{Efficiently implementing MCTS}{Efficiently Implementing MCTS}}
To efficiently implement the playout procedure
one should keep running values of $\bar{V}_T$ and $S_T$ for every node in the tree.
These values should be updated whenever a new node is added.
The standard formulation of MCTS bakes these updates into the algorithm specification.
Our formulation hides the procedure for computing $\bar{V}_T$ and $S_T$ to simplify exposition.

In addition, neural network evaluations of each node should only be performed once and a cached evaluation should be used when revisiting a node during a subsequent walk down the tree.

Our adversarial variants of MCTS use both of the above speedups.

\subsection[\ourmctssampleabbrev{}]{\ourmctssamplelong{} (\ourmctssampleabbrev{})}
\label{app:search:our-mcts-sample}

In this section, we describe in detail how our \ourmctssamplelong{} (\ourmctssampleabbrev{}) attack is implemented.
We build off of the framework for vanilla MCTS
as described in Appendix~\ref{app:mcts-review}.

\ourmctssampleabbrev{}, just like MCTS,
starts from a tree $T_0$ with a single root node
and adds nodes to the tree via a series of $N$ playouts.
We derive the next move distribution from the final game tree $T_N$
by sampling from the distribution proportional to
\begin{equation}
    S^\text{\ourmctsabbrev{}}_{T_N}(c)^{1 / \tau},
    \quad \text{where $c$ is a child of the root node of $T_N$}.
\end{equation}
Here, $S^\text{\ourmctsabbrev{}}_T$
is a modified version of $S_T$
that measures the size of a subtree
while ignoring non-terminal victim-nodes
(at victim-nodes it is the victim's turn to move,
and at self-nodes it is the adversary's turn to move).
Formally, $S^\text{\ourmctsabbrev{}}_T(x)$
is the sum of the weights of nodes in the subtree of $T$ rooted at $x$,
with weight function
\begin{align}
    w^\text{\ourmctsabbrev{}}_T(x)
    &=\begin{cases}
        1 & \text{if $x$ is self-node}, \\
        1 & \text{if $x$ is terminal victim-node}, \\
        0 & \text{if $x$ is non-terminal victim-node}.
    \end{cases}
    \label{eqn:ourmcts-weights}
\end{align}

We grow the tree by \ourmctsabbrev{} playouts.
At victim-nodes, we sample directly from the victim's policy $\pi^v$:
\begin{equation}
\label{eq:app:our-mcts-sample}
    \text{walk}^\text{\ourmctsabbrev{}}_T(x) := \text{sample from}\;\pi^v_T(x).
\end{equation}
This is a perfect model of the victim \emph{without} search.
However, it will tend to underestimate the strength of the victim when the victim plays with search.

At self-nodes, we instead take the move with the best upper confidence bound
just like in regular MCTS:
\begin{equation}
    \text{walk}^\text{\ourmctsabbrev{}}_T(x)
    := \displaystyle
        \underset{c}{\mathrm{argmax}}\quad
        \bar{V}^\text{\ourmctsabbrev{}}_T(c)
        + \alpha \cdot \hat{\pi}_T(x)[c]
                 \cdot \frac{\sqrt{S^\text{\ourmctsabbrev{}}_T(x) - 1}}
                            {1 + S^\text{\ourmctsabbrev{}}_T(c)}.
\end{equation}
Note this is similar to Eq~\ref{eqn:app:mcts-playout-dist} from the previous section.
The key difference is that
we use $S^\text{\ourmctsabbrev{}}_T(x)$
(a weighted version of $S_T(x)$)
and $\bar{V}^{\text{\ourmctsabbrev{}}}_T(c)$
(a weighted version of $\bar{V}_T(c)$).
Formally,
$\bar{V}^{\text{\ourmctsabbrev{}}}_T(c)$ is
the weighted average of the value function estimator $\hat{V}_T(x)$
across all nodes $x$ in the subtree of $T$ rooted at $c$,
weighted by $w^\text{\ourmctsabbrev{}}_T(x)$.
If $c$ does not exist in $T$ or is a terminal node,
we fall back to the behavior of $\bar{V}_T(c)$.

\subsection[\ourmctssamplesymmetryabbrev{}]{\ourmctssamplesymmetrylong{} (\ourmctssamplesymmetryabbrev{})}
\label{app:search:our-mcts-symmetry}
When computing the policy estimator $\hat\pi$ for the root node of a MCTS search (and when playing without tree-search, i.e. "policy-only"), KataGo will pass in different rotated/reflected copies of the game-board and average their results in order to obtain a more stable and symmetry-equivariant policy.
That is
\begin{equation*}
    \hat\pi_\text{root} = \frac{1}{|S|} \sum_{g \in S \subseteq D_4} g^{-1} \circ \hat\pi \circ g,
\end{equation*}
where $D_4$ is the symmetry group of a square (with 8 symmetries) and $S$ is a randomly sampled subset of $D_4$.\footnote{
     See \href{https://github.com/lightvector/KataGo/blob/eea5d0cbc3909ef55b81b364b19593d83f7aefa8/cpp/search/searchnnhelpers.cpp\#L63-L79}{\texttt{searchhelpers.cpp::initNodeNNOutput}} for how the symmetry averaging is implemented
     in KataGo. The size of $|S|$ is configured via the KataGo parameter \texttt{rootNumSymmetriesToSample}.
}

In \ourmctsabbrev{}, we ignore this symmetry averaging because modeling it would inflate the cost of simulating our victim by up to a factor of 8. By contrast, \ourmctssamplesymmetryabbrev{} accurately models this symmetry averaging at the cost of increased computational requirements.

\subsection[\ourmctsperfectabbrev{}]{\ourmctsperfectlong{} (\ourmctsperfectabbrev{})}
\label{app:search:our-mcts-perfect}
In \ourmctsperfectabbrev{}, we simulate the victim by starting a new (\emph{recursive}) MCTS search.
We use this simulation at victim-nodes, replacing the victim sampling step (Eq.~\ref{eq:app:our-mcts-sample}) in \ourmctssampleabbrev{}.
This simulation will be a perfect model of the victim when the MCTS search is configured to use the same number of visits and other settings as the victim.
However, since MCTS search is stochastic, the (random) move taken by the victim may still differ from that predicted by \ourmctsperfectabbrev{}.
Moreover, in practice, simulating the victim with its full visit count at every victim-node in the adversary's search tree can be prohibitively expensive.

\subsection[\ourmctsmodelabbrev{}]{\ourmctsmodellong{} (\ourmctsmodelabbrev{})}
\label{app:search:our-mcts-model}
In \ourmctsmodelabbrev{}, we propose fine-tuning a copy of the victim network to predict the moves played by the victim in games played against the adversarial policy.
This is similar to how the victim network itself was trained, but may be a better predictor as it is trained on-distribution. %
The adversary follows the same search procedure as in \ourmctssampleabbrev{} but samples from this predictive model instead of the victim.

\ourmctsmodelabbrev{} has the same inference complexity as \ourmctssampleabbrev{}, and is much cheaper than \ourmctsperfectabbrev{}.
However, it does impose a slightly greater training complexity due to the need to train an additional network.
Additionally, \ourmctsmodelabbrev{} requires white-box access in order to initialize the predictor to the victim network.

In principle, we could randomly initialize the predictor network, making the attack black-box.
Notably, imitating the victim has led to successful black-box adversarial policy attacks in other domains~\citep{bui2022}.
However, a randomly initialized predictor network would likely need a large number of samples to imitate the victim.
\citet{bui2022} use tens of millions of time steps to imitate continuous control policies, and we expect this number to be still larger in a game as complex as Go.

\subsection{\textSwitch{Pass-alive defense}{Pass-Alive Defense}}
\label{app:pass-alive}
Our hard-coded defense modifies KataGo's C++ code to directly remove passing moves from consideration after MCTS, setting their probability to zero. Since the victim must eventually pass in order for the game to end, we allow passing to be assigned nonzero probability when there are no legal moves, \emph{or} when the only legal moves are inside the victim's own pass-alive territory. We also do not allow the victim to play within its own pass-alive territory---otherwise, after removing highly confident pass moves from consideration, KataGo may play unconfident moves within its pass-alive territory, losing liberties and eventually losing the territory altogether. We use a pre-existing function inside the KataGo codebase, {\fontfamily{pcr}\selectfont Board::calculateArea}, to determine which moves are in pass-alive territory.

The term ``pass-alive territory'' is defined in the KataGo rules as follows~\citep{wu2021katago}:
\begin{quote}
    A \{maximal-non-black, maximal-non-white\} region R is \emph{pass-alive-territory} for \{Black, White\} if all \{black, white\} regions bordering it are pass-alive-groups, and all or all but one point in R is adjacent to a \{black, white\} pass-alive-group, respectively.
\end{quote}

The notion ``pass-alive group'' is a standard concept in Go~\citep{wu2021katago}:
\begin{quote}
    A black or white region R is a \emph{pass-alive-group} if there does not exist any sequence of consecutive pseudolegal moves of the opposing color that results in emptying R.
\end{quote}

KataGo uses an algorithm introduced by \citet{benson1976} to efficiently compute the pass-alive status of each group. For more implementation details, we encourage the reader to consult the official KataGo rules and the KataGo codebase on GitHub.

\subsubsection{\textSwitch{Vulnerability of defense in seki situations}{Vulnerability of Defense in Seki Situations}}
\label{app:pass-alive-defense-vulnerability}

Training against defended victims resulted in the cyclic-adversary which successfully exploits both the defended \hardened{\cpfivezerofive{}} and the undefended \cpfivezerofive{}, but adding the defense to victims in fact adds a vulnerability that undefended victims do not have. Because defended victims are usually not allowed to pass, they blunder seki situations where it is better to pass than play.

\begin{figure}
    \centering
    \includesvg[width=0.48\textwidth]{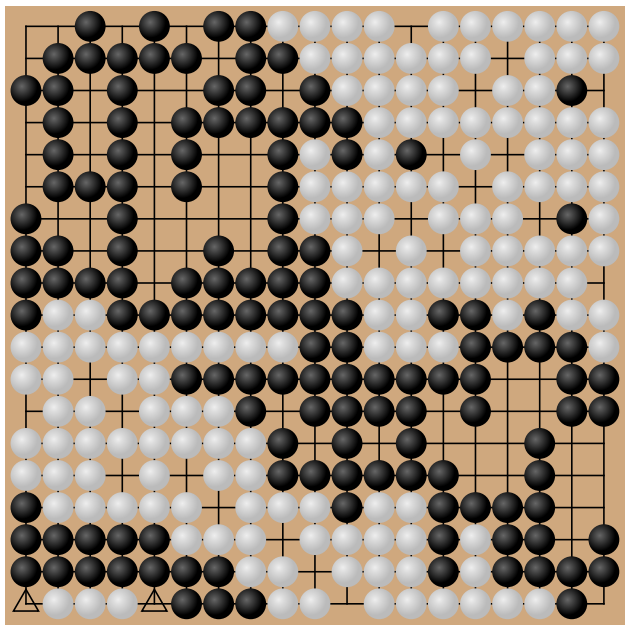}
	\caption{Black moves next in this game. There is a seki in the bottom left corner of the board. Neither black nor white should play in either square marked with $\Delta$, or else the other player will play in the other square and capture the opponent's stones. If \cpfivezerofive{} with 128 visits plays as black, it will pass. On the other hand, \hardened{\cpfivezerofive{}} with 128 visits playing as black will play in one of the marked squares and lose its stones.}
    \label{fig:app:cp505h-seki-blunder}
\end{figure}

For instance, consider the board shown in Figure~\ref{fig:app:cp505h-seki-blunder}. 
Black is next to play. 
At this point, the game is over unless one of the players severely blunders.
White cannot capture black's large group stretching from the top-left corner to the top-right corner, and black cannot capture white's two large groups.
There is a seki in the bottom-left corner of the board, where neither player wants to play in either of the two squares marked with $\Delta$ since then the other player could play in the other marked square and capture the opponent's stones.
Black is winning and should pass and wait for white to also pass or resign.
Indeed, \cpfivezerofive{} with 128 visits playing as black passes and eventually wins by 8.5 points.

\hardened{\cpfivezerofive{}}, however, is not allowed to pass, and instead plays in one of the squares marked by $\Delta$. White can then play in the other marked square to capture black's stones. Then white owns all the territory in the bottom-left corner and wins by 25.5 points.

We discovered this weakness of the pass-alive defense when we trained an adversary against \hardened{\cpfivezerofive{}} with the adversary's weights initialized to \cpsixtythree, an early KataGo checkpoint.
The adversary consistently set up similar seki situations to defeat \hardened{\cpfivezerofive{}}, but it would lose against the undefended \cpfivezerofive{}.

\clearpage
\section{\textSwitch{Hyperparameter settings}{Hyperparameter Settings}}
\label{app:hyperparameters}

We enumerate the key hyperparameters used in our training run in Table~\ref{tab:app:hyperparameters}.
For brevity, we omit hyperparameters that are the same as KataGo defaults and have only a minor effect on performance.

The key difference from standard KataGo training is that our adversarial policy uses a \texttt{b6c96} network architecture, consisting of 6 blocks and 96 channels.
By contrast, the victims we attack range from \texttt{b6c96} to \texttt{b40c256} in size.
We additionally disable a variety of game rule randomizations that help make KataGo a useful AI teacher in a variety of settings but are unimportant for our attack.
We also disable gatekeeping, designed to stabilize training performance, as our training has proved sufficiently stable without it.

We train at most $4$ times on each data row before blocking for fresh data.
This is comparable to the original KataGo training run, although the ratio during that run varied as the number of asynchronous self-play workers fluctuated over time.
We use an adversary visit count of $600$, which is comparable to KataGo, though the exact visit count has varied between their training runs.

In evaluation games we use a single search thread for KataGo unless otherwise specified.
We used 10 and 1024 search threads for evaluation of victims with $10^6$ and $10^7$ visits in order to ensure games complete in a reasonable time frame.
Holding visit count fixed, using more search threads tends to decrease the strength of an agent.
However increasing search threads enables more visits to be used in practice, ultimately enabling higher agent performance.

\begin{table}[h]
    \centering
    \begin{tabular}{lll}
         \toprule
         \textbf{Hyperparameter} & \textbf{Value} & \textbf{Different from KataGo?} \\
         \midrule
         Batch Size & 256 & Same \\
         Learning Rate Scale of Hard-coded Schedule & 1.0 & Same \\
         Minimum Rows Before Shuffling & 250,000 & Same \\
         Data Reuse Factor & 4 & Similar \\
         Adversary Visit Count & 600 & Similar \\
         Adversary Network Architecture & \texttt{b6c96} & Different \\
         Gatekeeping & Disabled & Different \\
         Auto-komi & Disabled & Different \\
         Komi randomization & Disabled & Different \\
         Handicap Games & Disabled & Different \\
         Game Forking & Disabled & Different \\
         Cheap Searches & Disabled & Different \\
         \bottomrule
    \end{tabular}
    \caption{Key hyperparameter settings for our adversarial training runs.}
    \label{tab:app:hyperparameters}
\end{table}

\subsection{\textSwitch{Configuration for curriculum against victim without search}{Configuration for Curriculum Against Victim Without Search}}
\label{app:hyperparameters:no-search}

In Section~\ref{sec:evaluation:no-search}, we train using a curriculum over checkpoints, moving on to the next checkpoint when the adversary's win rate exceeds 50\%. We ran the curriculum over the following checkpoints, all without search:
\begin{enumerate}
    \item Checkpoint 127: \texttt{b20c256x2-s5303129600-d1228401921} (\cponetwentyseven{}).
    \item Checkpoint 200: \texttt{b40c256-s5867950848-d1413392747}.
    \item Checkpoint 300: \texttt{b40c256-s7455877888-d1808582493}.
    \item Checkpoint 400: \texttt{b40c256-s9738904320-d2372933741}.
    \item Checkpoint 469: \texttt{b40c256-s11101799168-d2715431527}.
    \item Checkpoint 505: \texttt{b40c256-s11840935168-d2898845681} (\cpfivezerofive{}).
\end{enumerate}
These checkpoints can all be obtained from \citet{katagotraining:2022}.

We start with checkpoint 127 for computational efficiency: it is the strongest KataGo network of its size, 20 blocks or \texttt{b20}.
The subsequent checkpoints are all 40 block networks, and are approximately equally spaced in terms of training time steps.
We include checkpoint 469 in between 400 and 505 for historical reasons: we ran some earlier experiments against checkpoint 469, so it is helpful to include checkpoint 469 in the curriculum to check performance is comparable to prior experiments.

Checkpoint 505 is the latest \emph{confidently rated} network. 
There are some more recent, larger networks (\texttt{b60} = 60 blocks) that may have an improvement of up to 150 Elo.
However, they have had too few rated games to be confidently evaluated.

\subsection{\textSwitch{Configuration for curriculum against victim with passing defense}{Configuration for Curriculum Against Victim With Passing Defense}}
\label{app:hyperparameters:hardened-curriculum}

In Section~\ref{sec:evaluation:no-search-hardened}, we ran the curriculum over the following checkpoints, all with the pass-alive defense enabled:
\begin{enumerate}[leftmargin=40pt]
    \item Checkpoint 39: \texttt{b6c96-s45189632-d6589032} (\hardened{\cpthirtynine{}}), no search
    \item Checkpoint 49: \texttt{b6c96-s69427456-d10051148}, no search.
    \item Checkpoint 63: \texttt{b6c96-s175395328-d26788732}, no search.
    \item Checkpoint 79: \texttt{b10c128-s197428736-d67404019}, no search.
    \item Checkpoint 99: \texttt{b15c192-s497233664-d149638345}, no search.
    \item Checkpoint 127: \texttt{b20c256x2-s5303129600-d1228401921}, no search (\hardened{\cponetwentyseven{}}).
    \item Checkpoint 200: \texttt{b40c256-s5867950848-d1413392747}, no search
    \item Checkpoint 300: \texttt{b40c256-s7455877888-d1808582493}, no search.
    \item Checkpoint 400: \texttt{b40c256-s9738904320-d2372933741}, no search.
    \item Checkpoint 469: \texttt{b40c256-s11101799168-d2715431527}, no search.
    \item Checkpoint 505: \texttt{b40c256-s11840935168-d2898845681} (\hardened{\cpfivezerofive{}}), no search (1 visit).
    \item Checkpoint 505: \texttt{b40c256-s11840935168-d2898845681} (\hardened{\cpfivezerofive{}}), 2 visits.
    \item Checkpoint 505: \texttt{b40c256-s11840935168-d2898845681} (\hardened{\cpfivezerofive{}}), 4 visits.
    \item Checkpoint 505: \texttt{b40c256-s11840935168-d2898845681} (\hardened{\cpfivezerofive{}}), 8 visits.
    \item Checkpoint 505: \texttt{b40c256-s11840935168-d2898845681} (\hardened{\cpfivezerofive{}}), 16 visits.
    \item[16--20.] ...
    \item[21.] \texttt{b40c256-s11840935168-d2898845681} (\hardened{\cpfivezerofive{}}), 1024 visits.
    \item[22.] \texttt{b40c256-s11840935168-d2898845681} (\hardened{\cpfivezerofive{}}), 1600 visits.
    \item[23.] \texttt{b40c256-s11840935168-d2898845681} (\hardened{\cpfivezerofive{}}), 4096 visits.
    \item[24.] \texttt{b40c256-s11840935168-d2898845681} (\hardened{\cpfivezerofive{}}), 8192 visits.
    \item[25--27.] ...
    \item[28.] Checkpoint 505: \texttt{b40c256-s11840935168-d2898845681} (\hardened{\cpfivezerofive{}}), $2^{17} = 131072$ visits.
\end{enumerate}

We move on to the next checkpoint when the adversary's win rate exceeds 50\% until we reach \hardened{\cpfivezerofive{}} with 2 visits, at which point we increase the win rate threshold to 75\%. 

\clearpage
\section{\textSwitch{Compute estimates}{Compute Estimates}}
\label{app:compute-estimates}
In this section, we estimate the amount of compute that went into training our adversary and the amount of compute that went into training KataGo.

We estimate it takes \char`\~20.4 V100 GPU days to train our strongest pass-adversary, \char`\~2223.2 V100 GPU days to train our strongest cyclic-adversary, and at least 15,881 V100 GPU days to train the \cpfivezerofive{} KataGo checkpoint. Thus our pass-adversary and cyclic-adversary can be trained using 0.13\% and 14.0\% (respectively) of the compute it took to train KataGo. Moreover, an earlier checkpoint of the cyclic-adversary trained using only 7.6\% of the compute to train KataGo already achieves a 94\% win rate against \hardened{\cpfivezerofive{}} with 4096 visits.

As another point of reference, our strongest pass-adversary took $9.18 \times 10^{4}$ self-play games to train, our strongest cyclic-adversary took $1.01 \times 10^6$ self-play games to train, and \cpfivezerofive{} took $5.66 \times 10^7$ self-play games to train.\footnote{
To estimate the number of games for KataGo, we count the number of training games at \href{https://katagotraining.org/games/}{\texttt{katagotraining.org/games}} (only for networks prior to \cpfivezerofive{}) and \href{https://katagoarchive.org/g170/selfplay/index.html}{\texttt{katagoarchive.org/g170/selfplay/index.html}}.
}

Note that training our cyclic-adversary used 14\% of \cpfivezerofive{}'s compute, but less than 2\% of \cpfivezerofive{}'s games. This is because our cyclic-adversary was trained against high-visit count versions of \cpfivezerofive{} towards the end of its curriculum, and the compute required to generate a victim-play game scales proportionally with the amount of victim visits. See Figure~\ref{fig:compute:training-curves} for a visual illustration of this effect.

\subsection{\textSwitch{Estimating the compute used by our attack}{Estimating the Compute Used by Our Attack}}
\label{app:compute-estimates:our-attack}

To train our adversaries, we used A4000, A6000, A100 40GB, and A100 80GB GPUs. The primary cost of training is in generating victim-play games, so we estimated GPU-day conversions between these GPUs by benchmarking how fast the GPUs generated games.

We estimate that one A4000 GPU-day is 0.627 A6000 GPU-days, one A100 40GB GPU-day is 1.669 A6000 GPU-days, and one A100 80GB GPU-day is 1.873 A6000 GPU-days.
We estimate one A6000 GPU-day is 1.704 V100 GPU-days.

\begin{figure*}
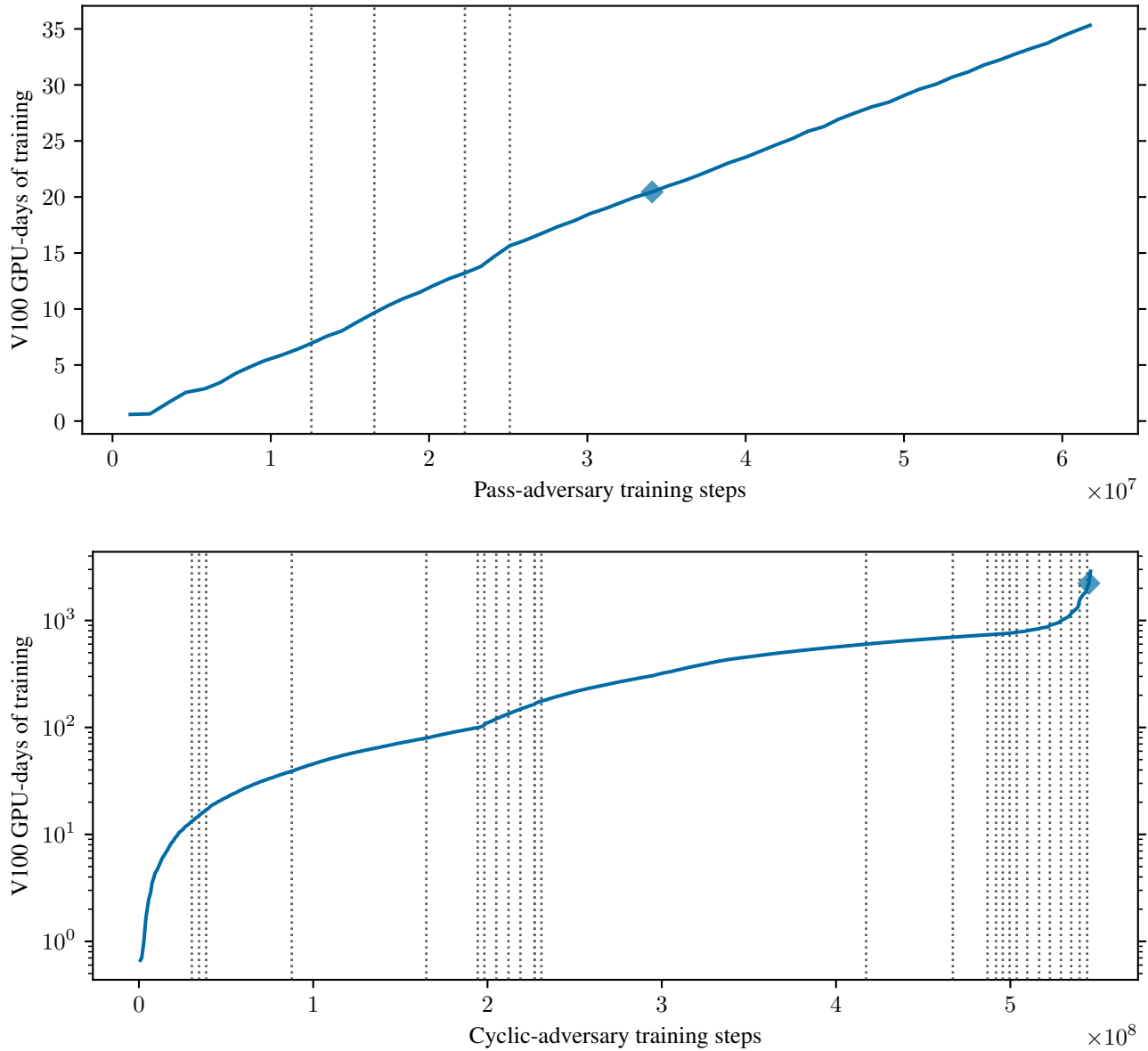

    \centering
    \input{figs/katagovisualizer/gpu-days-vs-steps-pass.pgf} \\
    \vspace{5mm}
    \input{figs/katagovisualizer/gpu-days-vs-steps-cyclic.pgf}
    \caption{
    The compute used for adversary training ($y$-axis) as a function of the number of adversary training steps taken ($x$-axis).
    The plots here mirror the structure of
    Figure~\ref{fig:evaluation:training-curve-no-search} and Figure~\ref{fig:evaluation:training-curve-cyclic}.
    Top: The compute of the pass-adversary is a linear function of its training steps because the pass-adversary was trained against victims of similar size, all of which used no search (Appendix~\ref{app:hyperparameters:no-search}). 
    Bottom: In contrast, the compute of the cyclic-adversary is highly non-linear due to training against a wider range of victim sizes and the exponential ramp up of victim search at the end of its curriculum (Appendix~\ref{app:hyperparameters:hardened-curriculum}).
    }
    \label{fig:compute:training-curves}
\end{figure*}

Figure~\ref{fig:compute:training-curves} plots the amount of compute used against the number of adversary training steps. To train the pass-adversary, we used 12.001 A6000 GPU-days, converting to 20.4 V100 GPU-days. To train the cyclic-adversary, we used 61.841 A4000 GPU-days, 348.582 A6000 GPU-days, 299.651 A100 40GB GPU-days, and 222.872 A100 80GB GPU-days, converting to 2223.2 V100 GPU-days.

\begin{figure*}
    \centering
    \input{figs/katagovisualizer/win-rate-vs-gpu-days-pass.pgf}
    \caption{The win rate achieved by the pass-adversary throughout training ($y$-axis) as a function of the training compute used ($x$-axis).
    This figure is the same as Figure~\ref{fig:evaluation:training-curve-no-search} but with V100 GPU-days on the $x$-axis instead of adversary training steps.
    }
    \label{fig:compute:win-rate-vs-gpu-days-pass}
\end{figure*}
\begin{figure*}
    \centering
    \input{figs/katagovisualizer/win-rate-vs-gpu-days-cyclic.pgf}
    \caption{
    The win rate achieved by the cyclic-adversary throughout training ($y$-axis) as a function of the training compute used ($x$-axis). 
    This figure is the same as Figure~\ref{fig:evaluation:training-curve-cyclic} but with V100 GPU-days on the $x$-axis instead of adversary training steps.
    }
    \label{fig:compute:win-rate-vs-gpu-days-cyclic}
\end{figure*}

The cyclic-adversary was already achieving high win rates against \hardened{\cpfivezerofive{}} with smaller amounts of training.
In Figure~\ref{fig:compute:win-rate-vs-gpu-days-cyclic}, earlier checkpoints of the cyclic-adversary achieved a win rate of 64.6\% against \hardened{\cpfivezerofive{}} with 4096 victim visits using 749.6 V100 GPU-days of training (4.7\% of the compute to train \cpfivezerofive{})
and a win rate of 94\% using 1206.2 V100 GPU-days of training (7.6\% of the compute to train \cpfivezerofive{}),
compared to a win rate of 95.7\% using 2223.2 V100 GPU-days of training.

\pagebreak

\subsection{\textSwitch{Estimating the compute used to train the \cpfivezerofive{} KataGo checkpoint}{Estimating the Compute Used to Train the \cpfivezerofive{} KataGo Checkpoint}}
\label{app:compute-estimates:katago}
The \cpfivezerofive{} KataGo checkpoint was obtained via distributed (i.e. crowdsourced) training starting from the strongest checkpoints in KataGo's ``third major run'' \citep{trainhistory2021}. The KataGo repository \href{https://github.com/lightvector/KataGo/blob/master/TrainingHistory.md\#third-major-run}{documents the compute used to train the strongest network of this run as}: 14 days of training with 28 V100 GPUs, 24 days of training with 36 V100 GPUs, and 119 days of training with 46 V100 GPUs. This totals to $14 \times 28 + 24 \times 36 + 119 \times 46 = 6730$ V100 GPU days of compute.

To lower-bound the remaining compute used by distributed training, we make the assumption that the average row of training-data generated during distributed training was more expensive to generate than the average row of data for the ``third major run''. We justify this assumption based on the following factors:\footnote{The biggest potential confounding factor is KataGo's neural network cache, which (per David Wu in private comms) ``is used if on a future turn you visit the same node that you already searched on the previous turn, or if multiple move sequences in a search lead to the same position''. Moreover, ``this [cache] typically saves somewhere between 20\% and 50\% of the cost of a search relative to a naive estimate based on the number of visits''. It is possible that distributed training has a significantly higher cache hit-rate than the ``third major run'', in which case our bound might be invalid. We assume that the stated factors are enough to overcome this and other potential confounding effects to yield a valid lower-bound.}
\begin{enumerate}
    \item The ``third major run'' used \texttt{b6}, \texttt{b10}, \texttt{b20}, \texttt{b30}, and \texttt{b40} nets while distributed training used only \texttt{b40} nets and larger, with larger nets being more costly to run (Table~\ref{tab:app:compute-by-network}).

    \item The ``third major run'' used less search during self-play than distributed training. Source: the following message from David Wu (the creator and primary developer of KataGo).
    \begin{quote}
        KataGo used 600 full / 100 cheap [visits] for roughly the first 1-2 days of training (roughly up through b10c128 and maybe between 1/2 and 1/4 of b15c192), 1000 full / 200 cheap [visits] for the rest of g170 (i.e. all the kata1 models that were imported from the former run g170 that was done on private hardware alone, before that run became the prefix for the current distributed run kata1), and then 1500 full / 250 cheap [visits] for all of distributed training so far. 
    \end{quote}
\end{enumerate}
\cpfivezerofive{} was trained with 2,898,845,681 data rows, while the strongest network of the ``third major run'' used 1,229,425,124 data rows. We thus lower bound the compute cost of training \cpfivezerofive{} at $2898845681 / 1229425124 \times 6730 \approx 15881$ V100 GPU days.
\begin{table}
    \centering
    \begin{tabular}{ll}
         \toprule
         \textbf{Network} & \textbf{FLOPs / forward pass} \\
         \midrule
         \texttt{b6c96} & $7.00 \times 10^8$ \\
         \texttt{b10c128} & $2.11 \times 10^9$ \\
         \texttt{b15c192} & $7.07 \times 10^9$ \\
         \texttt{b20c256} & $1.68 \times 10^{10}$ \\
         \texttt{b40c256} & $3.34 \times 10^{10}$ \\
         \bottomrule
    \end{tabular}
    \caption{Inference compute costs for different KataGo neural network architectures. These costs were empirically measured using \href{https://pypi.org/project/ptflops/}{\texttt{ptflops}} and \href{https://pypi.org/project/thop/}{\texttt{thop}}, and the reported numbers are averaged over the two libraries.}
    \label{tab:app:compute-by-network}
\end{table}

\clearpage
\section{\textSwitch{Strength of Go AI systems}{Strength of Go AI Systems}}
\label{app:experiments:strength}

In this section, we estimate the strength of KataGo's \cpfivezerofive{} network with and without search and the AlphaZero agent from \citet{schmid2021} playing with 800 visits.

\subsection{\textSwitch{Strength of KataGo without search}{Strength of KataGo Without Search}}
\label{app:experiments:strength:katago-no-search}
First, we estimate the strength of KataGo's \cpfivezerofive{} agent playing without search.
We use two independent methodologies and conclude that \cpfivezerofive{} without search is at the level of a weak professional.

One way to gauge the performance of \cpfivezerofive{} without search is to see how it fares against humans on online Go platforms. Per Table~\ref{tab:app:kgs-ranks}, on the online Go platform KGS, a slightly earlier (and weaker) checkpoint than \cpfivezerofive{} playing without search is roughly at the level of a top-100 European player. However, some caution is needed in relying on KGS rankings:
\begin{enumerate}
    \item Players on KGS compete under less focused conditions than in a tournament, so they may underperform.

    \item KGS is a less serious setting than official tournaments, which makes cheating (e.g., using an AI) more likely. Thus human ratings may be inflated.

    \item
    Humans can play bots multiple times and adjust their strategies, while bots remain static.
    In a sense, humans are able to run adversarial attacks on the bots, and are even able to do so in a white-box manner since the source code and network weights of a bot like KataGo are public.
\end{enumerate}

\begin{table}[h]
    \centering
    \begin{tabular}{lllll}
         \toprule
         \textbf{KGS handle} & \textbf{Is KataGo?} & \textbf{KGS rank} & \textbf{EGF rank} & \textbf{EGD Profile} \\
         \midrule
         Fredda &  & 22 & 25 & \href{https://www.europeangodatabase.eu/EGD/Player_Card.php?&key=13937946}{Fredrik Blomback}\\
         cheater &  & 25 & 6 & \href{https://europeangodatabase.eu/EGD/Player_Card.php?&key=12686597}{Pavol Lisy}\\
         TeacherD &  & 26 & 39 & \href{https://europeangodatabase.eu/EGD/Player_Card.php?&key=14225926}{Dominik Boviz}\\
         NeuralZ03 & \checkmark & 31 & & \\
         NeuralZ05 & \checkmark & 32 & & \\
         NeuralZ06 & \checkmark & 35 & & \\
         ben0 & & 39 & 16 & \href{https://europeangodatabase.eu/EGD/Player_Card.php?&key=14513532}{Benjamin Drean-Guenaizia} \\
         sai1732 & & 40 & 78 & \href{https://europeangodatabase.eu/EGD/Player_Card.php?&key=18486897}{Alexandr Muromcev} \\
         Tichu & & 49 & 64 & \href{https://europeangodatabase.eu/EGD/Player_Card.php?&key=15933973}{Matias Pankoke} \\
         Lukan & & 53 & 10 & \href{https://europeangodatabase.eu/EGD/Player_Card.php?&key=13201914}{Lukas Podpera} \\
         HappyLook & & 54 & 49 & \href{https://europeangodatabase.eu/EGD/Player_Card.php?&key=16049671}{Igor Burnaevskij} \\
         \bottomrule
    \end{tabular}
    \caption{
        Rankings of various humans and no-search KataGo bots on KGS~\citep{kgs2022}.
        Human players were selected to be those who have European Go Database (EGD) profiles~\citep{egd}, from which we obtained the European Go Federation (EGF) rankings in the table.
        The KataGo bots are running with a checkpoint slightly weaker than \cpfivezerofive{}, specifically Checkpoint 469 or \texttt{b40c256-s11101799168-d2715431527}~\citep{neuralz2022}. Per \citet{katagotraining:2022}, the checkpoint is roughly 10 Elo weaker than \cpfivezerofive{}.
    }
    \label{tab:app:kgs-ranks}
\end{table}

Another way to estimate the strength of \cpfivezerofive{} without search is to compare it to other AIs with known strengths and extrapolate performance across different amounts of search. Our analysis critically assumes the transitivity of Elo at high levels of play. We walk through our estimation procedure below:
\begin{enumerate}
    \item
    Our anchor is ELF OpenGo at 80,000 visits per move using its ``prototype'' model, which won all 20 games played against four top-30 professional players, including five games against the now world number one~\citep{tian2019}.
    We assume that ELF OpenGo at 80,000 visits is strongly superhuman,
    meaning it has a 90\%+ win rate over the strongest current human.\footnote{
    This assumption is not entirely justified by statistics, as a 20:0 record
    only yields a 95\% binomial lower confidence bound of an 83.16\% win rate
    against top-30 professional players in 2019.
    It does help however that the players in question were rated \#3, \#5, \#23, and \#30 in the world at the time.
    }
    At the time of writing, the top ranked player on Earth has an Elo of 3845 on goratings.org~\citep{goratings:2022}.
    Under our assumption, ELF OpenGo at 80,000 visits per move would have an Elo of 4245+ on goratings.org.

    \item ELF OpenGo's ``final'' model is about 150 Elo stronger than its prototype model~\citep{tian2019}, giving an Elo of 4395+ at 80,000 visits per move.

    \item
    The strongest network in the original KataGo paper was shown to be slightly stronger than ELF OpenGo's final network~\citep[Table~1]{wu2019} when both bots were run at 1600 visits per move. From Figure~\ref{fig:evaluation:elo-by-search}, we see that the relative strengths of KataGo networks is maintained across different amounts of search. We thus extrapolate that the strongest network in the original KataGo paper with 80,000 visits would also have an Elo of 4395+ on goratings.org.

    \item
    The strongest network in the original KataGo paper is comparable to the \texttt{b15c192-s1503689216-d402723070} checkpoint on katagotraining.org~\citep{katagotraining:2022}. We dub this checkpoint \cponezerothree{}.
    In a series of benchmark games, we found that \cpfivezerofive{} without search won 27/3200 games against \cponezerothree{} with 1600 visits. This puts \cponezerothree{} with 1600 visits \char`\~823 Elo points ahead of \cpfivezerofive{} without search.

    \item
    Finally, log-linearly extrapolating the performance of \cponezerothree{} from 1600 to 80,000 visits using Figure~\ref{fig:evaluation:elo-by-search} yields an Elo difference of \char`\~834 between the two visit counts.

    \item
    Combining our work, we get that \cpfivezerofive{} without search is roughly $823 + 834$ = \char`\~1657 Elo points weaker than ELF OpenGo with 80,000 visits. This would give \cpfivezerofive{} without search an Elo rating of $4395 - 1657$ = \char`\~2738 on goratings.org, putting it at the skill level of a weak professional.
\end{enumerate}

As a final sanity check on these calculations, the raw AlphaGo Zero neural network was reported to have an Elo rating of 3,055, comparable to AlphaGo Fan's 3,144 Elo.\footnote{The Elo scale used in \citet{silver2017} is not directly comparable to our Elo scale, although they should be broadly similar as both are anchored to human players.}
Since AlphaGo Fan beat Fan Hui, a 2-dan professional player~\citep{silver2017}, this confirms that well-trained neural networks can play at the level of human professionals.
Although there has been no direct comparison between KataGo and AlphaGo Zero, we would expect them to be not wildly dissimilar.
Indeed, if anything the latest versions of KataGo are likely stronger, benefiting from both a large distributed training run (amounting to over 10,000 V100 GPU days of training) and four years of algorithmic progress.

\clearpage
\subsection{\textSwitch{Strength of KataGo with search}{Strength of KataGo With Search}}
\label{app:experiments:strength:katago-search}
In the previous section, we established that \cpfivezerofive{} without search is at the level of a weak professional with rating around \char`\~2738 on goratings.org.

\begin{figure}
    \centering
    \input{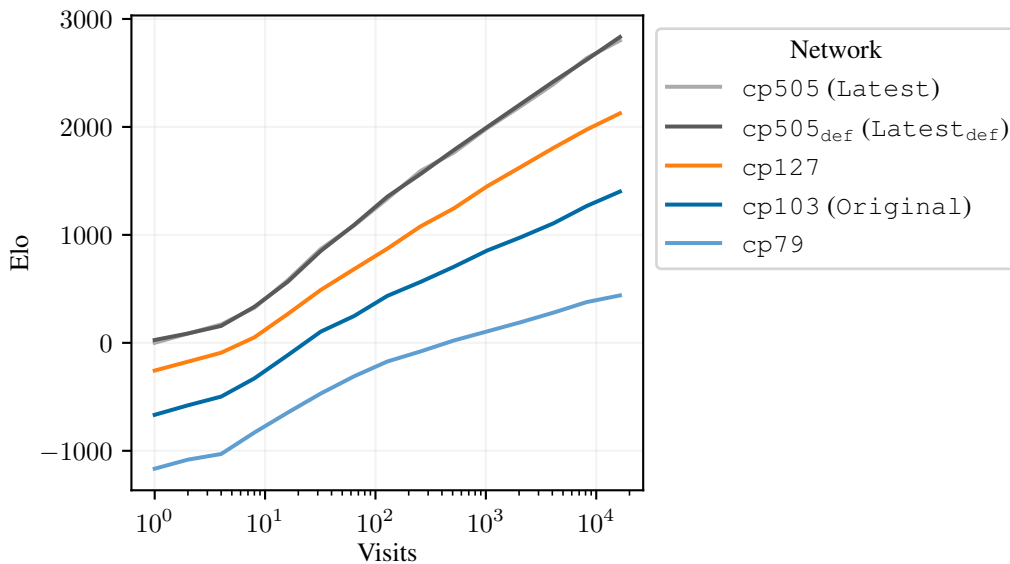}
    \caption{Elo ranking ($y$-axis) of networks (different colored lines) by visit count ($x$-axis). The lines are approximately linear on a log $x$-scale, with the different networks producing similarly shaped lines vertically shifted. This indicates that there is a \emph{consistent} increase in Elo, regardless of network strength, that is logarithmic in visit count. Elo ratings were computed from self-play games among the networks using a Bayesian Elo estimation algorithm~\citep{eloestimationscript2022}.}
    \label{fig:evaluation:elo-by-search}
\end{figure}

Assuming Elo transitivity,
we can estimate the strength of \cpfivezerofive{}
by utilizing Figure~\ref{fig:evaluation:elo-by-search}.
Our evaluation results tell us that
\cpfivezerofive{} with 8 playouts/move is roughly 325 Elo stronger
than \cpfivezerofive{} with no search.
This puts \cpfivezerofive{} with 8 playouts/move at an Elo of 
\char`\~3063 on goratings.org---within the top 500 in the world.
Beyond 128 playouts/move, \cpfivezerofive{} plays at a superhuman level.
\cpfivezerofive{} with 512 playouts/move, for instance, is roughly 1762 Elo stronger than \cpfivezerofive{} with no search, giving an Elo of 4500, over 600 points higher than the top player on goratings.org.

\subsection{Strength of AlphaZero}
\label{app:experiments:strength:alphazero}

Prior work from \citet{timbers2022} described in Section~\ref{sec:related_work} exploited the AlphaZero replica from \citet{schmid2021} playing with 800 visits.
Unfortunately, this agent has never been evaluated against KataGo or against any human player, making it difficult to directly compare its strength to those of our victims.
Moreover, since it is a proprietary model, we cannot perform this evaluation ourselves.
Accordingly, in this section we seek to estimate the strength of these AlphaZero agents using three anchors: GnuGo, Pachi and Lee Sedol.
Our estimates suggest AlphaZero with 800 visits ranges in strength from the top 300 of human players, to being slightly superhuman.

\begin{table}
    \centering
    \begin{tabular}{llll}
        \toprule
        Agent & Victim? & Elo (rel GnuGo) & Elo (rel victim)  \\
        \midrule
        AlphaZero(s=16k, t=800k) & & +3139 & +1040 \\
        AG0 3-day(s=16k) & & +3069 & +970 \\
        AlphaGo Lee(time=1sec) & & +2308 & +209 \\
        \textbf{AlphaZero(s=800,t=800k)} & \checkmark & \textbf{+2099} & 0 \\
        Pachi(s=100k) & & +869 & -1230 \\
        Pachi(s=10k) & & +231 & -1868 \\
        GnuGo(l=10) & & +0 & -2099 \\
        \bottomrule
    \end{tabular}
    \caption{Relative Elo ratings for AlphaZero, drawing on information from \citet[Table~4]{schmid2021}, \citet{silver2018} and \citet{silver2017}. s stands for number of steps, time for thinking time, and t for number of training steps.}
    \label{tab:app:alphazero-relative-elo}
\end{table}

We reproduce relevant Elo comparisons from prior work in Table~\ref{tab:app:alphazero-relative-elo}.
In particular, Table~4 of \citet{schmid2021} compares the victim used in \citet{timbers2022},  AlphaZero(s=800,t=800k), to two open-source AI systems, GnuGo and Pachi.
It also compares it to a higher visit count version AlphaZero(s=16k, t=800k), from which we can compare using \citet{silver2018} to AG0 3-day and from there using \citet{silver2017} to AlphaGo Lee which played Lee Sedol.

Our first strength evaluation uses the open-source anchor point provided by Pachi(s=10k).
The authors of Pachi~\citep{baudis2011} report it achieves a 2-dan ranking on KGS~\citep{pachi2020} when playing with 5000 playouts and using up to 15,000 when needed.
We conservatively assume this corresponds to a 2-dan EGF player (KGS rankings tend to be slightly inflated compared to EGF), giving Pachi(s=10k) an EGF rating of 2200 GoR.\footnote{
GoR is a special rating system (distinct from Elo) used by the European Go Federation. The probability that a player $A$ with a GoR of $G_A$ beats a player $B$ with a GoR of $G_B$ is $1 / (1 + \left(\frac{3300 - G_A}{3300 - G_B}\right)^7)$.
}
The victim AlphaZero(s=800,t=800k) is 1868 Elo stronger than Pachi(s=10k), so assuming transitivity, AlphaZero(s=800,t=800k) would have an EGF rating of 3063 GoR.\footnote{
This is a slightly non-trivial calculation: we first calculated the win-probability $x$ implied by an 1868 Elo difference, and then calculated the GoR of AlphaZero(s=800,t=800k) as the value that would achieve a win-probability of $x$ against Pachi(s=10k) with 2200 GoR. We used the following notebook to perform this and subsquent Elo-GoR conversion calculations: \gorcolablink.
}
The top EGF professional Ilya Shiskin has an EGF rating of 2830 GoR~\citep{egfpros:2022} at the time of writing, and 2979 Elo on goratings.org~\citep{goratings:2022}.
Using Ilya as an anchor, this would give AlphaZero(s=800,t=800k) a rating of 3813 Elo on goratings.org.
This is near-superhuman, as the top player at the time of writing has an rating of 3845 Elo on goratings.org.

However, some caution is needed here---the Elo gap between Pachi(s=10k) and AlphaZero(s=800,t=800k) is huge, making the exact value unreliable.
The gap from Pachi(s=100k) is smaller, however unfortunately to the best of our knowledge there is no public evaluation of Pachi at this strength.
However, the results in \citet{pachi2020} strongly suggest it would perform at no more than a 4-dan KGS level, or at most a 2400 GoR rating on EGF.\footnote{In particular, \citet{pachi2020} report that Pachi achieves a 3-dan to 4-dan ranking on KGS when playing on a cluster of 64 machines with 22 threads, compared to 2-dan on a 6-core Intel i7.
Figure~4 of \citet{baudis2011} confirms playouts are proportional to the number of machines and number of threads, and we'd therefore expect the cluster to have 200x as many visits, or around a million visits.
If 1 million visits is at best 4-dan, then 100,000 visits should be weaker.
However, there is a confounder: the 1 million visits was distributed across 64 machines, and Figure~4 shows that distributed playouts do worse than playouts on a single machine.
Nonetheless, we would not expect this difference to make up for a 10x difference in visits.
Indeed, \citet[Figure~4]{baudis2011} shows that 1 million playouts spread across 4 machines (red circle) is substantially better than 125,000 visits on a single machine (black circle), achieving an Elo of around 150 compared to -20.}
Repeating the analysis above then gives AlphaZero(s=800,t=800k) a rating of 2973 GoR on EGF and a rating of 3419 Elo on goratings.org.
This is a step below superhuman level, and is roughly at the level of a top-100 player in the world.

If we instead take GnuGo level 10 as our anchor, we get a quite different result.
It is known to play between 10 and 11kyu on KGS~\citep{gnugo2:2022}, or at an EGF rating of 1050 GoR.
This gives AlphaZero(s=800,t=800k) an EGF rating of 2900 GoR, or a goratings.org rating of 3174 Elo.
This is still strong, in the top \char`\~300 of world players, but is far from superhuman.

The large discrepancy between these results led us to seek a third anchor point: how AlphaZero performed relative to previous AlphaGo models that played against humans.
A complication is that the version of AlphaZero that \citeauthor{timbers2022} use differs from that originally reported in \citet{silver2018}, however based on private communication with \citeauthor{timbers2022} we are confident the performance is comparable:
\begin{quotation}
These agents were trained identically to the original AlphaZero paper, and were trained for the full 800k steps. We actually used the original code, and did a lot of validation work with Julian Schrittweiser \& Thomas Hubert (two of the authors of the original AlphaZero paper, and authors of the ABR paper) to verify that the reproduction was exact. We ran internal strength comparisons that match the original training runs.
\end{quotation}

Table~1 of \citet{silver2018} shows that AlphaZero is slightly stronger than AG0 3-day (AlphaGo Zero, after 3 days of training), winning 60 out of 100 games giving an Elo difference of +70.
This tournament evaluation was conducted with both agents having a thinking time of 1 minute.
Table~S4 from \citet{silver2018} reports that 16k visits are performed per second, so the tournament evaluation used a massive 960k visits--significantly more than reported on in Table~\ref{tab:app:alphazero-relative-elo}.
However, from Figure~\ref{fig:evaluation:elo-by-search} we would expect the \emph{relative} Elo to be comparable between the two systems at different visit counts,
so we extrapolate AG0 3-day at 16k visits as being an Elo of $3139-70=3069$ relative to GnuGo.

Figure~3a from \citet{silver2017} report that AG0 3-day achieves an Elo of around 4500.
This compares to an Elo of 3,739 for AlphaGo Lee.
To the best of our knowledge, the number of visits achieved per second of AlphaGo Lee has not been reported.
However, we know that AG0 3-day and AlphaGo Lee were given the same amount of thinking time, so we can infer that AlphaGo Lee has an Elo of $-761$ relative to AG0 3-day.
Consequently, AlphaGo Lee(time=1sec) thinking for 1 second has an Elo relative to GnuGo of $3069-761=2308$.

Finally, we know that AlphaGo Lee beat Lee Sedol in four out of five matches, giving AlphaGo Lee a +240 Elo difference relative to Lee Sedol, and that Lee Sedol has an Elo of 2068 relative to GnuGo level 10.
This would imply that the victim is slightly stronger than Lee Sedol.
However, this result should be taken with some caution.
First, it relies on transitivity through many different versions of AlphaGo.
Second, the match between AlphaGo Lee and Lee Sedol was played under two hours of thinking time with 3 byoyomi periods of 60 seconds per move~\citet[page~30]{silver2018}.
We are extrapolating from this to some hypothetical match between AlphaGo Lee and Lee Sedol with only 1 second of thinking time per player.
Although the Elo rating of Go AI systems seems to improve log-linearly with thinking time, it is unlikely this result holds for humans.

\clearpage
\section{\textSwitch{More evaluations of adversaries against KataGo}{More Evaluations of Adversaries Against KataGo}}
\label{app:unhardened-results}

In this section we provide more evaluations of our attacks from Section~\ref{sec:evaluation}.

\subsection{\textSwitch{Evolution of pass-adversary over training}{Evolution of Pass-Adversary Over Training}}

\begin{figure}[h]
    \vspace{1cm}
    \centering
    \input{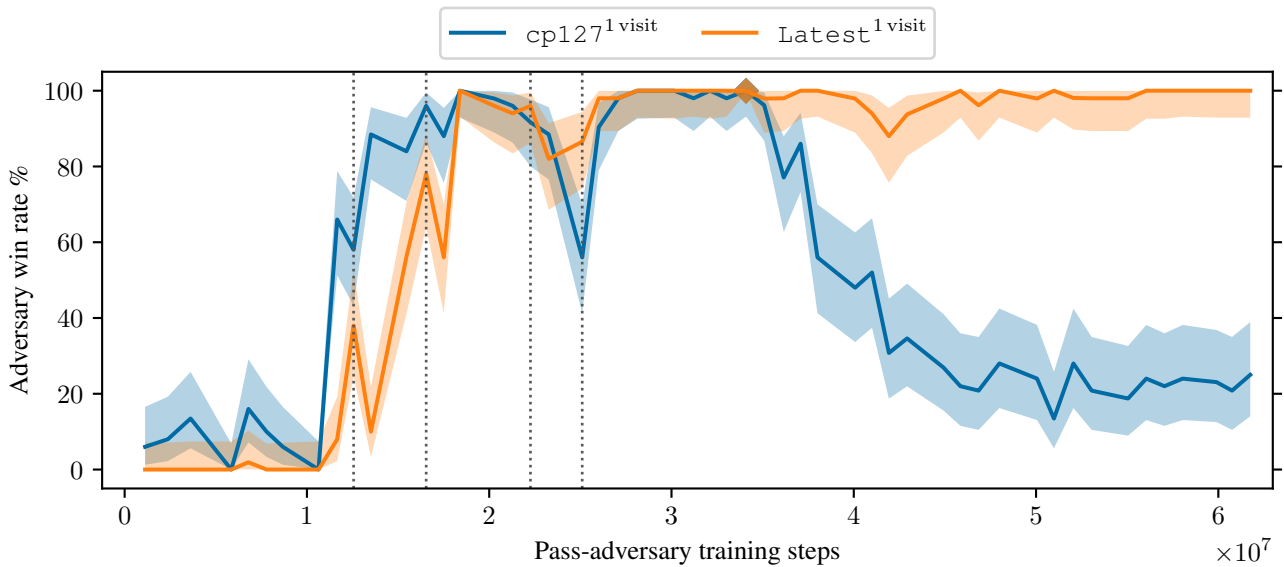}
    \caption{The win rate ($y$-axis) of the pass-adversary from Section~\ref{sec:evaluation:no-search} over time ($x$-axis) against the \cponetwentyseven{} and \cpfivezerofive{} victim policy networks playing without search. The strongest adversary checkpoint (marked $\blacklozenge$) wins 1047/1048 games against $\cpfivezerofive{}$. The adversary overfits to \cpfivezerofive{}, winning less often against \cponetwentyseven{} over time. Shaded interval is a 95\% Clopper-Pearson interval over $n=50$ games per checkpoint. The adversarial policy is trained with a curriculum, starting from \cponetwentyseven{} and ending at \cpfivezerofive{} (see Appendix~\ref{app:hyperparameters:no-search}). Vertical dashed lines denote switches to a later victim policy.}
    \label{fig:evaluation:training-curve-no-search}
\end{figure}

In Figure~\ref{fig:evaluation:training-curve-no-search} we evaluate the pass-adversary from Section~\ref{sec:evaluation:no-search} against \cponetwentyseven{} and \cpfivezerofive{} throughout the training process of the adversary.
We find the pass-adversary attains a large (>$90\%$) win rate against both victims throughout much of training.
However, over time the adversary overfits to \cpfivezerofive{}, with the win rate against \cponetwentyseven{} falling to around 22\%.

\begin{figure}
    \centering
    \begin{subfigure}[b]{\textwidth}
        \centering
        \input{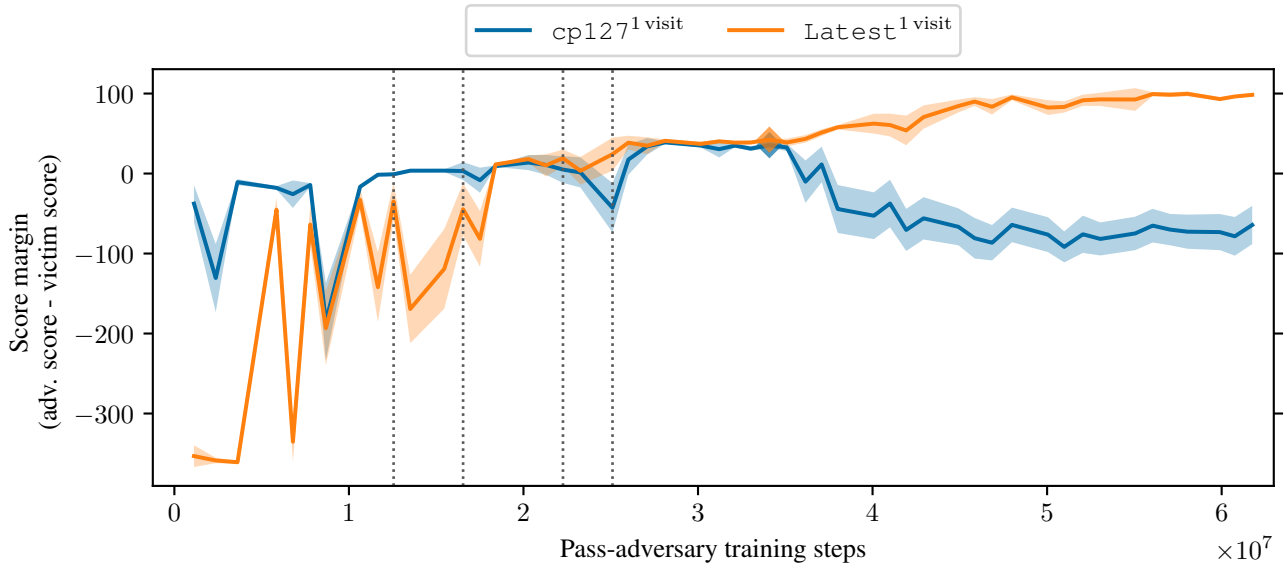}
        \caption{%
        Final score margin from adversary's perspective (i.e. adversary score $-$ victim score) on $y$-axis vs. adversary training steps on $x$-axis.
        }
        \label{fig:evaluation:training-curve-no-search-scorediff-overall}
    \end{subfigure}
    \\[0.5cm]
    \begin{subfigure}[b]{0.48\textwidth}
        \centering
        \resizebox{\textwidth}{!}{\input{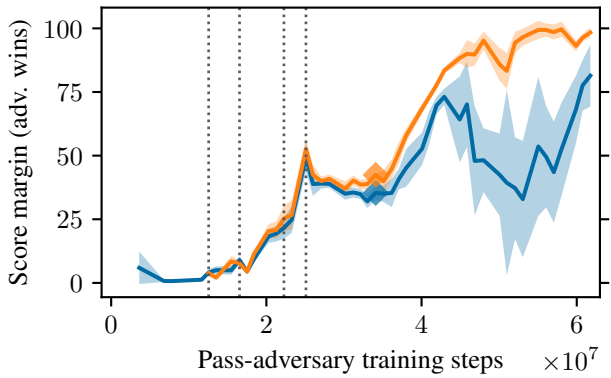}}
        \caption{Score margin, restricted to games adversary won.}
        \label{fig:evaluation:training-curve-no-search-scorediff-advwin}
    \end{subfigure}
    \hfill
    \begin{subfigure}[b]{0.48\textwidth}
        \centering
        \resizebox{\textwidth}{!}{\input{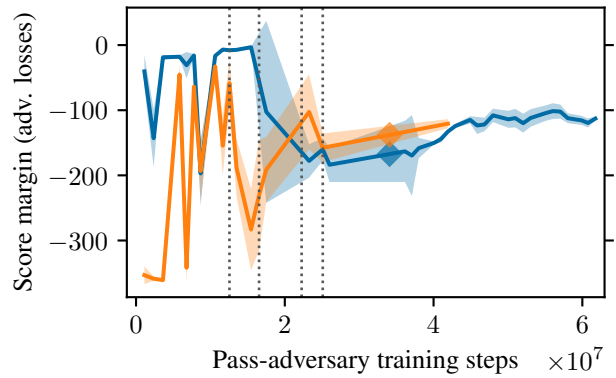}}
        \caption{Score margin, restricted to games adversary lost.}
        \label{fig:evaluation:training-curve-no-search-scorediff-advlose}
    \end{subfigure}
    \caption{We evaluate the average margin of victory for the pass-adversary from Section~\ref{sec:evaluation:no-search} against \cpfivezerofive{} without search as the training process progresses. Shaded regions are 95\% T-intervals over $n=50$ games per checkpoint.}
    \label{fig:evaluation:training-curve-no-search-scorediff}
\end{figure}

In Figure~\ref{fig:evaluation:training-curve-no-search-scorediff}, the context is the same as the preceding figure but instead of win rate we report the margin of victory. In the win-only and loss-only subfigures, we plot only points with at least 5 wins or losses. Note that standard Go has no incentives for winning by a larger margin; we examine these numbers for solely additional insight into the training process of our adversary. We see that even after win rate is near 100\% against \cpfivezerofive{}, the win margin continues to increase, suggesting the adversary is still learning.

\clearpage
\subsection{\textSwitch{Score margin of the cyclic-adversary}{Score Margin of the Cyclic-Adversary}}

\begin{figure}[h]
    \centering
    \begin{subfigure}[b]{\textwidth}
        \centering
        \input{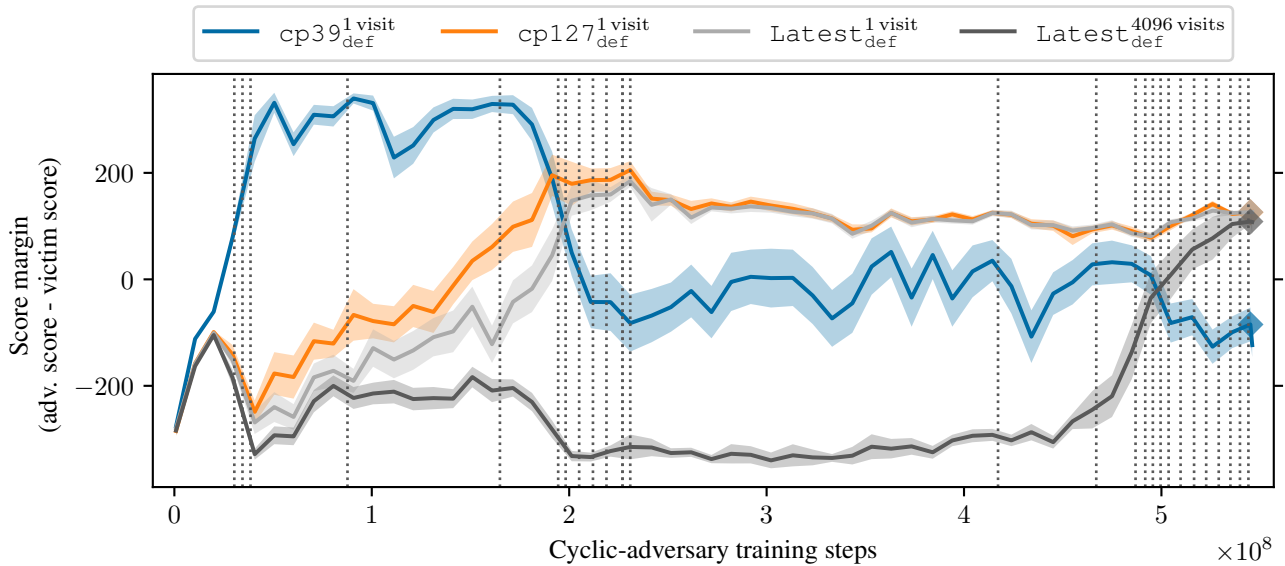}
        \caption{%
        Final score margin from adversary's perspective (i.e. adversary score $-$ victim score) on $y$-axis vs. adversary training steps on $x$-axis.
        }
        \label{fig:evaluation:545mil-training-curve-scorediff-overall}
    \end{subfigure}
    \\[0.5cm]
    \begin{subfigure}[b]{0.48\textwidth}
        \centering
        \resizebox{\textwidth}{!}{\input{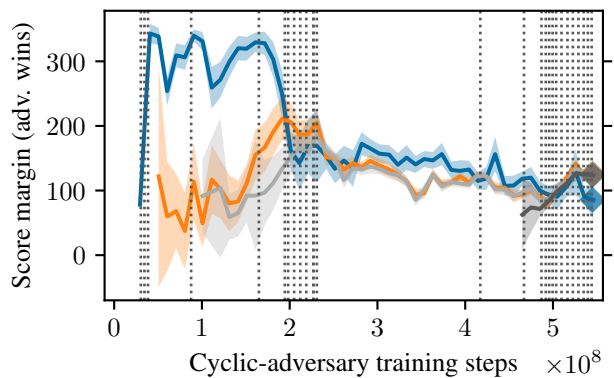}}
        \caption{Score margin, restricted to games adversary won.}
        \label{fig:evaluation:545mil-training-curve-scorediff-advwin}
    \end{subfigure}
    \hfill
    \begin{subfigure}[b]{0.48\textwidth}
        \centering
        \resizebox{\textwidth}{!}{\input{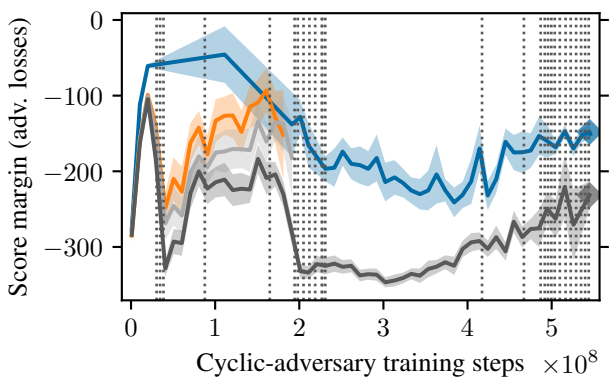}}
        \caption{Score margin, restricted to games adversary lost.}
        \label{fig:evaluation:545mil-training-curve-scorediff-advlose}
    \end{subfigure}
    \caption{We evaluate the average margin of victory for the cyclic-adversary from Section~\ref{sec:evaluation:no-search-hardened} against various victims as the training process progresses. Shaded regions are 95\% T-intervals over $n=50$ games per checkpoint.}
    \label{fig:evaluation:545mil-training-curve-scorediff}
\end{figure}

In Figure~\ref{fig:evaluation:545mil-training-curve-scorediff}, we show the margin of victory over the training process of the cyclic-adversary from Section~\ref{sec:evaluation:no-search-hardened} against victims with the pass-alive defense. The corresponding win rate is shown in Figure~\ref{fig:evaluation:training-curve-cyclic}. Compared to Figure~\ref{fig:evaluation:training-curve-no-search-scorediff}, we see that the margin of victory is typically larger. This is likely because the cyclic-adversary either captures a large group or gives up almost everything in a failed attempt. After approximately 250 million training steps, the margins are relatively stable, but we do see a gradual reduction in the loss margin against \hardened{\cpfivezerofive{}} with 4096 visits (preceding the eventual spike in win rate against that victim).

\clearpage
\subsection{\textSwitch{Pass-adversary vs. victims with search}{Pass-Adversary vs. Victims With Search}}

\begin{figure}[h]
    \vspace{1cm}
    \centering
    \begin{subfigure}[b]{0.48\textwidth}
        \centering
        \input{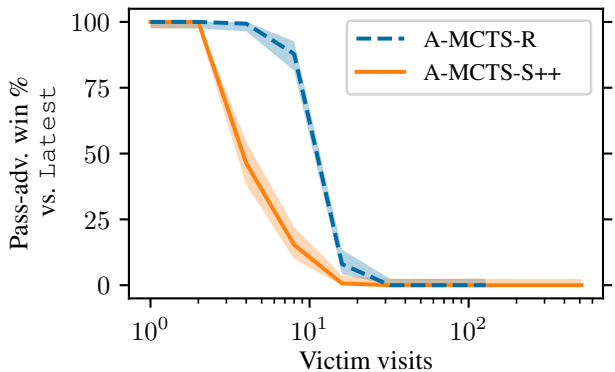}
        \caption{%
        Win rate by number of victim visits ($x$-axis) for \ourmctssampleabbrev{} and \ourmctsperfectabbrev{}.
        The adversary is run with 200 visits. The adversary is unable to exploit \cpfivezerofive{} when it plays with at least 32 visits.
        }
        \label{fig:evaluation:search:victim-visits}
    \end{subfigure}
    \hfill
    \begin{subfigure}[b]{0.48\textwidth}
        \centering
        \input{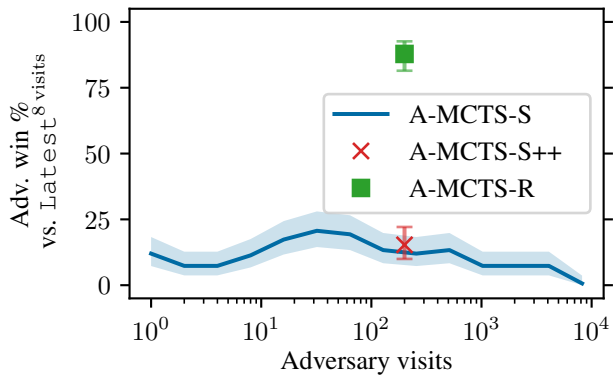}
        \caption{Win rate by number of adversary visits with \ourmctssampleabbrev{}, playing against \cpfivezerofive{} with 8 visits. In this case, scaling up the number of adversary visits does not lead to stronger attack.\\}
        \label{fig:evaluation:search:adv-visits}
    \end{subfigure}
    \caption{We evaluate the ability of the pass-adversary from Section~\ref{sec:evaluation:no-search} trained against \cpfivezerofive{} without search to transfer to \cpfivezerofive{} with search.}
    \label{fig:evaluation:search}
\end{figure}

We evaluate the ability of the pass-adversary to exploit \cpfivezerofive{} playing \emph{with} search (the pass-adversary was trained only against no-search victims).
Although the pass-adversary with \ourmctssampleabbrev{} and 200 visits achieves a win rate of 100\% over 160 games against \cpfivezerofive{} without search, in Figure~\ref{fig:evaluation:search:victim-visits} we find the win rate drops to 15.3\% at 8 victim visits.
However, \ourmctssampleabbrev{} models the victim as having no search at both training and inference time.
We also test \ourmctsperfectabbrev{}, which correctly models the victim at inference by performing an MCTS search at each victim-node in the adversary's tree.
We find that our pass-adversary with \ourmctsperfectabbrev{} performs somewhat better, obtaining an 87.8\% win rate against \cpfivezerofive{} with 8 visits, but performance drops to 8\% at 16 visits.

Of course, \ourmctsperfectabbrev{} is more computationally expensive than \ourmctssampleabbrev{}.
An alternative way to spend our inference-time compute budget is to perform \ourmctssampleabbrev{} with a greater \emph{adversary} visit count. We see in Figure~\ref{fig:evaluation:search:adv-visits}, however, that this does not increase the win rate of the pass-adversary against \cpfivezerofive{} with 8 visits. It seems that \cpfivezerofive{} at a modest number of visits quickly becomes resistant to our pass-adversary, no matter how we spend our inference-time compute budget.

\clearpage
\subsection{\textSwitch{Transferring attacks between checkpoints}{Transferring Attacks Between Checkpoints}}
\label{app:checkpoint-transfer}

\begin{figure}[h]
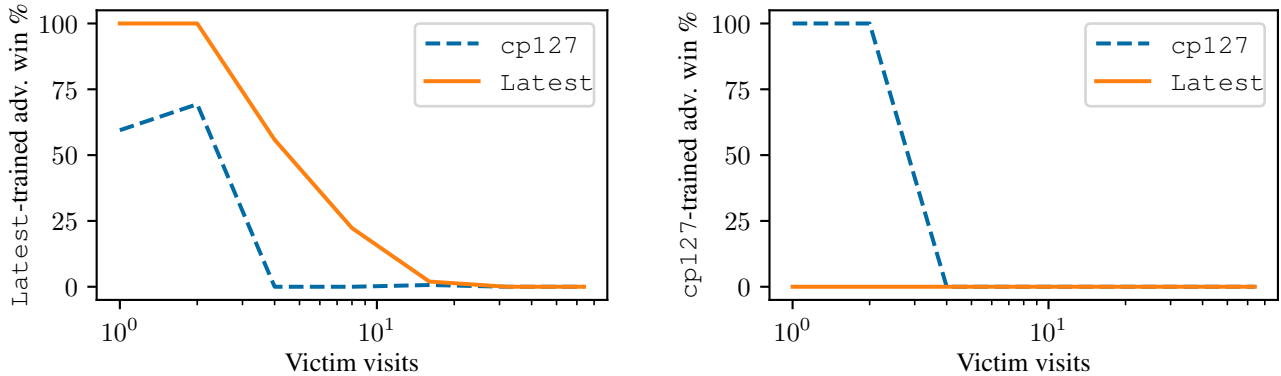

    \vspace{1cm}
    \centering
    \begin{subfigure}[b]{0.48\textwidth}
        \centering
        \input{figs/katagovisualizer/adv505-transfer.pgf}
    \end{subfigure}
    \hfill
    \begin{subfigure}[b]{0.48\textwidth}
        \centering
        \input{figs/katagovisualizer/adv127-transfer.pgf}
    \end{subfigure}
    \caption{An adversary trained against \cpfivezerofive{} (left) or \cponetwentyseven{} (right), evaluated against both \cpfivezerofive{} and \cponetwentyseven{} at various visit counts. The adversary always uses 600 visits/move.}
    \label{fig:app:checkpoint-transfer}
\end{figure}

In Figure~\ref{fig:app:checkpoint-transfer}, we train adversaries against the \cpfivezerofive{} and \cponetwentyseven{} checkpoints respectively and evaluate against both checkpoints.
An adversary trained against \cpfivezerofive{} does better against \cpfivezerofive{} than \cponetwentyseven{}, despite \cpfivezerofive{} being a stronger agent.
The converse also holds: an agent trained against \cponetwentyseven{} does better against \cponetwentyseven{} than \cpfivezerofive{}.
This pattern holds across visit counts.
These results support the conclusion that different checkpoints have unique vulnerabilities.

\clearpage
\subsection{\textSwitch{Baseline attacks}{Baseline Attacks}}
\label{app:baseline-attacks}

We also test \emph{hard-coded} baseline adversarial policies. These baselines were inspired by the behavior of our trained adversary.
The \emph{Edge} attack plays random legal moves in the outermost $\ell^{\infty}$-box available on the board. 
The \emph{Spiral} attack is similar to the \emph{Edge} attack, except that it plays moves in a deterministic counterclockwise order, forming a spiral pattern.
The \emph{Random} attack plays uniformaly random legal moves.
Finally, we also implement \emph{Mirror Go}, a classic strategy that plays the opponent's last move reflected about the $y = x$ diagonal. If the opponent plays on $y = x$, Mirror Go plays that move reflected along the $y = -x$ diagonal. If the mirrored vertex is taken, Mirror Go plays the closest legal move by $\ell^{1}$ distance.

For each of these baseline policies, if the victim passes, then the policy will pass to end the game if passing is a winning move.

\begin{figure}[h]
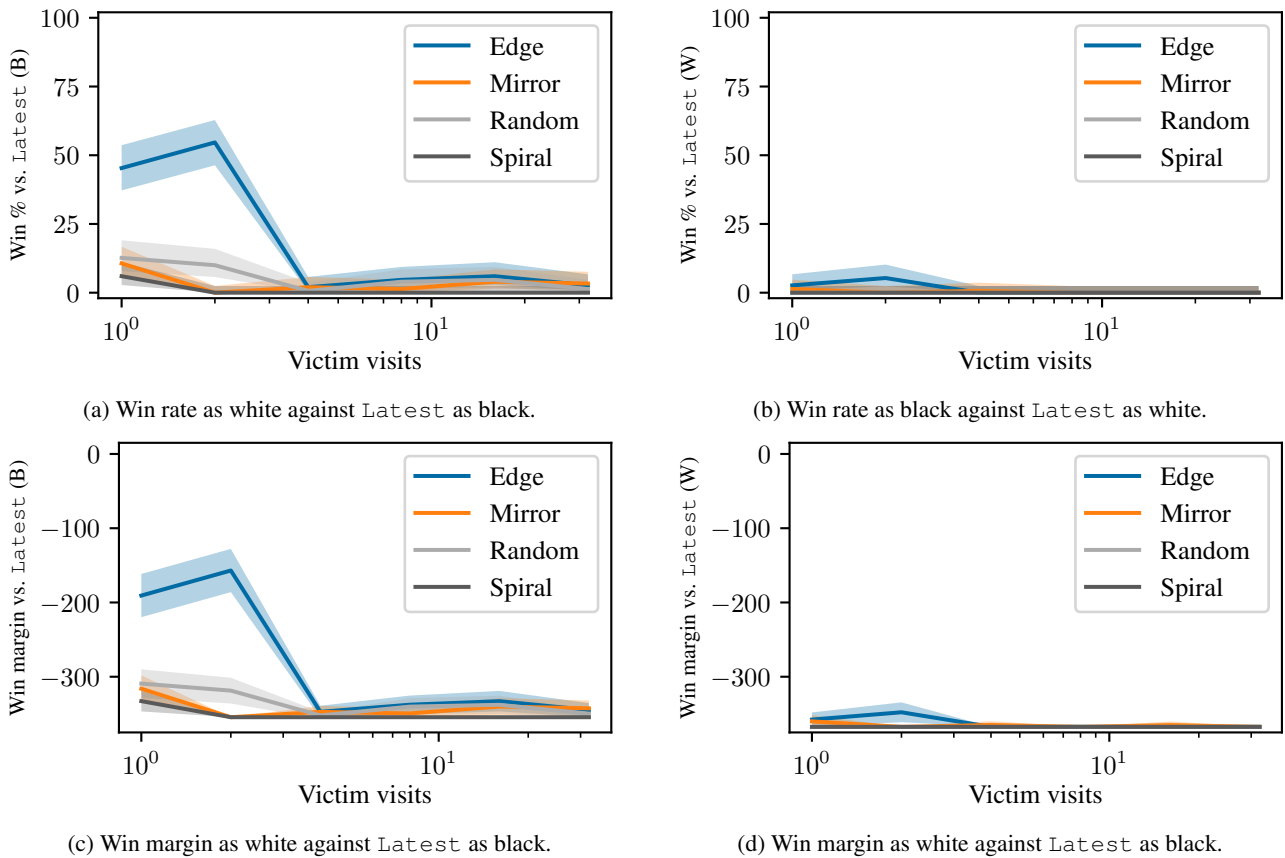

    \centering
    \begin{subfigure}[b]{0.48\textwidth}
        \input{figs/katagovisualizer/baseline-attack-cp505-b-win-rate.pgf}
        \caption{Win rate as white against \cpfivezerofive{} as black.}
    \end{subfigure}
    \hfill
    \begin{subfigure}[b]{0.48\textwidth}
        \input{figs/katagovisualizer/baseline-attack-cp505-w-win-rate.pgf}
        \caption{Win rate as black against \cpfivezerofive{} as white.}
    \end{subfigure}
    \\[0.1cm]
    \begin{subfigure}[b]{0.48\textwidth}
        \input{figs/katagovisualizer/baseline-attack-cp505-b-win-margin.pgf}
        \caption{Win margin as white against \cpfivezerofive{} as black.}
    \end{subfigure}
    \hfill
    \begin{subfigure}[b]{0.48\textwidth}
        \input{figs/katagovisualizer/baseline-attack-cp505-w-win-margin.pgf}
        \caption{Win margin as white against \cpfivezerofive{} as black.}
    \end{subfigure}
    \caption{Win rates and win margins of different baseline attacks versus \cpfivezerofive{} at varying visit counts ($x$-axis). 95\% confidence intervals are shown. The win margins are negative, indicating that on average the victim gains more points than the attack does.}
    \label{fig:app:baseline-vs-cp505}
\end{figure}

In Figure~\ref{fig:app:baseline-vs-cp505}, we plot the win rate and win margin of the baseline attacks against the KataGo victim \cpfivezerofive{}. 
The edge attack is the most successful, achieving a 45\% win rate when \cpfivezerofive{} plays as black with no search. None of the attacks work well once \cpfivezerofive{} is playing with at least 4 visits.

\begin{figure}
    \centering
    \begin{subfigure}[b]{0.48\textwidth}
        \input{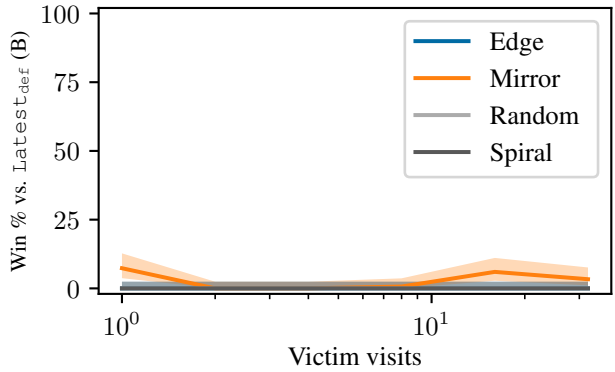}
        \caption{Win rate as white against \hardened{\cpfivezerofive{}} as black.}
    \end{subfigure}
    \hfill
    \begin{subfigure}[b]{0.48\textwidth}
        \input{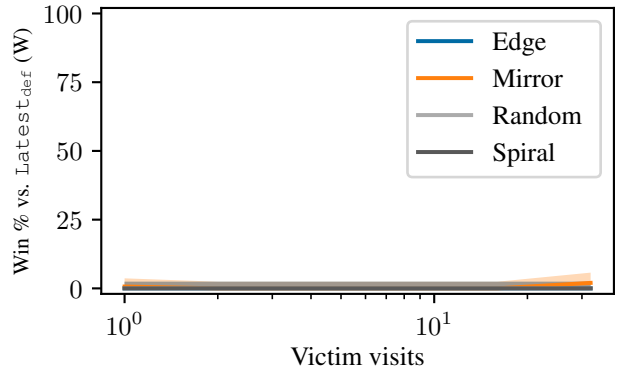}
        \caption{Win rate as black against \hardened{\cpfivezerofive{}} as white.}
    \end{subfigure}
    \\[0.5cm]
    \begin{subfigure}[b]{0.48\textwidth}
        \input{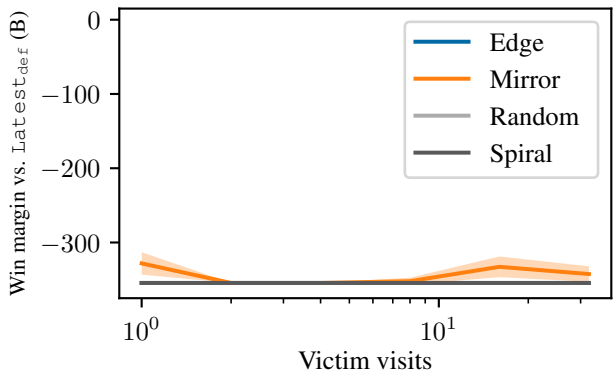}
        \caption{Win margin as white against \hardened{\cpfivezerofive{}} as black.}
    \end{subfigure}
    \hfill
    \begin{subfigure}[b]{0.48\textwidth}
        \input{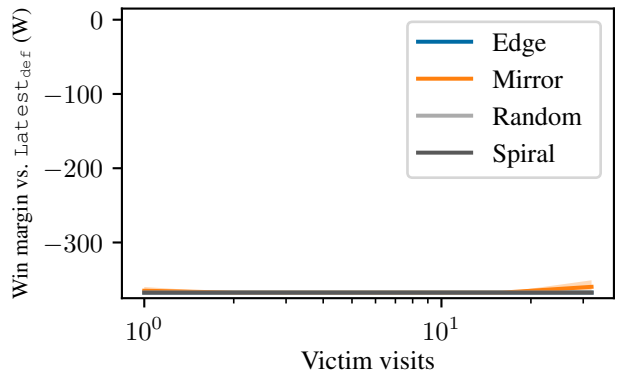}
        \caption{Win margin as white against \hardened{\cpfivezerofive{}} as black.}
    \end{subfigure}
    \caption{Win rates and win margins of different baseline attacks versus \hardened{\cpfivezerofive{}} at varying visit counts ($x$-axis). 95\% confidence intervals are shown. None of the attacks see much success.}
    \label{fig:app:baseline-vs-cp505h}
\end{figure}

In Figure~\ref{fig:app:baseline-vs-cp505h}, we plot the win rate and win margin against \hardened{\cpfivezerofive{}}. In this setting, none of the attacks work well even when \hardened{\cpfivezerofive{}} is playing with no search, though the mirror attack wins very occasionally.

We also run the baseline attacks against the weaker \cponetwentyseven{}, with Figure~\ref{fig:app:baseline-vs-cp127} plotting the win rate and win margin of the baseline attacks against \cponetwentyseven{} and Figure~\ref{fig:app:baseline-vs-cp127h} plotting the same statistics against \hardened{\cponetwentyseven{}}. 
\cponetwentyseven{} without search is shockingly vulnerable to simple attacks, losing all of its games against the edge and random attacks. Still, like \cpfivezerofive{}, \cponetwentyseven{} becomes much harder to exploit once it is playing with at least 4 visits, and \hardened{\cponetwentyseven{}} only suffers losses to the mirror attack.

\begin{figure}
    \centering
    \begin{subfigure}[b]{0.48\textwidth}
        \input{figs/katagovisualizer/baseline-attack-cp127-b-win-rate.pgf}
        \caption{Win rate as white against \cponetwentyseven{} as black.}
    \end{subfigure}
    \hfill
    \begin{subfigure}[b]{0.48\textwidth}
        \input{figs/katagovisualizer/baseline-attack-cp127-w-win-rate.pgf}
        \caption{Win rate as black against \cponetwentyseven{} as white.}
    \end{subfigure}
    \\[0.5cm]
    \begin{subfigure}[b]{0.48\textwidth}
        \input{figs/katagovisualizer/baseline-attack-cp127-b-win-margin.pgf}
        \caption{Win margin as white against \cponetwentyseven{} as black.}
    \end{subfigure}
    \hfill
    \begin{subfigure}[b]{0.48\textwidth}
        \input{figs/katagovisualizer/baseline-attack-cp127-w-win-margin.pgf}
        \caption{Win margin as white against \cponetwentyseven{} as black.}
    \end{subfigure}
    \caption{Win rates and win margins of different baseline attacks versus \cponetwentyseven{} at varying visit counts ($x$-axis). 95\% confidence intervals are shown.}
    \label{fig:app:baseline-vs-cp127}
\end{figure}

\begin{figure}
    \centering
    \begin{subfigure}[b]{0.48\textwidth}
        \input{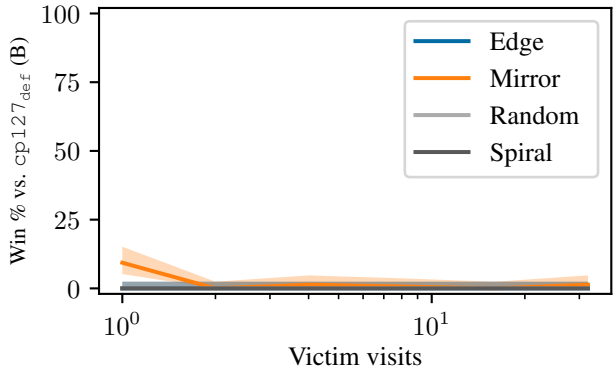}
        \caption{Win rate as white against \hardened{\cponetwentyseven{}} as black.}
    \end{subfigure}
    \hfill
    \begin{subfigure}[b]{0.48\textwidth}
        \input{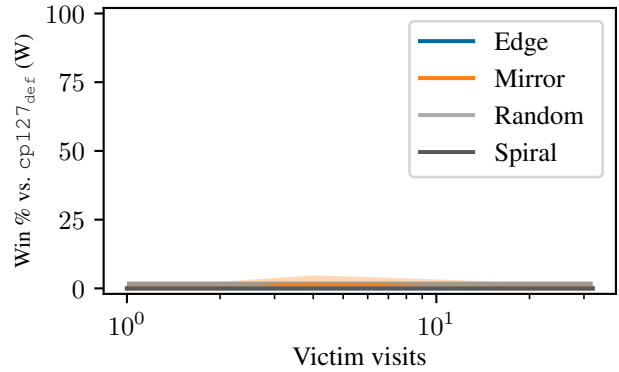}
        \caption{Win rate as black against \hardened{\cponetwentyseven{}} as white.}
    \end{subfigure}
    \\[0.5cm]
    \begin{subfigure}[b]{0.48\textwidth}
        \input{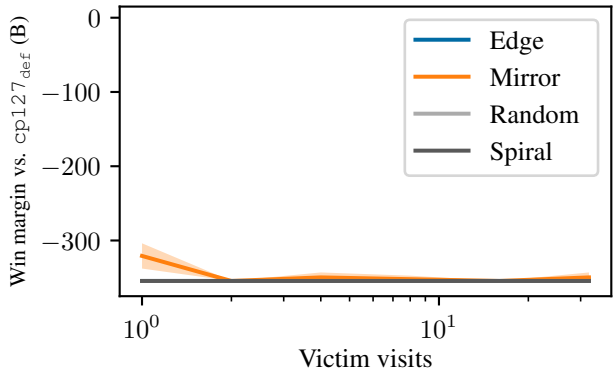}
        \caption{Win margin as white against \hardened{\cponetwentyseven{}} as black.}
    \end{subfigure}
    \hfill
    \begin{subfigure}[b]{0.48\textwidth}
        \input{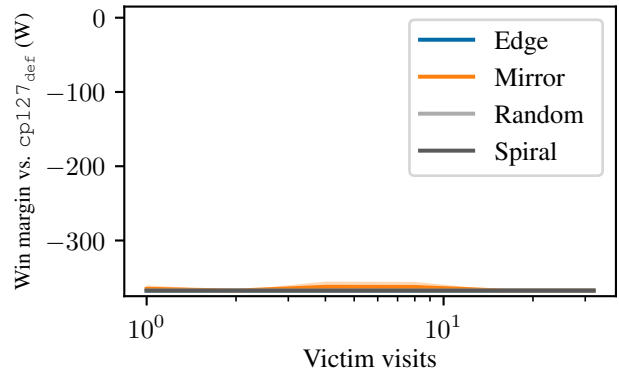}
        \caption{Win margin as white against \hardened{\cponetwentyseven{}} as black.}
    \end{subfigure}
    \caption{Win rates and win margins of different baseline attacks versus \hardened{\cponetwentyseven{}} at varying visit counts ($x$-axis). 95\% confidence intervals are shown.}
    \label{fig:app:baseline-vs-cp127h}
\end{figure}

\clearpage
\subsection{\textSwitch{Understanding the pass-adversary}{Understanding the Pass-Adversary}}
\label{app:experiments:understand-pass-adversary}

We observed in Figure~\ref{fig:cp505vis1-board:unhardened} that the pass-adversary appears to win by tricking the victim into passing prematurely, at a time favorable to the adversary.
In this section, we seek to answer three key questions.
First, \emph{why} does the victim pass even when it leads to a guaranteed loss?
Second, is passing \emph{causally} responsible for the victim losing, or would it lose anyway for a different reason?
Third, is the adversary performing a \emph{simple} strategy, or does it contain some hidden complexity?

Evaluating the \cpfivezerofive{} victim without search against the pass-adversary over $n = 250$ games, we find that \cpfivezerofive{} passes (and loses) in $247$ games and does not pass (and wins) in the remaining $3$ games.
In all cases, \cpfivezerofive{}'s value head estimates a win probability of greater than 99.5\% after the final move it makes, although its true win percentage is only 1.2\%.
\cpfivezerofive{} predicts it will \emph{win} by $\mu = 134.5$ points ($\sigma = 27.9$) after its final move, and passing would be reasonable if it were so far ahead.
But in fact it is just one move away from losing by an average of $86.26$ points.

We conjecture that the reason why the victim's prediction is so mistaken is that the games induced by playing against the adversarial policy are very different from those seen during the victim's self-play training.
Certainly, there is no fundamental inability of neural networks to predict the outcome correctly.
The adversary's value head achieves a mean-squared error of only $3.18$ (compared to $49{,}742$ for the victim) on the adversary's penultimate move.
The adversary predicts it will win $98.6\%$ of the time---extremely close to the true $98.8\%$ win rate in this sample.

To verify whether this pathological passing behavior is the reason the adversarial policy wins, we design a hard-coded defense for the victim, the pass-alive defense described in Section~\ref{sec:evaluation:no-search-hardened}.
Whereas the pass-adversary won greater than $99\%$ of games against vanilla \cpfivezerofive{}, we find that it \emph{loses} all 1600 evaluation games against \hardened{\cpfivezerofive{}}.
This confirms the pass-adversary wins via passing.

Unfortunately, this ``defense'' is of limited effectiveness: as we saw in Section~\ref{sec:evaluation:no-search-hardened}, repeating the attack method finds the cyclic-adversary that can beat it.
Moreover, the defense causes KataGo to continue to play even when a game is clearly won or lost, which is frustrating for human opponents.
The defense also relies on hard-coded knowledge about Go, using a search algorithm to compute the pass-alive territories.

Finally, we seek to determine if the adversarial policy is winning by pursuing a simple high-level strategy, or via a more subtle exploit such as forming an adversarial example by the pattern of stones it plays.
We start by evaluating the hard-coded baseline adversarial policies described in Appendix~\ref{app:baseline-attacks}.
In Figure~\ref{fig:app:baseline-vs-cp505}, we see that all of our baseline attacks perform substantially worse than our pass-adversary (Figure~\ref{fig:evaluation:search:victim-visits}). Moreover, when our baseline attacks do win it is usually due to the komi bonus given to white (as compensation for playing second), and therefore they almost never win as black. By contrast, our pass-adversary wins playing as either color, and often by a large margin (in excess of 50 points).

{
}

\clearpage
\subsection{\textSwitch{Performance of adversaries on other board sizes}{Performance of Adversaries on Other Board Sizes}}

Throughout this paper, we have been only reporting on the performance of our adversaries on $19 \times 19$ boards. During training, however,
our adversaries played games on different board sizes from $7 \times 7$ up to $19 \times 19$ with the default KataGo training frequencies listed in Table~\ref{tab:app:board-size-freq}, so our adversaries are also able to play on smaller board sizes.

\begin{table}[h]
    \centering
    \begin{tabular}{llllllllllllll}
        \toprule
         \textbf{Board size ($n \times n$)} & 7 & 8 & 9 & 10 & 11 & 12 & 13 & 14 & 15 & 16 & 17 & 18 & 19 \\
         \textbf{Training frequency (\%)} & 1 & 1 & 4 & 2 & 3 & 4 & 10 & 6 & 7 & 8 & 9 & 10 & 35 \\
         \bottomrule
    \end{tabular}
    \caption{Percentage of games played at each board size throughout the training of our adversaries. These percentages are the defaults for KataGo training.}
    \label{tab:app:board-size-freq}
\end{table}

\begin{figure}[h]
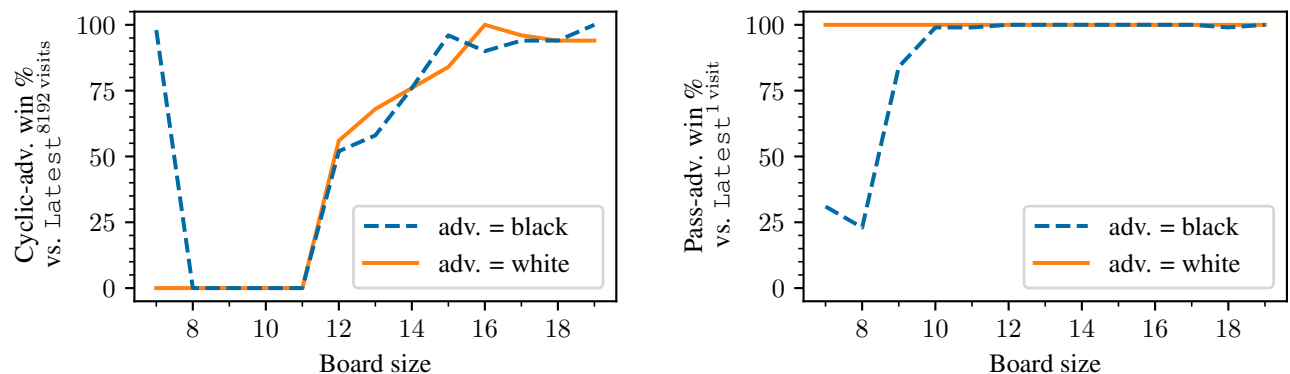

    \vspace{1cm}
    \centering
    \begin{subfigure}[b]{0.48\textwidth}
        \centering
        \input{figs/katagovisualizer/win-rate-vs-board-size-s545m.pgf}
        \caption{
          Cyclic-adversary with 600 visits versus \cpfivezerofive{} with 8192 visits.
        }
        \label{fig:app:cyclic-win-by-board-size}
    \end{subfigure}
    \hfill
    \begin{subfigure}[b]{0.48\textwidth}
        \centering
        \input{figs/katagovisualizer/win-rate-vs-board-size-s34m.pgf}
        \caption{
          Pass-adversary with 600 visits versus \cpfivezerofive{} without search.
        }
        \label{fig:app:pass-win-by-board-size}
    \end{subfigure}
    \caption{
        Win rate of our adversaries playing as each color against \cpfivezerofive{} on different board sizes.
    }
    \label{fig:app:win-by-board-size-adv}
\end{figure}

Figure~\ref{fig:app:win-by-board-size-adv} plots the win rate across different board sizes for the cyclic-adversary against \cpfivezerofive{} playing with 8192 visits (Figure~\ref{fig:app:cyclic-win-by-board-size}) and the pass-adversary against \cpfivezerofive{} playing without search (Figure~\ref{fig:app:pass-win-by-board-size}).
The komi is 8.5 for $7 \times 7$ boards, 9.5 for $8 \times 8$ boards, and 6.5 otherwise. These values were taken from analysis by David Wu, creator of KataGo, on fair komis for different board sizes under Chinese Go rules.\footnote{The komi analysis is at \url{https://lifein19x19.com/viewtopic.php?p=259358\#p259358}.}
These are the same komi settings we used during training, except that we had a configuration typo that swapped the komis for $8 \times 8$ boards and $9 \times 9$ boards during training.

The cyclic-adversary sets up the cyclic structure on board sizes of at least $12 \times 12$, and not coincidentally, those are board sizes on which the cyclic-adversary achieves wins.
The pass-adversary achieves wins on all board sizes via getting the victim to pass early, but on board sizes of $12 \times 12$ and smaller, the adversary sometimes plays around the edge of the board instead of playing primarily in one corner.

\begin{figure}[t]
    \centering
    \input{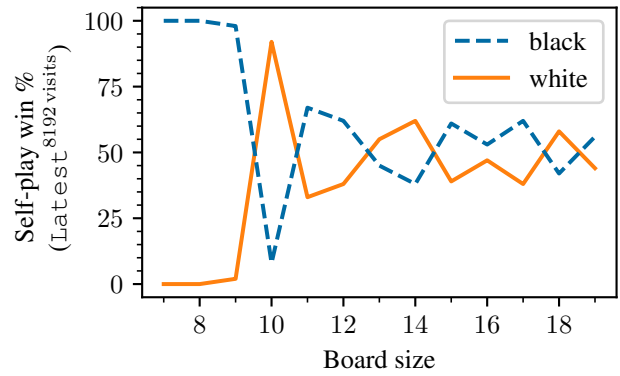}
    \caption{
        Win rate of \cpfivezerofive{} with 8192 visits playing against itself on different board sizes.
    }
    \label{fig:app:win-by-board-size-cp505}
\end{figure}

For comparison, Figure~\ref{fig:app:win-by-board-size-cp505} plots the win rate of \cpfivezerofive{} with 8192 visits playing against itself.

\vfill

\clearpage
\section{\textSwitch{Transfer of cyclic-adversary to other Go systems}{Transfer of Cyclic-Adversary to Other Go Systems}}
\label{app:transfer}

\subsection{\textSwitch{Algorithmic transfer}{Algorithmic Transfer}}
\label{app:transfer:algorithmic}

The cyclic-adversary transferred zero-shot to attacking Leela Zero and ELF OpenGo. Note that in this transfer, not only are the weights of the adversary trained against a different model (i.e. KataGo), the simulated victim in the search (\ourmctssampleabbrev{} simulating KataGo) is also different from the actual victim.

We ran ELF OpenGo with its final model and 80,000 rollouts.
A weaker model with 80,000 rollouts was already strong enough to consistently defeat several top-30 Go players~\citep{tian2019}.
We ran Leela Zero with its final model (released February 15, 2021), unlimited time, and a maximum of 40,000 visits per move. We turned off resignation for both ELF and Leela. We expect that ELF and Leela play at a superhuman level with these parameters. Confirming this, we found that ELF and Leela with these parameter settings defeat \cpfivezerofive{} with 128 visits a majority of the time, and we estimate in Appendix \ref{app:experiments:strength:katago-search} that \cpfivezerofive{} with 128 visits plays at a superhuman level.

Our adversary won 8/132 = 6.1\% games against Leela Zero and 5/142 = 3.5\% games against ELF OpenGo. Although this win rate is considerably lower than that attained against KataGo, to beat these systems at all zero-shot is significant given that even the best human players almost never win against these systems.

\subsection{\textSwitch{Human transfer}{Human Transfer}}
\label{app:transfer:human}

The cycle attack discovered by our algorithmic adversaries can also be implemented by humans. An author, who is a Go expert, successfully attacked a variety of Go programs including KataGo and Leela Zero playing with 100,000 visits, both of which even top professional players are normally unable to beat. They also won 14/15 games against \texttt{JBXKata005}, a custom KataGo implementation not affiliated with the authors, which was the strongest bot available to play on the KGS Go Server at the time of the test. In addition, they also tested giving JBXKata005 3, 5, and 9 handicap stones (additional moves at the beginning of the game), and won in all cases.

\begin{figure}
    \centering
    \begin{subfigure}{0.48\textwidth}
        \centering
        \includesvg[inkscapelatex=false, width=0.9\textwidth]{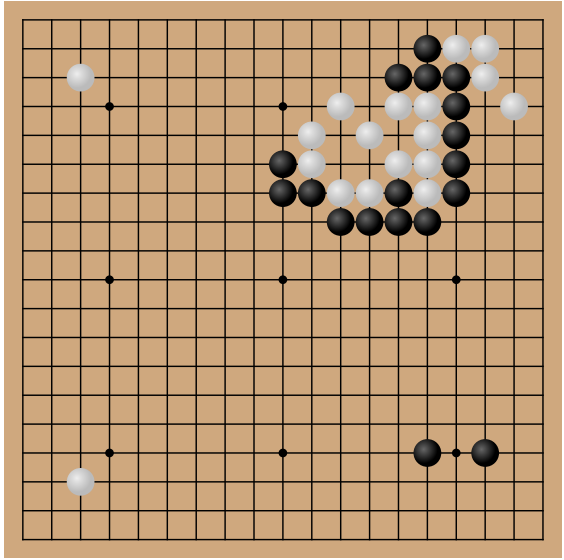}
        
	\caption{Move 36: human has set up the inner group (top right middle) around which to lure the victim to create a cycle.}
    \label{fig:humanatk-katago100k:move-36}
    \end{subfigure}
    \quad
    \begin{subfigure}{0.48\textwidth}
        \centering
        \includesvg[inkscapelatex=false, width=0.9\textwidth]{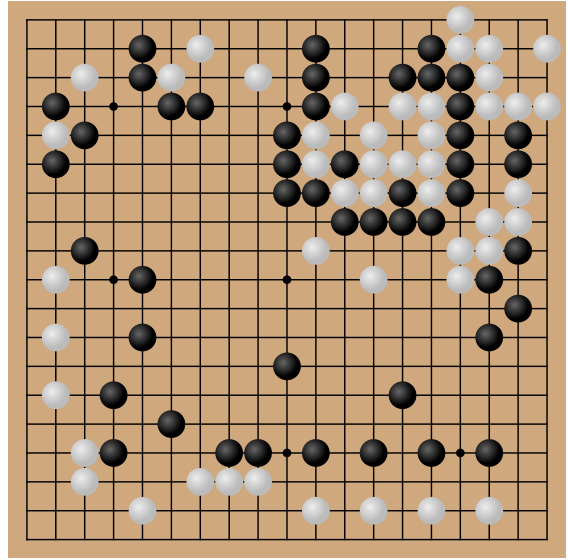}
	\caption{Move 95: human set up a position where it is optimal for black to fill in the missing part of the cycle.}
    \label{fig:humanatk-katago100k:move-95}
    \end{subfigure}
    \\
    \vspace{2mm}
    \begin{subfigure}{0.48\textwidth}
        \centering
        \includesvg[inkscapelatex=false, width=0.9\textwidth]{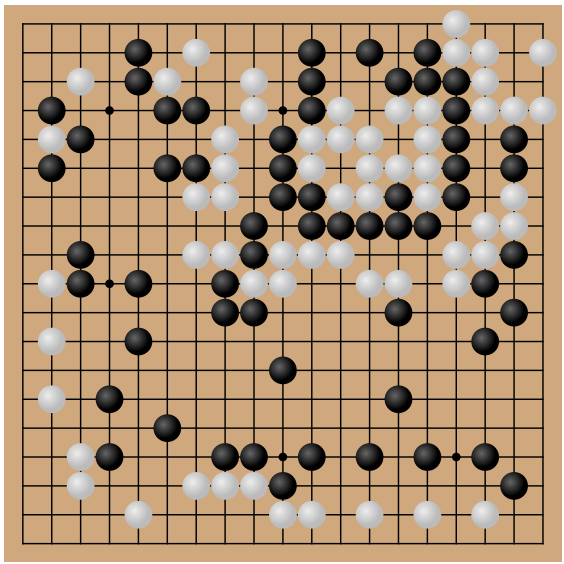}
	\caption{Move 122: victim's cycle group is now surrounded. It remains to capture it before the victim catches on and defends.}
    \label{fig:humanatk-katago100k:move-122}
    \end{subfigure}
    \quad
    \begin{subfigure}{0.48\textwidth}
        \centering
        \includesvg[inkscapelatex=false, width=0.9\textwidth]{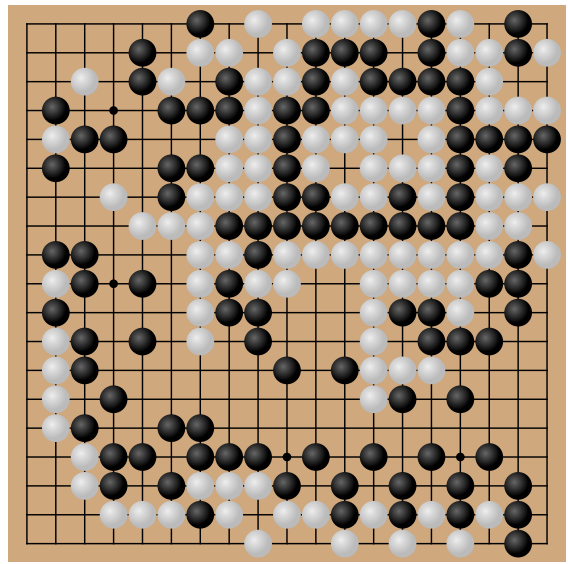}
	\caption{Move 210: by now none of the victim's stones in the top right can avoid capture. The victim finally realizes and resigns.}
    \label{fig:humanatk-katago100k:move-210}
    \end{subfigure}
	\caption{Human (white) beats KataGo with 100k visits (black).
	}
    \label{fig:humanatk-katago100k}
\end{figure}

\begin{figure}
    \centering
    \begin{subfigure}{0.48\textwidth}
        \centering
        \includesvg[inkscapelatex=false, width=0.9\textwidth]{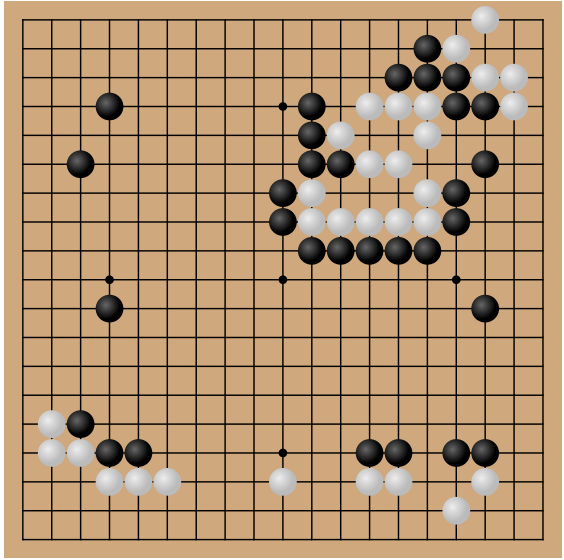}
	    \caption{Move 61: human has set up the inner group (top right middle) around which to lure the victim to create a cycle.}
        \label{fig:humanatk-LZ100k:move-36}
    \end{subfigure}
    \quad
    \begin{subfigure}{0.48\textwidth}
        \centering
        \includesvg[inkscapelatex=false, width=0.9\textwidth]{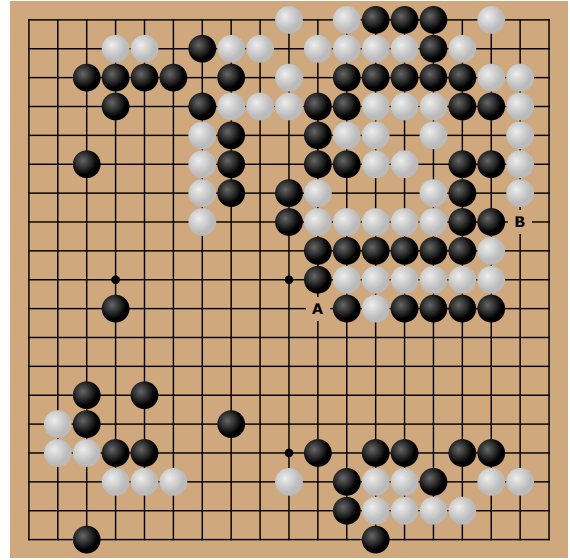}
	    \caption{Move 95: human next plays A, instead of the safer connecting move at B, to attempt to encircle victim's cyclic group.}
        \label{fig:humanatk-LZ100k:move-95}
    \end{subfigure}
    \\
    \vspace{2mm}
    \begin{subfigure}{0.48\textwidth}
        \centering
        \includesvg[inkscapelatex=false, width=0.9\textwidth]{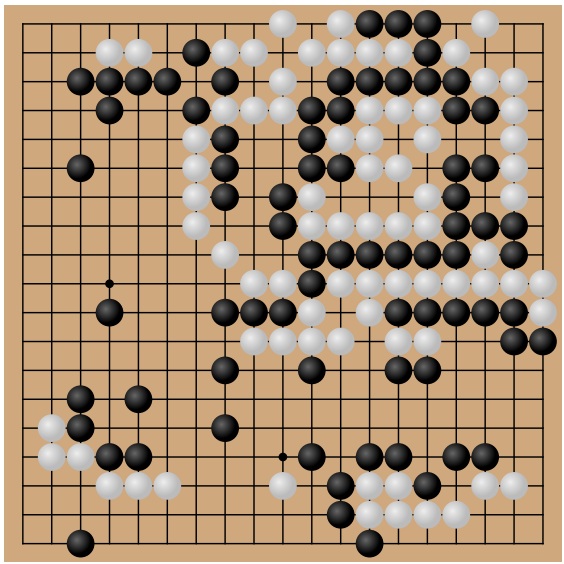}
	    \caption{Move 156: the encirclement is successful. Victim could survive by capturing one of the encircling groups, but will it?}
        \label{fig:humanatk-LZ100k:move-156}
    \end{subfigure}
    \quad
    \begin{subfigure}{0.48\textwidth}
        \centering
        \includesvg[inkscapelatex=false, width=0.9\textwidth]{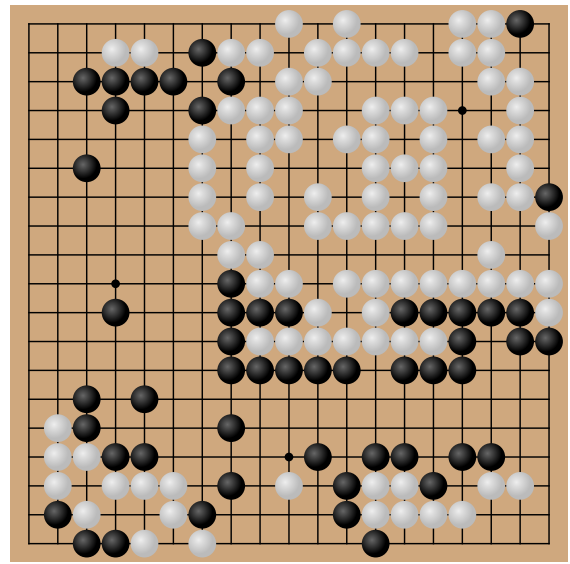}
	    \caption{Move 199: the victim failed to see the danger in time, was captured, and resigns.}
        \label{fig:humanatk-LZ100k:move-210}
    \end{subfigure}
	\caption{Human (white) beats Leela Zero with 100k visits (black).}
    \label{fig:humanatk-LZ100k}
\end{figure}

\begin{figure}
    \centering
    \begin{subfigure}{0.48\textwidth}
        \centering
        \includesvg[inkscapelatex=false, width=0.9\textwidth]{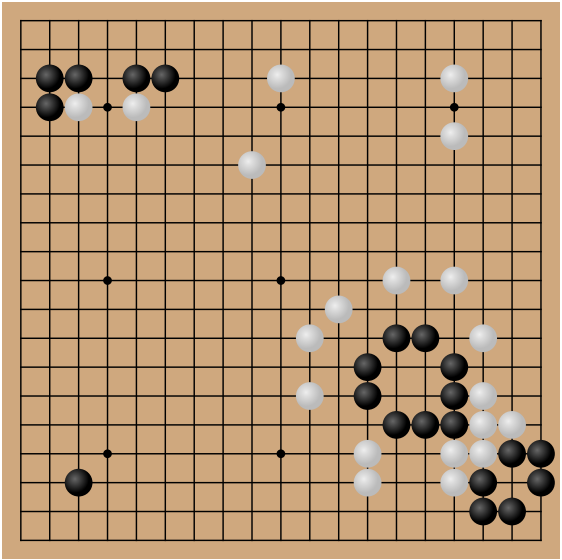}
	    \caption{Move 41: the frame of the cyclic group is set up (lower right middle)}
        \label{fig:humanatk-JBXKata005:move-41}
    \end{subfigure}
    \quad
    \begin{subfigure}{0.48\textwidth}
        \centering
        \includesvg[inkscapelatex=false, width=0.9\textwidth]{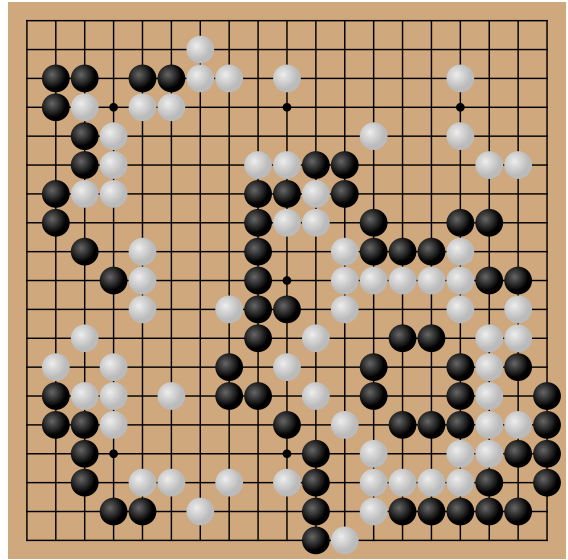}
	    \caption{Move 133: human has completed loose encirclement of the victim's cyclic group.}
        \label{fig:humanatk-JBXKata005:move-133}
    \end{subfigure}
    \\
    \vspace{2mm}
    \begin{subfigure}{0.48\textwidth}
        \centering
        \includesvg[inkscapelatex=false, width=0.9\textwidth]{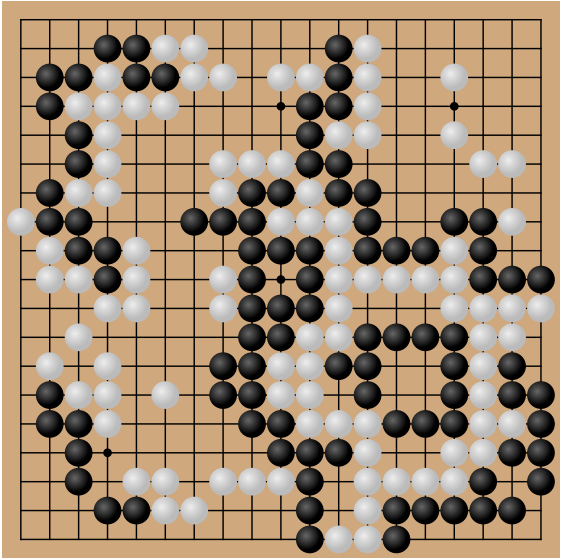}
	    \caption{Move 189: although victim's cyclic group has a number of liberties left, it can no longer avoid capture and the game is decided.}
        \label{fig:humanatk-JBXKata005:move-189}
    \end{subfigure}
    \quad
    \begin{subfigure}{0.48\textwidth}
        \centering
        \includesvg[inkscapelatex=false, width=0.9\textwidth]{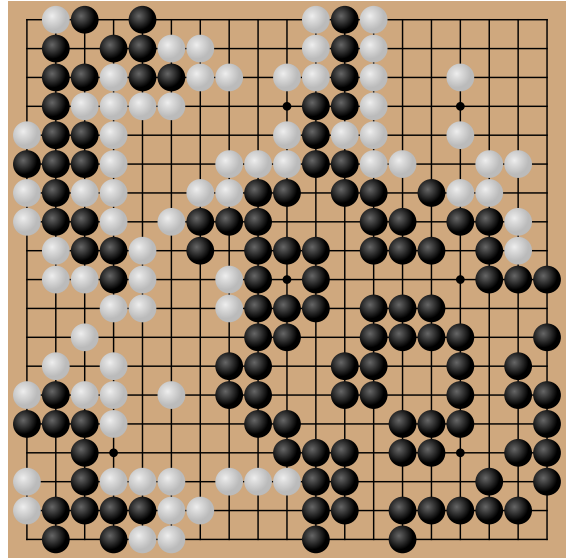}
	    \caption{Move 237: after nearly 40 more moves the cyclic group is captured. Victim realizes game is lost and resigns.\\}
        \label{fig:humanatk-JBXKata005:move-210}
    \end{subfigure}
	\caption{Human (black) beats JBXKata005 (white).}
    \label{fig:humanatk-JBXKata005}
\end{figure}

\begin{figure}
    \centering
    \begin{subfigure}{0.48\textwidth}
        \centering
        \includesvg[inkscapelatex=false, width=0.9\textwidth]{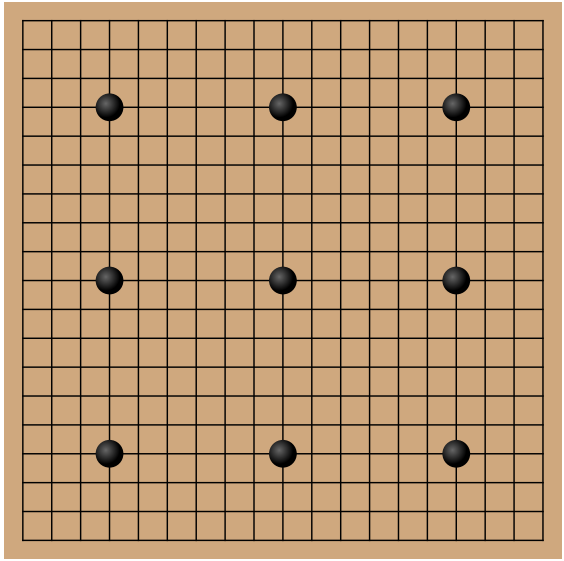}
	    \caption{Move 0: starting board position. In contrast to a normal game starting with an empty board, here the victim received 9 handicap stones, giving it an enormous initial advantage.}
        \label{fig:humanatk-JBXKata005-handi9:move-0}
    \end{subfigure}
    \quad
    \begin{subfigure}{0.48\textwidth}
        \centering
        \includesvg[inkscapelatex=false, width=0.9\textwidth]{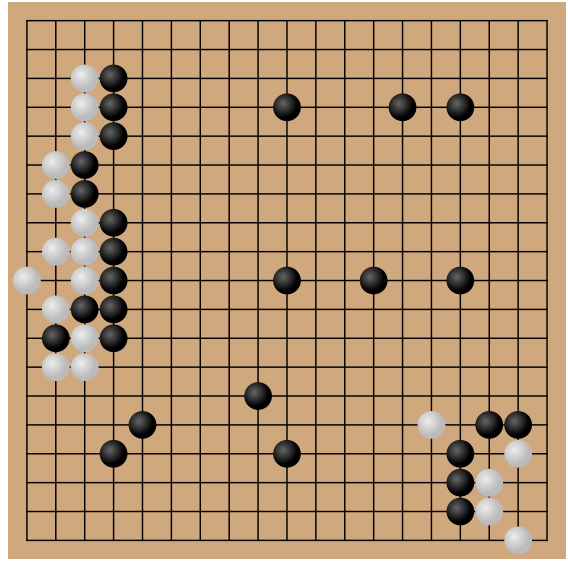}
	    \caption{Move 38: setting up the inside group is slightly more challenging here, since the victim has it surrounded from the start.\\}
        \label{fig:humanatk-JBXKata005-handi9:move-38}
    \end{subfigure}
    \\
    \vspace{2mm}
    \begin{subfigure}{0.48\textwidth}
        \centering
        \includesvg[inkscapelatex=false, width=0.9\textwidth]{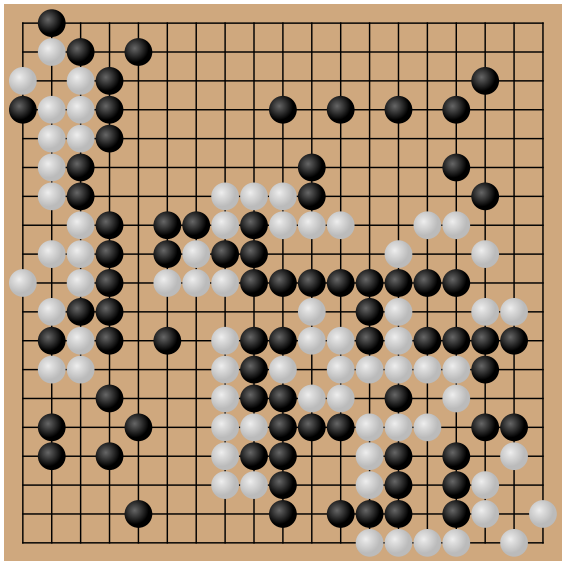}
	    \caption{Move 141: an encirclement is complete, but there are numerous defects. Victim could easily live inside or capture key stones.}
        \label{fig:humanatk-JBXKata005-handi9:move-141}
    \end{subfigure}
    \quad
    \begin{subfigure}{0.48\textwidth}
        \centering
        \includesvg[inkscapelatex=false, width=0.9\textwidth]{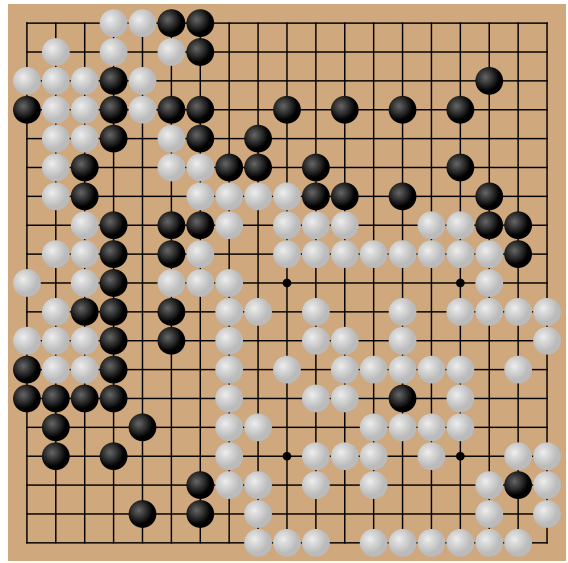}
	    \caption{Move 227: victim fails to grasp any option to survive. Its group is captured and it resigns.}
        \label{fig:humanatk-JBXKata005-handi9:move-227}
    \end{subfigure}
	\caption{Human (white) beats JBXKata005 (black), giving it 9 handicap stones.}
    \label{fig:humanatk-JBXKata005-handi9}
\end{figure}

In the following figures we present selected positions from the games. The full games are available on our \href{\demosite/human-evaluation#human_vs_kata100k}{website}. First, Figure~\ref{fig:humanatk-katago100k} shows key moments in a game against KataGo with 100k visits. Figure~\ref{fig:humanatk-LZ100k} shows the same against LeelaZero with 100k visits. Figure~\ref{fig:humanatk-JBXKata005} shows a normal game against JBXKata005, while Figure~\ref{fig:humanatk-JBXKata005-handi9} shows a game where JBXKata005 received the advantage of a 9 stone handicap at the start. In each case the strategy is roughly the following: first, set up an ``inside'' group and let or lure the victim to surround it, creating a cyclic group. Second, surround the cyclic group. Third, guarantee the capture before the victim realizes it is in danger and defends. In parallel to all these steps, one must also make sure to secure enough of the rest of the board that capturing the cyclic group will be enough to win.

\clearpage
\section{\textSwitch{The role of search in robustness}{The Role of Search in Robustness}}
\label{app:role-of-search}

Asymptotically, search leads to robustness: with infinite search, a model could perfectly evaluate every possible move and never make a mistake. However, this level of search is computationally impossible. Our results show that in computationally feasible regimes, search is insufficient to produce full robustness. Does search have a meaningful impact at all on robustness in realistic regimes? In this appendix we show that it does substantially improve robustness, and consequently, while not a full solution, it is nonetheless a practical tool for creating more robust models and pipelines.

\begin{figure}[h]
    \centering
    \input{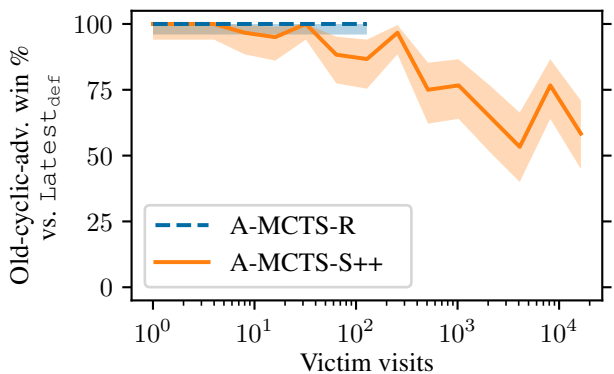}
    \caption{
    We evaluate the win rate of an older version of our cyclic-adversary playing with 200 visits / move against \hardened{\cpfivezerofive{}} playing with varying amounts of search. The cyclic-adversary was trained for 498 million steps and had a curriculum whose victims only went up to 256 visits of search.
    Shaded regions and error bars denote 95\% Clopper-Pearson confidence intervals over 60 games for \ourmctssamplesymmetryabbrev{} and 90 games for \ourmctsperfectabbrev{}. The adversary does better and wins all its games when performing \ourmctsperfectabbrev{}, which models the victim perfectly.
}
    \label{fig:app:s497mil-amcts-r-vs-victim-visits}
\end{figure}

In results discussed previously, we see that for a fixed adversary, increasing victim search improves its win rate (e.g. Figure~\ref{fig:evaluation:search:victim-visits-hardened}). This provides evidence search increases robustness. However, there are potential confounders. First, the approximation that \ourmctssampleabbrev{} and \ourmctssamplesymmetryabbrev{} makes of the victim becomes less accurate the more search the victim has, making it harder to exploit for this algorithm regardless of its general change in robustness. (Indeed, we see in Figure~\ref{fig:app:s497mil-amcts-r-vs-victim-visits} that \ourmctsperfectabbrev{}, which perfectly models the victim, achieves a higher win rate than \ourmctssamplesymmetryabbrev{}.) Second, for a fixed adversary, the further the victim search diverges from the training, the more out-of-distribution the victim becomes. Third, it is possible that higher search improves winrate not through improved robustness or judgment but because it simply has less tendency to create cyclic positions. A person who hates mushrooms is less likely to eat a poisonous one, regardless of their judgment identifying them or towards risk in general.

In order to remove some of these confounders, we analyze board positions from games of the cyclic-adversary vs. victim which the victim lost. The cyclic-adversary examined here was trained for 498 million steps, making it an older version whose curriculum only went up to victims with 256 visits. The positions were selected manually by an author who is an expert Go player to be the last opportunity for the victim to win the game with subsequent best play by both sides. To facilitate accurate and informative analysis, they selected positions with relatively decisive correct and incorrect moves. This is necessary because many positions have numerous inconclusive moves that postpone resolution of the conflict until a later move (typically, a strong threat which requires an immediate answer, but does not change the overall situation). We vary the level of victim search from the original 1.6k visits, to 5k, 50k, 100k, 250k, and 500k and examine how much search is needed for the victim to rank a correct move as the best one. This corresponds roughly to asking "how much search is needed for the victim to play correctly in this position?" (ignoring stochasticity in the search and in the move choice, which depends on the chosen temperature hyperparameter).

Results are shown in Table~\ref{tab:app:position-visits}. ``\cmark'' indicates a correct move that should lead to victory was ranked \#1, ``\xmark'' indicates a wrong move that should lead to defeat was \#1, while ``?'' indicates an inconclusive move ranked \#1.

\begin{table}[t]
    \centering
    \makebox[\textwidth]{%
    \begin{tabular}{lllllllllll}
         \toprule
         \textbf{Visits} & \textbf{Game 0} & \textbf{Game 1} & \textbf{Game 2} & \textbf{Game 3} & \textbf{Game 4} & \textbf{Game 5} & \textbf{Game 6} & \textbf{Game 7} & \textbf{Game 8} & \textbf{Game 9} \\
         \midrule
         \textbf{1.6k} & \nosymbol & \nosymbol & \nosymbol & \nosymbol  & \nosymbol & \nosymbol & \nosymbol & \nosymbol & \nosymbol & \nosymbol\\
         \textbf{5k} & \nosymbol & \yessymbol & ? &  \nosymbol  & \nosymbol & \yessymbol & \nosymbol & \nosymbol  & \nosymbol & \nosymbol \\
         \textbf{50k} & \yessymbol & \yessymbol & ? & ? & \nosymbol & \yessymbol & \nosymbol & \yessymbol  & \nosymbol & \yessymbol\\
         \textbf{100k} & \yessymbol & \yessymbol & ? & ? & \yessymbol & \yessymbol & \nosymbol & ?  & \nosymbol & \yessymbol\\
         \textbf{250k} & \yessymbol & \yessymbol & \yessymbol & ? & \yessymbol & \yessymbol & \nosymbol & \yessymbol  & \nosymbol & \yessymbol\\
         \textbf{500k} & \yessymbol & \yessymbol & \yessymbol & ? & \yessymbol & \yessymbol & \yessymbol & \yessymbol & \nosymbol & \yessymbol\\
         \bottomrule
    \end{tabular}
    }
    \caption{
        Examining how much search is needed to make the correct move in deciding positions. The original victim, which played the wrong move and consequently lost, used 1.6k visits of search. Higher visits leads to more correct moves in these positions, suggesting improved robustness.
    }
    \label{tab:app:position-visits}
\end{table}

We also investigated games played by the fully-trained adversary (i.e. the adversary whose curriculum goes up to 131k visits) against KataGo with 10 million visits. 
We find that when the adversary wins in this setting, the decisive move is played a greater number of moves before the cyclic group is captured than in the previous setting.
This means that more victim search is needed to see the correct result.
The adversary has likely learned to favor such positions during the additional training against higher search victims.
There is also likely a selection bias, as the victim will likely win when the attack is less concealed, although as the adversary achieves a 76.7\% win rate this effect cannot be substantial.

To test this impression quantitatively, we randomly sampled 25 games in which the adversary wins from each set of opponents. We resampled 1 outlier involving an abnormal, very complicated triple ko. For each game, we determined the last move from which the victim could have won. We then measure the number of moves from that move until the cyclic group is captured. For this measurement we consider the fastest possible sequence to capture, which might slightly differ from the actual game, because in the actual game the victim might resign before the final capture, or the adversary might not capture immediately if there is no way for the victim to escape. We include in the count any moves which postpone the capture that the victim actually played. This represents a middle ground: including all possible moves to postpone the capture could result in counting many moves that were irrelevant to the search (e.g. moves that require an answer but have no effect on the final result, which the victim realized without significantly affecting the search). On the other hand, removing all moves that postpone the capture might ignore moves that the victim thought were beneficial and had a significant effect on the search. Since the goal is to determine if there is a difference in how well hidden the attack is vis-a-vis the search, this middle ground is the most informative.

We find the lower search games had a mean moves-to-capture of 6.36 moves with a standard deviation of 2.87, while the higher search games had a mean of 8.36 with a standard deviation of 2.69. With a standard t-test for difference in means, this is significant at the 5\% level ($p = 0.0143$). This also matches a qualitative assessment that the higher visit positions are more complex and have more potential moves, even if they are not part of the optimal sequence. 
Overall, this suggests that increased search leads to increased robustness, but that the adversary is able to partially combat this by setting up complex positions.

We observe that with lower search, there are 6 games which have a 3 move difference between the deciding move and the capture, while with higher search there are none less than 5. Is a 3 move trap too few to catch a high search victim? We examine these 6 positions (shown in Figures~\ref{fig:search-position-examples-round2} and \ref{fig:search-position-examples-round2-part2}) further by varying the amount of search, as in the preceding experiment. Results are shown in Table~\ref{tab:app:position-visits-round2}. Similar to the positions examined previously, higher search typically leads to a correct move, although there is one exception where none of the visit levels tested fixed the victim's mistake. We tested this one position with 1 million, 2.5 million, and 10 million visits, and found that 1 million is still insufficient but 2.5 million and 10 million find the correct move. Therefore, it does seem these positions are not enough to fool a high search victim. Once again, this indicates overall that search does not give full robustness but still yields improvements.

We see that in 8 out of 10 positions, 500k visits leads to a winning move, and in many of the positions a winning move is found with substantially fewer visits. These search numbers are well within a feasible range. Although the sample size is limited due to the substantial manual analysis needed for each position, the results provide consistent evidence that adding a reasonable amount of search is indeed beneficial for robustness.

We show the board positions analyzed in Figures~\ref{fig:search-position-examples1}, \ref{fig:search-position-examples2}, and \ref{fig:search-position-examples3}. Moves are marked according to the preceding table, though note the markings for wrong and inconclusive moves are non-exhaustive. Full game records are available on our \href{\demosite/game-analysis#position-analysis}{website}.

To further confirm that reasonable amounts of search are not sufficient for robustness, we examined 5 positions from random games where our adversary beat a KataGo victim with 1 million visits. We determined the last chance for victory as above, and gave KataGo 1 \textit{billion} visits to find a correct move. The positions are shown in Figure~\ref{fig:search-position-examples-1b} and full game records are available on our \href{\demosite/game-analysis#position-analysis-1b}{website}. In all 5 positions, a wrong game-losing move was still selected. This is around two orders of magnitude beyond the number of visits used in typical computer vs. computer tournaments, let alone normal games against humans. Consequently, short of revolutionary progress in hardware, we are unlikely to be able to solve this vulnerability through increasing search alone.

\begin{table}[h]
    \centering
    \begin{tabular}{lllllll}
         \toprule
         \textbf{Visits} & \textbf{Game 0} & \textbf{Game 1} & \textbf{Game 2} & \textbf{Game 3} & \textbf{Game 4} & \textbf{Game 5}  \\
         \midrule
         \textbf{1.6k} & \nosymbol & \nosymbol & \nosymbol & \nosymbol  & \nosymbol & \nosymbol \\
         \textbf{5k} & \nosymbol & \nosymbol & \nosymbol &  \nosymbol  & \nosymbol & \nosymbol  \\
         \textbf{50k} & \yessymbol & \yessymbol & \nosymbol & \nosymbol & \nosymbol & \nosymbol \\
         \textbf{100k} & \yessymbol & \yessymbol & \nosymbol & \nosymbol & \nosymbol & \nosymbol \\
         \textbf{250k} & \yessymbol & \yessymbol & \yessymbol & \nosymbol & \nosymbol & \yessymbol \\
         \textbf{500k} & \yessymbol & \yessymbol & \yessymbol & \nosymbol & \yessymbol & \yessymbol \\
         \bottomrule
    \end{tabular}
    \caption{
        Examining how much search is needed to make the correct move in positions with a 3 move difference between the deciding move and capture. Similar to the preceding table, the original victim had 1.6k visits. Higher visits again leads to more correct moves and improved robustness.
    }
    \label{tab:app:position-visits-round2}
\end{table}

\begin{figure}
    \centering
    \begin{subfigure}{0.48\textwidth}
        \centering
        \includesvg[inkscapelatex=false, width=0.9\textwidth]{figs/boardstates/specific-positions/game-0-2col}
        
	\caption{White to play.}
    \end{subfigure}
    \quad
    \begin{subfigure}{0.48\textwidth}
        \centering
        \includesvg[inkscapelatex=false, width=0.9\textwidth]{figs/boardstates/specific-positions/game-1-2col}
	\caption{Black to play.}
    \end{subfigure}
    \\
    \vspace{2mm}
    \begin{subfigure}{0.48\textwidth}
        \centering
        \includesvg[inkscapelatex=false, width=0.9\textwidth]{figs/boardstates/specific-positions/game-2-2col}
	\caption{White to play.}
    \end{subfigure}
    \quad
    \begin{subfigure}{0.48\textwidth}
        \centering
        \includesvg[inkscapelatex=false, width=0.9\textwidth]{figs/boardstates/specific-positions/game-3-2col}
	\caption{Black to play.}
    \end{subfigure}

	\caption{Part 1 of positions analyzed with varying levels of search. Correct moves are marked ``\cmark'', and non-exhaustive examples of incorrect and inconclusive moves that the victim likes to play are marked with ``\xmark'' and ``?'' respectively.
	}
    \label{fig:search-position-examples1}
\end{figure}

\begin{figure}
    \centering
    \begin{subfigure}{0.48\textwidth}
        \centering
        \includesvg[inkscapelatex=false, width=0.9\textwidth]{figs/boardstates/specific-positions/game-4-2col}
        
	\caption{White to play.}
    \end{subfigure}
    \quad
    \begin{subfigure}{0.48\textwidth}
        \centering
        \includesvg[inkscapelatex=false, width=0.9\textwidth]{figs/boardstates/specific-positions/game-5-2col}
	\caption{Black to play.}
    \end{subfigure}
    \\
    \vspace{2mm}
    \begin{subfigure}{0.48\textwidth}
        \centering
        \includesvg[inkscapelatex=false, width=0.9\textwidth]{figs/boardstates/specific-positions/game-6-2col}
	\caption{White to play.}
    \end{subfigure}
    \quad
    \begin{subfigure}{0.48\textwidth}
        \centering
        \includesvg[inkscapelatex=false, width=0.9\textwidth]{figs/boardstates/specific-positions/game-7-2col}
	\caption{Black to play.}
    \end{subfigure}

	\caption{Part 2 of positions analyzed with varying levels of search. Correct moves are marked ``\cmark'', and non-exhaustive examples of incorrect and inconclusive moves that the victim likes to play are marked with ``\xmark'' and ``?'' respectively.
	}
    \label{fig:search-position-examples2}
\end{figure}

\begin{figure}
    \centering
    \begin{subfigure}{0.48\textwidth}
        \centering
        \includesvg[inkscapelatex=false, width=0.9\textwidth]{figs/boardstates/specific-positions/game-8-2col}
        
	\caption{Black to play.}
    \end{subfigure}
    \quad
    \begin{subfigure}{0.48\textwidth}
        \centering
        \includesvg[inkscapelatex=false, width=0.9\textwidth]{figs/boardstates/specific-positions/game-9-2col}
	\caption{White to play.}
    \end{subfigure}

	\caption{Part 3 of positions analyzed with varying levels of search. Correct moves are marked ``\cmark'', and non-exhaustive examples of incorrect and inconclusive moves that the victim likes to play are marked with ``\xmark'' and ``?'' respectively.
	}
    \label{fig:search-position-examples3}
\end{figure}

\begin{figure}
    \centering
    \begin{subfigure}{0.48\textwidth}
        \centering
        \includesvg[inkscapelatex=false, width=0.9\textwidth]{figs/boardstates/specific-positions/round2-game-0-2col}
        
	\caption{Black to play.}
    \end{subfigure}
    \quad
    \begin{subfigure}{0.48\textwidth}
        \centering
        \includesvg[inkscapelatex=false, width=0.9\textwidth]{figs/boardstates/specific-positions/round2-game-1-2col}
	\caption{White to play.}
    \end{subfigure}
    \\
    \vspace{2mm}
    \begin{subfigure}{0.48\textwidth}
        \centering
        \includesvg[inkscapelatex=false, width=0.9\textwidth]{figs/boardstates/specific-positions/round2-game-2-2col}
	\caption{White to play.}
    \end{subfigure}
    \quad
    \begin{subfigure}{0.48\textwidth}
        \centering
        \includesvg[inkscapelatex=false, width=0.9\textwidth]{figs/boardstates/specific-positions/round2-game-3-2col}
	\caption{White to play.}
    \end{subfigure}
	\caption{Positions with a 3 move difference between deciding move and capture, analyzed with varying levels of search. Correct moves are marked ``\cmark'', and non-exhaustive examples of incorrect and inconclusive moves that the victim likes to play are marked with ``\xmark'' and ``?'' respectively.
	}
    \label{fig:search-position-examples-round2}
\end{figure}

\begin{figure}
    \centering
    \begin{subfigure}{0.48\textwidth}
        \centering
        \includesvg[inkscapelatex=false, width=0.9\textwidth]{figs/boardstates/specific-positions/round2-game-4-2col}
        
	\caption{Black to play.}
    \end{subfigure}
    \quad
    \begin{subfigure}{0.48\textwidth}
        \centering
        \includesvg[inkscapelatex=false, width=0.9\textwidth]{figs/boardstates/specific-positions/round2-game-5-2col}
	\caption{White to play.}
    \end{subfigure}
	\caption{Part 2 of positions with a 3 move difference between deciding move and capture, analyzed with varying levels of search. Correct moves are marked ``\cmark'', and non-exhaustive examples of incorrect and inconclusive moves that the victim likes to play are marked with ``\xmark'' and ``?'' respectively.
	}
    \label{fig:search-position-examples-round2-part2}
\end{figure}

\begin{figure}
    \centering
    \begin{subfigure}{0.31\textwidth}
        \centering
        \includesvg[inkscapelatex=false, width=0.9\textwidth]{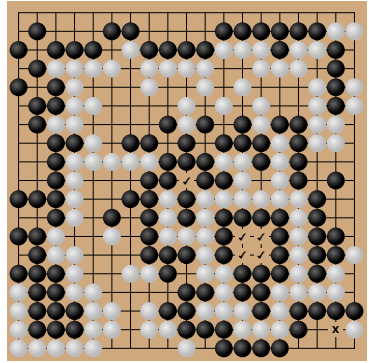}
        
	\caption{White to play.}
    \end{subfigure}
    \quad
    \begin{subfigure}{0.31\textwidth}
        \centering
        \includesvg[inkscapelatex=false, width=0.9\textwidth]{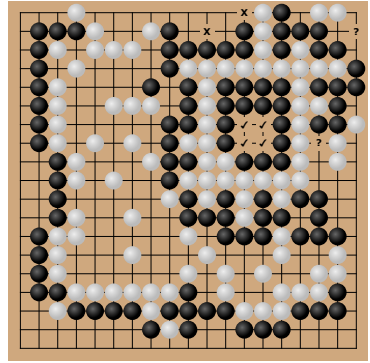}
	\caption{White to play.}
    \end{subfigure}
    \\
    \vspace{2mm}
    \begin{subfigure}{0.31\textwidth}
        \centering
        \includesvg[inkscapelatex=false, width=0.9\textwidth]{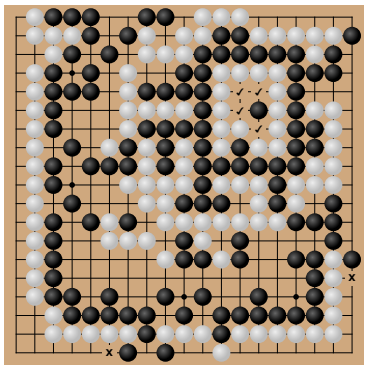}
	\caption{Black to play.}
    \end{subfigure}
    \quad
    \begin{subfigure}{0.31\textwidth}
        \centering
        \includesvg[inkscapelatex=false, width=0.9\textwidth]{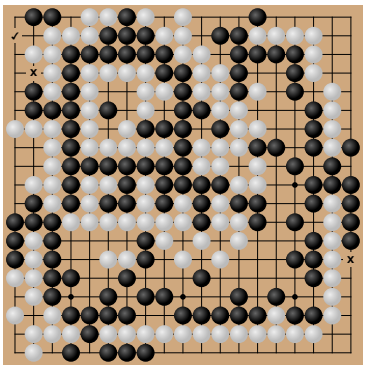}
	\caption{Black to play.}
    \end{subfigure}
    \quad
    \begin{subfigure}{0.31\textwidth}
        \centering
        \includesvg[inkscapelatex=false, width=0.9\textwidth]{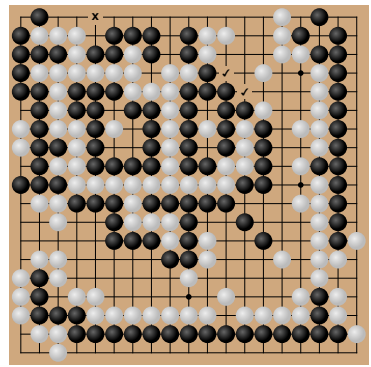}
	\caption{White to play.}
    \end{subfigure}
    
	\caption{Positions that are not solved with even 1 billion visits.
	}
    \label{fig:search-position-examples-1b}
\end{figure}

\clearpage
\section{\textSwitch{A-MCTS-R analysis}{A-MCTS-R Analysis}}
\label{app:recursive}

In this section we further examine the effect of \ourmctsperfectabbrev{} to more accurately model the victim compared to \ourmctssamplesymmetryabbrev{} by simulating the victim's search as well as its policy.
We find \ourmctsperfectabbrev{} yields modest but consistent improvements, even without simulating the entirety of victim search.

\paragraph{Position Analysis} We examined 50 games the \ourmctssamplesymmetryabbrev{} adversary played against a victim with $10^6$ visits/move, looking for cases the adversary lost but had a winning advantage. Out of these 50 games, the adversary lost 9, and a Go expert (Kellin Pelrine) found in 5 of those it had a winning position (i.e.\ with optimal subsequent play from both sides, the adversary would be guaranteed to win). These are positions where a cyclic group was created and trapped, but then at some point the adversary made a mistake that let the victim break the encirclement and escape.

We took the last position where the adversary still had a guaranteed win and examined if \ourmctsperfectabbrev{} would secure this win where \ourmctssamplesymmetryabbrev{} had failed to. The positions are shown in Figure~\ref{fig:app:amctsr-positions}. We simulated the victim with $1, 10, 10^2, 10^3, 10^4, 10^5$ visits in \ourmctsperfectabbrev{} and checked if each level of simulated victim search would lead to a correct move played by the adversary. In all cases, including the original games, the adversary uses 600 visits.

Results are shown in Table~\ref{tab:app:amctsr-pos-analysis}. The $10^0$ case means the adversary simulated the victim without search by sampling directly from the policy head, i.e. using \ourmctssampleabbrev{}.
In these cases the adversary makes a game-losing move, replicating the results of the original games.
By contrast, the adversary playing with \ourmctsperfectabbrev{} eventually find the correct move in each position.
The number of simulated victim visits needed varies from only $10$ to $10^5$.
The results here are monotonic, i.e., if the correct move is found with one number of simulated victim visits then it is maintained with a higher number.

Considering these games represent 5 out of the 9 total losses in 50 games against this victim, these results suggest \ourmctsperfectabbrev{} can produce a solid increase in win rate. While fully simulating a high-search victim is prohibitively expensive computationally, the examples here show a benefit can be gained against a high-search victim even with a low-search simulated one. In such regimes, this low-search simulated victim does not substantially increase the computation needed -- for example, against a victim with $10^6$ visits and an adversary with 600 visits, simulating 100 victim visits would lead to approximately $600 * 100 = 60,000$ additional visits per move, which is small compared to the $10^6$ victim visits. Consequently, this might improve adversary training, and even in evaluation alone can provide a stronger challenge for robustness. We therefore recommend this strategy for future work with high-visit victim systems.

\begin{table}[h]
    \centering
    \begin{tabular}{llllll}
         \toprule
         \textbf{Visits} & \textbf{Game 0} & \textbf{Game 1} & \textbf{Game 2} & \textbf{Game 3} & \textbf{Game 4}  \\
         \midrule
         \textbf{1} & \nosymbol & \nosymbol & \nosymbol & \nosymbol  & \nosymbol  \\
         \textbf{10} & \yessymbol & \nosymbol & \yessymbol &  \nosymbol  & \nosymbol  \\
         \textbf{100} & \yessymbol & \nosymbol & \yessymbol & \yessymbol & \nosymbol  \\
         \textbf{1,000} & \yessymbol & \nosymbol & \yessymbol & \yessymbol & \nosymbol  \\
         \textbf{10,000} & \yessymbol & \nosymbol & \yessymbol & \yessymbol & \yessymbol  \\
         \textbf{100,000} & \yessymbol & \yessymbol & \yessymbol & \yessymbol & \yessymbol  \\
         \bottomrule
    \end{tabular}
    \caption{
	    Varying the number of victim visits simulated in the recursive part of \ourmctsperfectabbrev{}, in 5 positions where the adversary blundered (\nosymbol). In each case, with enough simulated visits the blunder is avoided (\yessymbol), and in several cases even a small number is sufficient.
    }
    \label{tab:app:amctsr-pos-analysis}
\end{table}

\begin{figure}
    \centering
    \input{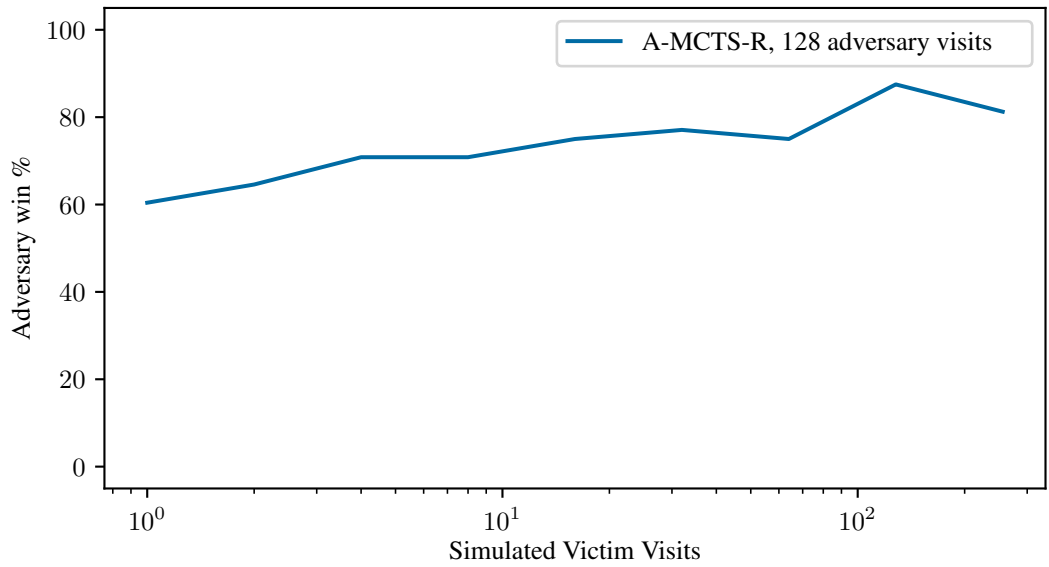}
    \caption{Testing adversary win rate with varying levels of \ourmctsperfectabbrev{} simulated victim visits. Even while well below the actual victim's 8K visits, \ourmctsperfectabbrev{} can provide some improvement.}
    \label{fig:amctsr-against-8kv}
\end{figure}

\begin{figure}
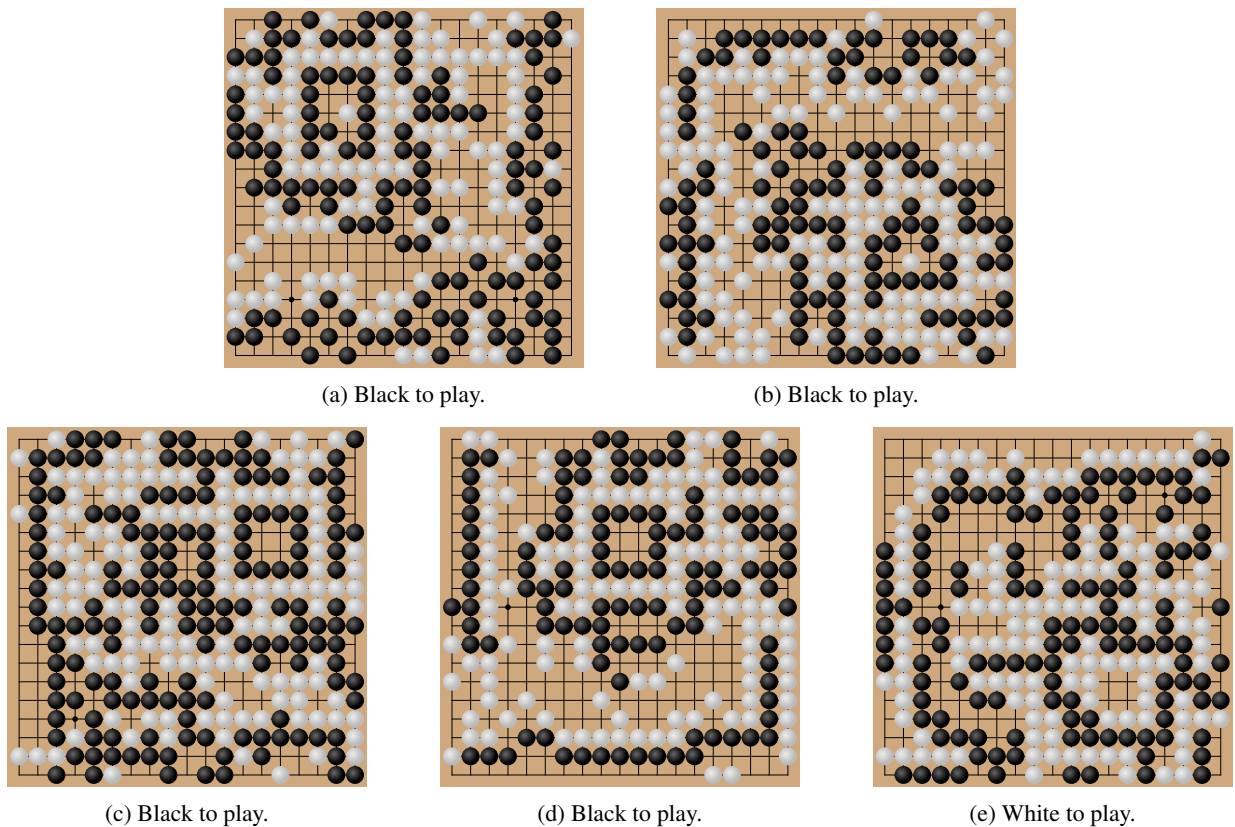

    \centering
    \begin{subfigure}{0.31\textwidth}
        \centering
        \includesvg[inkscapelatex=false, width=0.9\textwidth]{figs/boardstates/recursive-positions/pos1-amctsr-2col.svg}
	\caption{Black to play.}
    \end{subfigure}
    \quad
    \begin{subfigure}{0.31\textwidth}
        \centering
        \includesvg[inkscapelatex=false, width=0.9\textwidth]{figs/boardstates/recursive-positions/pos2-amctsr-2col.svg}
	\caption{Black to play.}
    \end{subfigure}
    \\
    \vspace{2mm}
    \begin{subfigure}{0.31\textwidth}
        \centering
        \includesvg[inkscapelatex=false, width=0.9\textwidth]{figs/boardstates/recursive-positions/pos3-amctsr-2col.svg}
	\caption{Black to play.}
    \end{subfigure}
    \quad
    \begin{subfigure}{0.31\textwidth}
        \centering
        \includesvg[inkscapelatex=false, width=0.9\textwidth]{figs/boardstates/recursive-positions/pos4-amctsr-2col.svg}
	\caption{Black to play.}
    \end{subfigure}
    \quad
    \begin{subfigure}{0.31\textwidth}
        \centering
        \includesvg[inkscapelatex=false, width=0.9\textwidth]{figs/boardstates/recursive-positions/pos5-amctsr-2col.svg}
	\caption{White to play.}
    \end{subfigure}
    
	\caption{Positions where the \ourmctssamplesymmetryabbrev{} adversary blundered. We found \ourmctsperfectabbrev{} does better. 
	}
    \label{fig:app:amctsr-positions}
\end{figure}

\paragraph{Match Analysis} A potential limitation of the analysis above is that it only considers specific positions rather than full games. In Figure~\ref{fig:amctsr-against-8kv}, we show results of our adversary with 503 million training steps and 128 visits against a victim with 8192 visits. This is a weaker version of our adversary compared to the 545 million training steps with 600 visits that we use in the main experiments. The adversary here has an \ourmctssamplesymmetryabbrev{} win rate well below 100\%, so we can look at the impact of varying the \ourmctsperfectabbrev{} simulated victim visits. Each data point is the result of 48 games.

We see that performance trends upwards, providing additional evidence that a low amount of recursive victim simulations can improve the winrate against a significantly higher visit real victim. We note that the results here are not perfectly monotonic, but this is likely due to the limited sample size. In future work, we plan to run a similar experiment with more samples against a $10^6$ visit victim to further confirm these results.

\clearpage
\section{\textSwitch{Human experiments and analysis}{Human Experiments and Analysis}}
\label{app:human-experiments}

\subsection{\textSwitch{Humans vs. adversarial policies}{Humans vs. Adversarial Policies}}
\label{app:experiments:human-vs-adversary}

\begin{figure}
    \centering
    \begin{subfigure}[b]{0.48\textwidth}
        \centering
        \includesvg[width=0.98\textwidth]{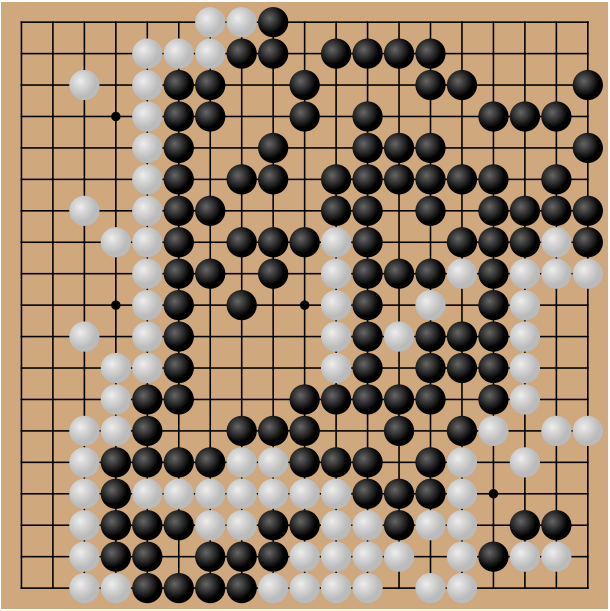}
    	\caption{An author (B) defeats the strongest cyclic-adversary from Figure~\ref{fig:evaluation:training-curve-cyclic} by 36.5 points. \href{\demosite//human-evaluation?row=1\#amateur_vs_advh_545mil}{Explore the game}.}
	\end{subfigure}
	\hfill
	\begin{subfigure}[b]{0.48\textwidth}
        \centering
        \includesvg[width=0.98\textwidth]{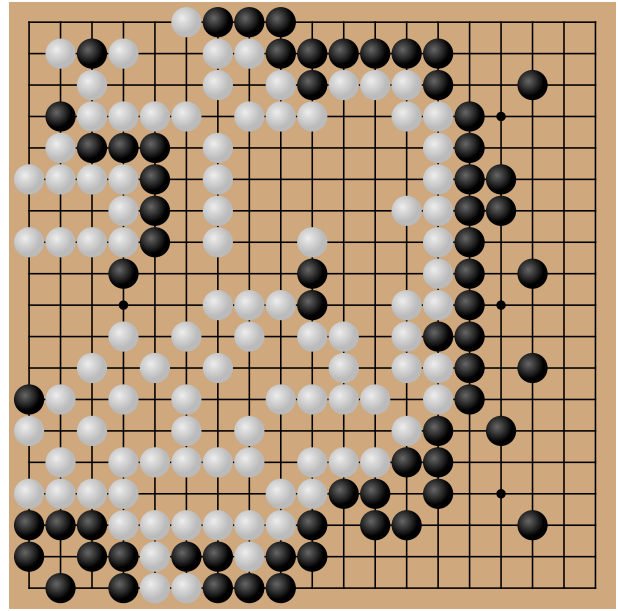}
    	\caption{An author (W) defeats the strongest cyclic-adversary from Figure~\ref{fig:evaluation:training-curve-cyclic} by 65.5 points. \href{\demosite//human-evaluation?row=0\#amateur_vs_advh_545mil}{Explore the game}.}
	\end{subfigure}
    \\[5mm]
    \begin{subfigure}[b]{0.48\textwidth}
        \centering
        \includesvg[width=0.98\textwidth]{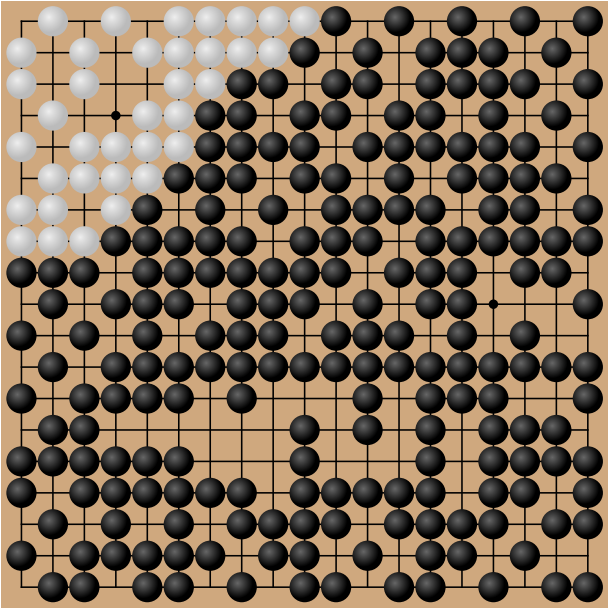}
	\caption{An author (B) defeats the strongest pass-adversary from Figure~\ref{fig:evaluation:training-curve-no-search}. \href{\demosite//human-evaluation?row=1\#amateur_vs_adv}{Explore the game}.}
    \end{subfigure}
    \hfill
    \begin{subfigure}[b]{0.48\textwidth}
        \centering
        \includesvg[width=0.98\textwidth]{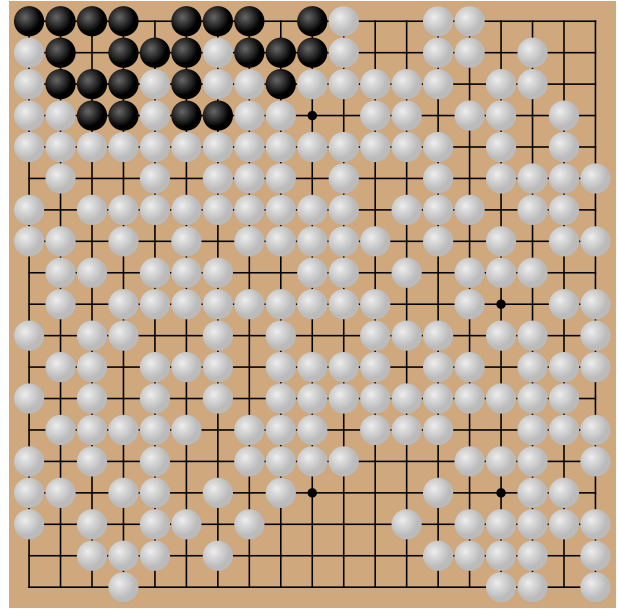}
    	\caption{An author (W) defeats the strongest pass-adversary from Figure~\ref{fig:evaluation:training-curve-no-search} using \ourmctssamplesymmetryabbrev{}. \href{\demosite//human-evaluation?row=0\#amateur_vs_adv}{Explore the game}.}
	\end{subfigure}
 
    \vspace{0.2cm}
	\caption{Games between an author of this paper (who is a Go amateur) and the strongest adversaries from
	Figure~\ref{fig:evaluation:training-curve-cyclic} and Figure~\ref{fig:evaluation:training-curve-no-search}. In all games, the author achieves a victory. The adversaries used 600 playouts / move and used \cpfivezerofive{} as the model of its human opponent. The adversaries used \ourmctssampleabbrev{} for all games except the one marked otherwise.}
    \label{fig:app:tony-vs-adversary}
\end{figure}

An author who is a Go novice played manual games against
both the strongest cyclic-adversary from Figure~\ref{fig:evaluation:training-curve-cyclic}
and the strongest pass-adversary from Figure~\ref{fig:evaluation:training-curve-no-search}.
In the games against the pass-adversary, the author was able to achieve an overwhelming victory.
In the games against the cyclic-adversary, the author won but with a much smaller margin.
See Figure~\ref{fig:app:tony-vs-adversary} for details.

We also set up a bot running our strongest cyclic-adversary on the KGS Go server under the username \href{https://www.gokgs.com/graphPage.jsp?user=Adversary0}{\texttt{Adversary0}}. This bot was available for the public to play for a period of a month. It played over 2300 games and was ranked around 17kyu. This is further evidence that our cyclic-adversary plays at the level of a novice Go player.

Our evaluation is imperfect in one significant way: the adversaries did not play with an accurate model of their human opponents (rather they modeled their opponents as \cpfivezerofive{} with 1 visit).
However, given the limited transferability of our adversaries to different KataGo checkpoints (see Figure~\ref{fig:evaluation:training-curve-cyclic}, Figure~\ref{fig:evaluation:training-curve-no-search}, Appendix~\ref{app:checkpoint-transfer}, and Appendix~\ref{app:adv-training:adv-fine-tuning}), 
we conjecture that our adversaries would not win significantly more even if they had access to an accurate model of their human opponents.

\clearpage
\subsection{\textSwitch{Human analysis of the cyclic-adversary}{Human Analysis of the Cyclic-Adversary}}
\label{app:kellin-donut-analysis}

In the following we present human analysis of games with the cyclic-adversary (the type shown in Figure~\ref{fig:cp505vis1-board:hardened}) playing against \hardened{\cpfivezerofive{}} with 1600 visits. This analysis was done by an expert-level Go player on our team. We first analyze in detail a game where the adversary won. We then summarize a sample of games where the adversary lost.

\begin{figure}
    \centering
    \begin{subfigure}[t]{0.48\textwidth}
        \centering
        \includesvg[inkscapelatex=false, width=0.9\textwidth]{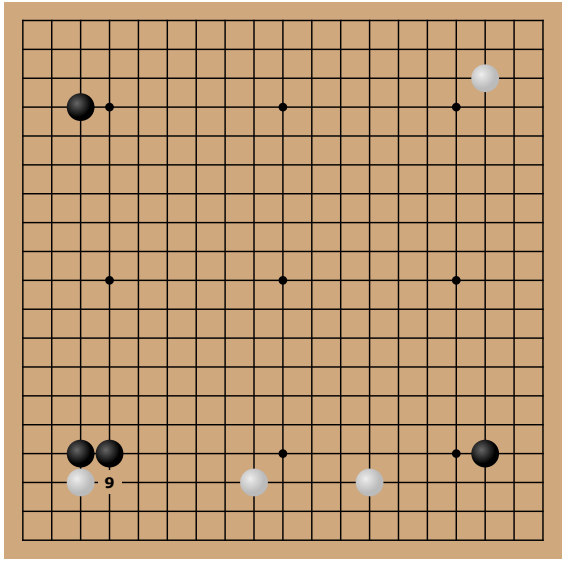}
	    \caption{Move 9: after this move victim already has the advantage, if it were robust.}
        \label{fig:human_analysis_of_dragonslayer:move_9}
    \end{subfigure}
    \quad
    \begin{subfigure}[t]{0.48\textwidth}
        \centering
        \includesvg[inkscapelatex=false, width=0.9\textwidth]{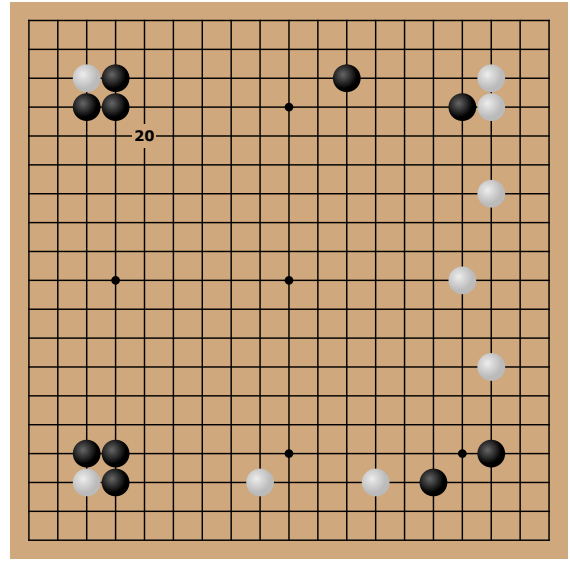}
	    \caption{Move 20: adversary initiates a key tactic to create a cycle group.}
        \label{fig:human_analysis_of_dragonslayer:move_20}
    \end{subfigure}
    \\
    \vspace{2mm}
    \begin{subfigure}[t]{0.48\textwidth}
        \centering
        \includesvg[inkscapelatex=false, width=0.9\textwidth]{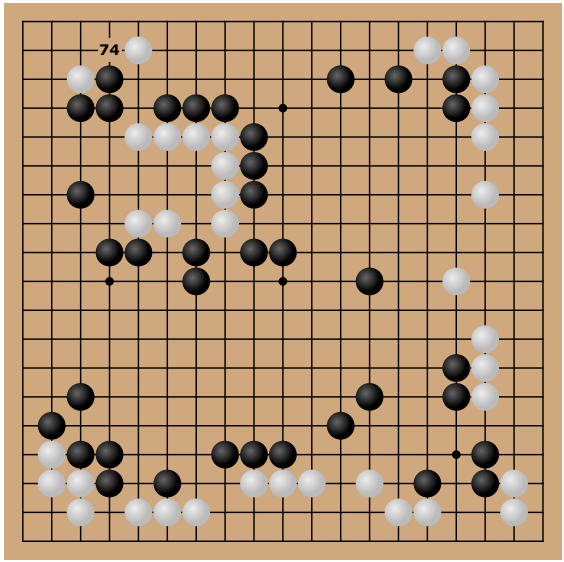}
	    \caption{Move 74: adversary slowly begins to surround victim.\\}
        \label{fig:human_analysis_of_dragonslayer:move_74}
    \end{subfigure}
    \quad
    \begin{subfigure}[t]{0.48\textwidth}
        \centering
        \includesvg[inkscapelatex=false, width=0.9\textwidth]{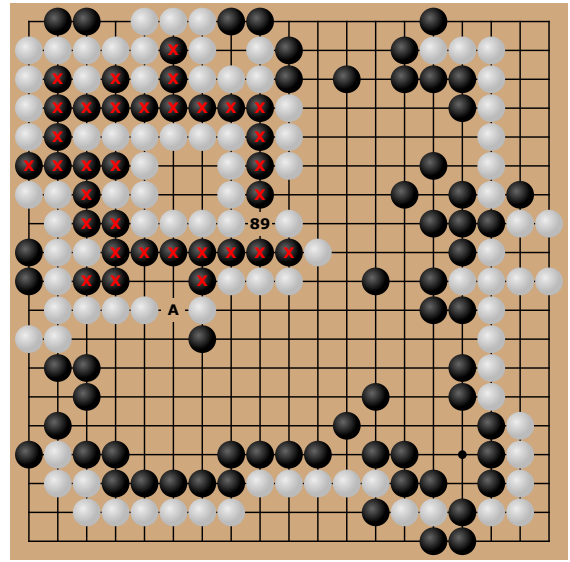}
	    \caption{Move 189 (89): victim could have saved \textcolor{red}{X} group by playing at "A" instead, but now it will be captured.}
        \label{fig:human_analysis_of_dragonslayer:move_189}
    \end{subfigure}
	\caption{The cyclic-adversary (white) exploiting a KataGo victim (black) by capturing a large group that a human could easily save. The subfigures show different moves in the game. Explore \href{\demosite/game-analysis\#qualitative}{the full game}.}
    \label{fig:human_analysis_of_dragonslayer}
\end{figure}

\paragraph{Adversary win analysis} The game in Figure~\ref{fig:human_analysis_of_dragonslayer} shows typical behavior and outcomes with this adversary: the victim gains an early and soon seemingly insurmountable lead. The adversary sets a trap that would be easy for a human to see and avoid. But the victim is oblivious and collapses.

In this game the victim plays black and the adversary white. The full game is available on our \href{\demosite/game-analysis#qualitative}{website}. We see in Figure~\ref{fig:human_analysis_of_dragonslayer:move_9} that the adversary plays non-standard, subpar moves right from the beginning. The victim's estimate of its win rate is over 90\% before move 10, and a human in a high-level match would likewise hold a large advantage from this position.

On move 20 (Figure~\ref{fig:human_analysis_of_dragonslayer:move_20}), the adversary initiates a tactic we see consistently, to produce a "dead" (at least, according to normal judgment) square 4 group in one quadrant of the board. Elsewhere, the adversary plays low, mostly second and third line moves. This is also common in its games, and leads to the victim turning the rest of the center into its sphere of influence. We suspect this helps the adversary later play moves in that area without the victim responding directly, because the victim is already strong in that area and feels confident ignoring a number of moves.

On move 74 (Figure~\ref{fig:human_analysis_of_dragonslayer:move_74}), the adversary begins mobilizing its "dead" stones to set up an encirclement. Over the next 100+ moves, it gradually surrounds the victim in the top left. A key pattern here is that it leads the victim into forming an isolated group that loops around and connects to itself (a group with a cycle instead of tree structure). David Wu, creator of KataGo~\citep{wu2019}, suggested Go-playing agents like the victim struggle to accurately judge the status of such groups, but they are normally very rare. This adversary seems to produce them consistently.

Until the adversary plays move 189 (Figure~\ref{fig:human_analysis_of_dragonslayer:move_189}), the victim could still save that cycle group (marked with \textcolor{red}{X}), and in turn still win by a huge margin. There are straightforward moves to do so that would be trivial to find for any human playing at the victim's normal level. Even a human who has only played for a few months or less might find them. For instance, on 189 it could have instead played at the place marked "A." But after 189, it is impossible to escape, and the game is reversed. The victim seems to have been unable to detect the danger. Play continues for another 109 moves but there is no chance for the victim (nor would there be for a human player) to get out of the massive deficit it was tricked into.

\paragraph{Adversary loss analysis} In all cases examined where the adversary lost, it did set up a cycle group, or a cycle group with one stone missing, which is likely still a cycle as perceived by the neural net of the victim (see Figure~\ref{fig:human_analysis_of_dragonslayer:move_189} for an example where it misjudges such a position).

\begin{figure}[th]
    \centering
        \centering
        \includesvg[inkscapelatex=false, width=0.48\textwidth]{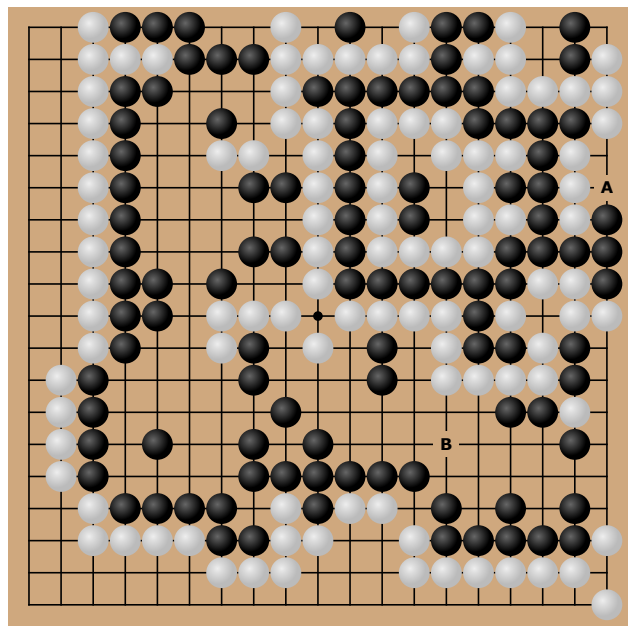}
	\caption{A game the cyclic-adversary (white) lost. The adversary could take a decisive lead by capturing at A, but instead plays B and lets the victim (black) save their group.}
    \label{fig:adv-lost-but-could-win-example}
    
\end{figure}

In four out of ten cases, the adversary could either immediately capture the cycle group or could capture it on its next turn if it played correctly. An example is shown in Figure~\ref{fig:adv-lost-but-could-win-example}. But instead it allowed the victim to save the group and win the game. We found this is due in some situations to imperfect modeling of the victim, i.e., modeling a victim without search in \ourmctssamplesymmetryabbrev{} even though the true victim has search. This can lead to the adversary thinking the victim will not defend, and therefore there is no need to capture immediately, while in reality the victim is about to defend and the opportunity will disappear. In such cases, \ourmctsperfectabbrev{} leads to the correct move. Besides this search limitation, other contributing factors potentially include the adversary itself not being completely immune to misjudging cycle groups, or the adversary's skill at Go in general being too low, resulting in many mistakes.

In the other six cases the adversary never has any clear opportunity to capture the cycle group. This is because the victim breaks through the attempted encirclement in some fashion, either by capturing some surrounding stones or simply connecting to one of its other groups. Although this could indicate the victim recognized the danger to the cycle group, the moves are typically also consistent with generic plays to wrap up the game with the large lead that it has established.

\clearpage
\subsection{\textSwitch{Human analysis of the adversary's training progression}{Human Analysis of the Adversary's Training Progression}}
\label{app:training-games-analysis}
To understand the evolution of our adversary over the training process, we randomly sampled 5 games against each of 4 victims at approximately 10\% training step intervals up to 545 million steps. The 4 victims were \hardened{\cpthirtynine} with no search, \hardened{\cponetwentyseven} with no search, \hardened{\cpfivezerofive{}} with no search, and \hardened{\cpfivezerofive{}} with 4096 visits. These correspond to the victims in Figure~\ref{fig:evaluation:training-curve-cyclic}. The full sampled games are available on our \href{\demosite/training-sample#contents}{website}. An expert Go player on our team analyzed the games, looking for patterns in how the adversary won (or lost).

We first note that in all cases against a defended victim, the adversary wins as a result of a key capture which its opponent somehow misjudged. In no game that we analyzed did the victim simply play too conservatively and lose without a fight.

Early in the training process, the adversary quickly achieves a good winrate against \hardened{\cpthirtynine}, but is only gradually improving against the other victims. Here the attack is not very consistent. When the adversary is successful, the exploit typically has low moves-to-capture (defined in Appendix~\ref{app:role-of-search}). Our analysis did not reveal a consistent strategy understandable by humans. Overall, these early training games suggest that without search, KataGo's early checkpoints have significant trouble accurately judging liberties in a variety of positions.

\begin{figure}
    \centering
    \begin{subfigure}{0.48\textwidth}
        \centering
        \includesvg[inkscapelatex=false, width=0.9\textwidth]{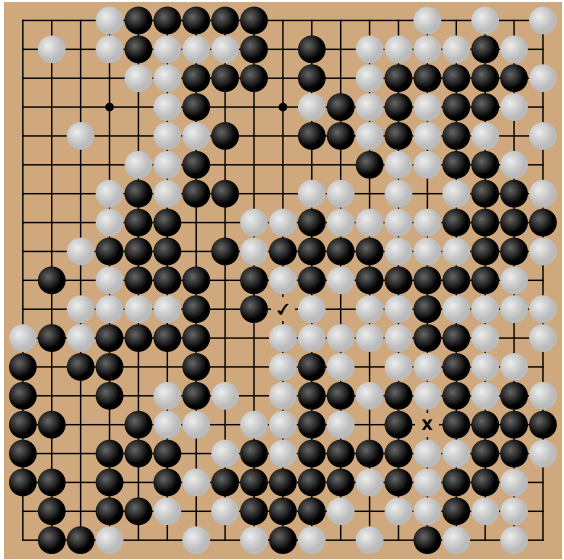}
        
	\caption{Black should have played at the location marked ``\cmark'', giving up some of their stones on the bottom in order to secure the main group. But instead black played the move marked ``\xmark'', which makes their entire right side capturable on white's next move.}
    \end{subfigure}
    \quad
    \begin{subfigure}{0.48\textwidth}
        \centering
        \includesvg[inkscapelatex=false, width=0.9\textwidth]{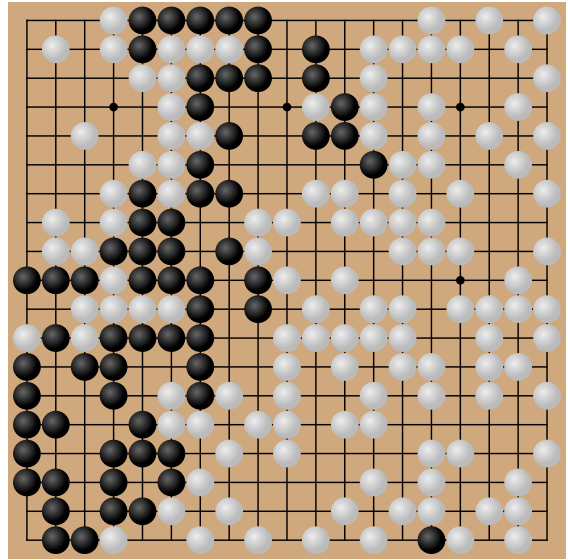}
	\caption{Both players exchange a few moves on the left side, giving the victim more chances to escape, but it doesn't see the danger. At this point White captures everything on the right, leading to victory by a large margin.}
    \end{subfigure}
	\caption{The losing move and decisive capture in a game between \hardened{\cpfivezerofive{}} (black) and the adversary with 220 million training steps (white). 
	}
    \label{fig:training-sample-game-dragon}
\end{figure}

Around 220 million training steps, the adversary is winning consistently against \hardened{\cponetwentyseven} and \hardened{\cpfivezerofive{}} with no search. We now see consistent patterns. The adversary lures the victim into creating a large group with few liberties. It sets up numerous kos and many other single stones in atari (i.e., could be captured by the victim on the next move) around that group. The victim does not realize it is about to run out of liberties, and the adversary captures the large group, leading to a win. The finishing blow is often a move where the victim fills one of their own last two liberties in order to make sure two of their groups stay connected, but this leads to losing everything. An example is shown in Figure~\ref{fig:training-sample-game-dragon}. 

This attack has similarities to the cycle attack, in that it attacks a large group that the victim seems unable to judge accurately. However, there is a far stronger emphasis on kos and stones in atari, and the moves-to-capture is generally minimal (i.e., up until the actual capture the victim could save its group, or the key portion of it). The targeted group is sometimes a cyclic group but more often not. Furthermore, unlike later versions of the cycle attack, this attack seems inadequate to beat a very moderate amount of search -- the adversary still has near 0\% winrate against \hardened{\cpfivezerofive{}} with 4096 visits.

At the same time, the win percentage against \hardened{\cpthirtynine} begins to fall. Analyzing the games, we hypothesize that this is not due to \hardened{\cpthirtynine} judging these positions more accurately, but that it more frequently plays defensive moves when far ahead. Compared to earlier attacks, this one requires giving the opponent an enormous point lead, because to set up the large target group the adversary gives up most of the board. In addition, leaving many stones in atari provides numerous straightforward defensive captures that happen to save the large group. We observe in practice that \hardened{\cpthirtynine} makes many of these captures, as well as other straightforward defensive moves, even when they are not mandatory and there are other places on the board that would give more points. At the same time, although the adversary's win percentage falls against this victim, it never goes to 0; the adversary still wins a non-trivial number of games. In combination, this suggests that \hardened{\cpthirtynine} does not clearly see the trap being created or intentionally defend against it, but rather just plays many defensive moves that often happen to save it. This has similarities to the behavior of human beginners, who often make many random defensive moves because they cannot tell clearly if a defense is needed. As in \hardened{\cpthirtynine}'s games, in many cases they are unnecessary, but in some cases they avert a disaster.

Around 270 million steps and beyond, the adversary is mostly using the cycle attack, only occasionally making non-cycle attacks. It is doing very well against \hardened{\cponetwentyseven} and \hardened{\cpfivezerofive{}} without search. However, until nearly 500 million steps the adversary still struggles against opponents with search. We hypothesize the main factor is that the moves-to-capture is too low to fool a victim with this level of search - successful attacks against this victim seem to have a moves-to-capture of at least 3, while the attacks produced at this stage in the training still often have fewer (frequently due to the many kos and stones in atari, which seemed helpful for the previous attack but not always helpful here). Towards the end of the training, the adversary starts producing cycles with higher moves-to-capture, and starts to win consistently against stronger victims with more search.

\clearpage
\section{\textSwitch{Activation analysis}{Activation Analysis}}
\label{app:activations}

We examine activations in different positions and with different versions of KataGo to try to identify where and ultimately why the system misjudges cyclic positions. 

\paragraph{Setup}

We examine the activations of each main convolutional layer in the network. Specifically, using the TensorFlow version of KataGo (before KataGo switched to using PyTorch in KataGo version 1.12) we look at ``conv1'' for the input layer, then the equivalent ``rconvK'' in subsequent layers (where K ranges from 2 to 41), then the final ``trunk'' layer. These layers are described in detail in Appendix A of the KataGo paper~\citep{wu2019}.

\begin{figure}[h]
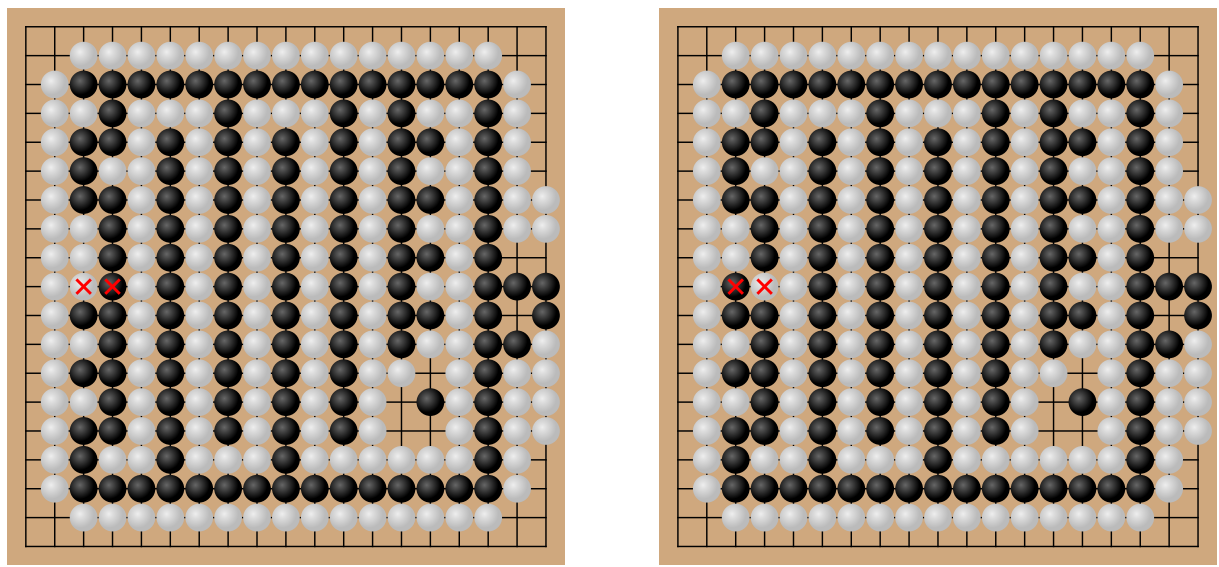

    \vspace{1.5cm}
    \centering
    \begin{subfigure}{0.48\textwidth}
        \centering
        \includesvg[inkscapelatex=false, width=0.9\textwidth]{figs/boardstates/activations/connection_test_position_A-2col.svg}
	\caption{Here the cycle is intact and complete.}
    \end{subfigure}
    \quad
    \begin{subfigure}{0.48\textwidth}
        \centering
        \includesvg[inkscapelatex=false, width=0.9\textwidth]{figs/boardstates/activations/connection_test_position_B-2col.svg}
	\caption{Here the cycle is broken.}
    \end{subfigure}
	\caption{In this manually constructed position, by swapping the colors of the stones marked \textcolor{red}{X}, we can complete or break the cyclic group. The impact on the actual game state is minimal---the score does not change, nor the life and death status or liberty counts of any stones, nor the best subsequent moves. But the two boards are evaluated dramatically differently by many KataGo checkpoints.
	}
    \label{fig:activation-pos-manual}
\end{figure}

\begin{figure}
    \centering
    \begin{subfigure}{0.48\textwidth}
        \centering
        \includesvg[inkscapelatex=false, width=0.9\textwidth]{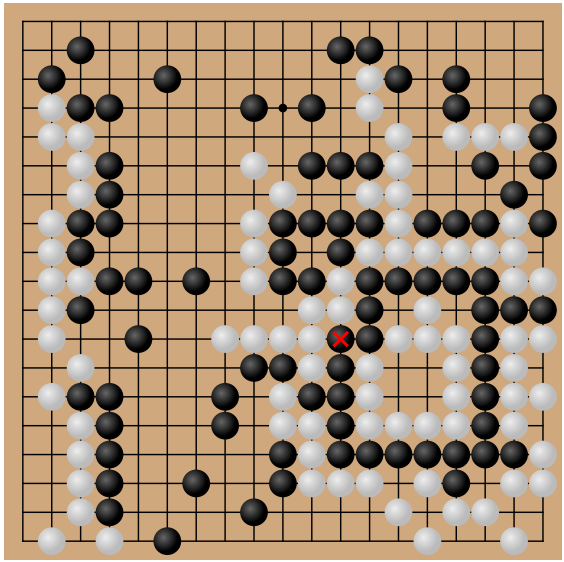}
        
	\caption{Here the cycle is intact and complete.}
        \label{fig:activation-pos-realgame-a}
    \end{subfigure}
    \quad
    \begin{subfigure}{0.48\textwidth}
        \centering
        \includesvg[inkscapelatex=false, width=0.9\textwidth]{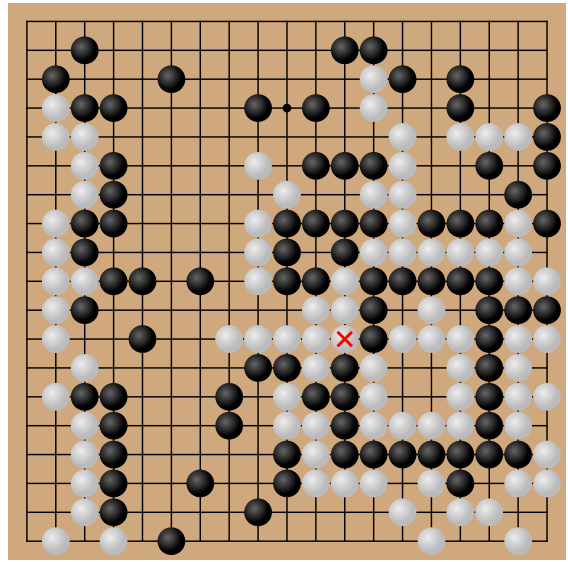}
	\caption{Here the cycle is broken.}
    \end{subfigure}
	\caption{In this real game position, by swapping the color of the stone marked \textcolor{red}{X}, we can complete or break the cyclic group. The impact on the actual game state is minimal---the score is changed by at most one point which does not affect who is winning, it does not affect the life and death status or liberty counts of any stones, and it does not change the best subsequent moves. But it dramatically changes the evaluation of many KataGo checkpoints.
	}
    \label{fig:activation-pos-realgame}
\end{figure}

\begin{figure}
    \centering
    \begin{subfigure}{0.48\textwidth}
        \centering
        \includesvg[inkscapelatex=false, width=0.9\textwidth]{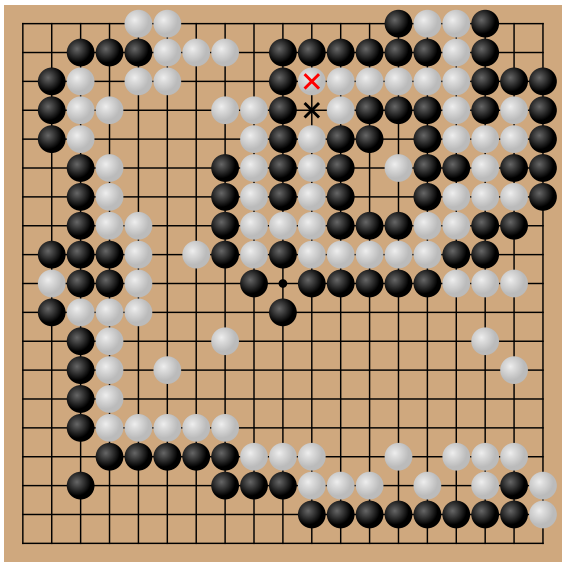}
        
	\caption{Cycle is complete.}
    \end{subfigure}
    \quad
    \begin{subfigure}{0.48\textwidth}
        \centering
        \includesvg[inkscapelatex=false, width=0.9\textwidth]{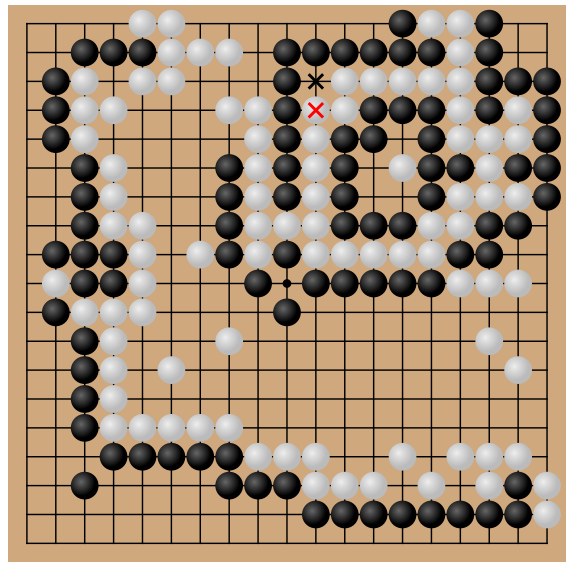}
	\caption{Cycle is incomplete, but still could be completed.}
    \end{subfigure}
	\caption{In this position, by swapping the position of the stone marked \textcolor{red}{X}, we can complete or leave open the option to complete the cycle group. As in preceding positions, the impact on the actual game state is minimal---the score does not change, it does not affect the life and death status or liberty counts of any stones, and it does not change the best subsequent moves.
	}
    \label{fig:activation-pos-realgame3}
\end{figure}

We examine positions with a cyclic group, and minimal perturbations of such a position where the cycle is broken or incomplete.
We consider a manually constructed position which has few moves available and low complexity (Figure~\ref{fig:activation-pos-manual}), an example from a real game played by an author against KataGo (Figure~\ref{fig:activation-pos-realgame}), and an example based on a game played by our adversary against KataGo (Figure~\ref{fig:activation-pos-realgame3}).
In each case a position and perturbation is chosen such that the game state is mostly unchanged (connected groups, status of all groups, best move, player in the lead, etc.) but the cycle is broken.
The perturbation between the two positions is also made minimal by either swapping the color of a single stone or a pair of adjacent stones, or moving a single stone one space over.

In the first two cases the cyclic group is dead and the victim is doomed, and in the last case it is alive and the victim currently has a winning position. In the last case, the two positions themselves are a slight modification of the real game played by KataGo and our adversary. Specifically, in the real game the victim had one of the marked stones already and played the other one that completes the cycle. Here, to avoid playing moves in the board locations we want to perturb, we have it play a nearby connecting move that does not change the game state.

\begin{figure}
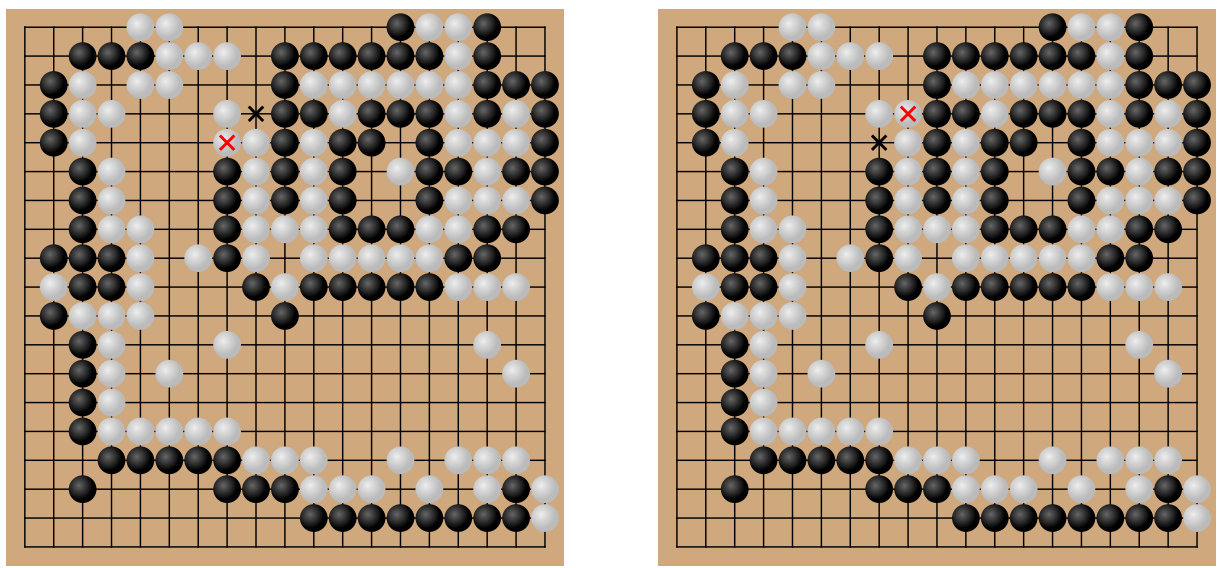

    \centering
    \begin{subfigure}{0.48\textwidth}
        \centering
        \includesvg[inkscapelatex=false, width=0.9\textwidth]{figs/boardstates/activations/connection_test_realgame3_E-2col.svg}

    \end{subfigure}
    \quad
    \begin{subfigure}{0.48\textwidth}
        \centering
        \includesvg[inkscapelatex=false, width=0.9\textwidth]{figs/boardstates/activations/connection_test_realgame3_F-2col.svg}
    \end{subfigure}
	\caption{These positions, derived from the same game as the preceding pair, have a near-cycle that is already broken. Swapping the most recent move \textcolor{red}{X} impacts neither the cyclic-ness of the position nor the game state.
	}
    \label{fig:activation-pos-realgame3-negative}
\end{figure}

Finally, we also consider a situation where the cyclic group is already broken and the perturbation is irrelevant to both the game state and the cyclic-ness. This case is derived from the preceding one (Figure~\ref{fig:activation-pos-realgame3}), and shown in Figure~\ref{fig:activation-pos-realgame3-negative}.

\clearpage
\paragraph{Progression of models}
We examine a progression of model checkpoints before and after adversarial training began, to attempt to disentangle evolution due to normal training from changes specific to adversarial training. Specifically, we look at 9 checkpoints centered on \cpfivezerofive{}, resulting in four before it and four after. Of these, the two most recent ones, \texttt{cp559} and \texttt{cp580}, have been adversarially trained on cyclic positions and demonstrate improved performance in some of them.
The specific models examined are:

\begin{enumerate}[leftmargin=40pt]
    \item Checkpoint 455: \texttt{b40c256-s10823908608-d2638763986}.
    \item Checkpoint 468: \texttt{b40c256-s11078294784-d2707780120}.
    \item Checkpoint 478: \texttt{b40c256-s11290411776-d2760978415}.
    \item Checkpoint 492: \texttt{b40c256-s11574569216-d2829125899}.
    \item \textbf{Checkpoint 505}: \texttt{b40c256-s11840935168-d2898845681} (\cpfivezerofive{}).
    \item Checkpoint 522: \texttt{b40c256-s12096598272-d2984620981}.
    \item Checkpoint 535: \texttt{b40c256-s12350780416-d3055274313}.
    \item Checkpoint 559: \texttt{b40c256-s12604774912-d3126339815} (\texttt{cp559}).
    \item Checkpoint 580: \texttt{b40c256-s12860905472-d3197353276} (\texttt{cp580}).
\end{enumerate}

These were selected to approximate equal training steps, taking into account the inexact schedule of checkpoint releases. In each case we use the network to directly evaluate the positions without tree search.

\paragraph{Results}

\begin{figure}[t]
    \centering
    \input{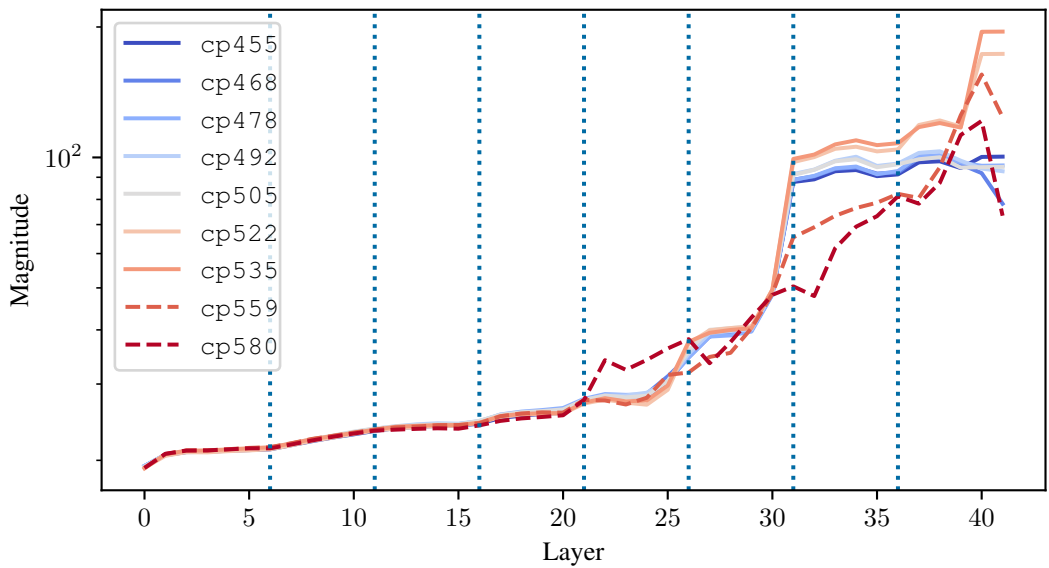}
    \caption{The magnitude of the maximum activation per layer across 9 checkpoints. The two most recent checkpoints, which have been adversarially trained, have a markedly different curve.}
    \label{fig:raw-max-activation}
\end{figure}

We begin by analyzing the progression of activation magnitudes in the position of Figure~\ref{fig:activation-pos-realgame-a}. Figure~\ref{fig:raw-max-activation} shows the maximum activation per layer. We see that the maximum grows as we progress through the network. The two most recent checkpoints (and particularly the most recent one), which have adversarial training and are shown with dashed lines, exhibit some differences from the other checkpoints. 

\begin{figure}
    \centering
    \input{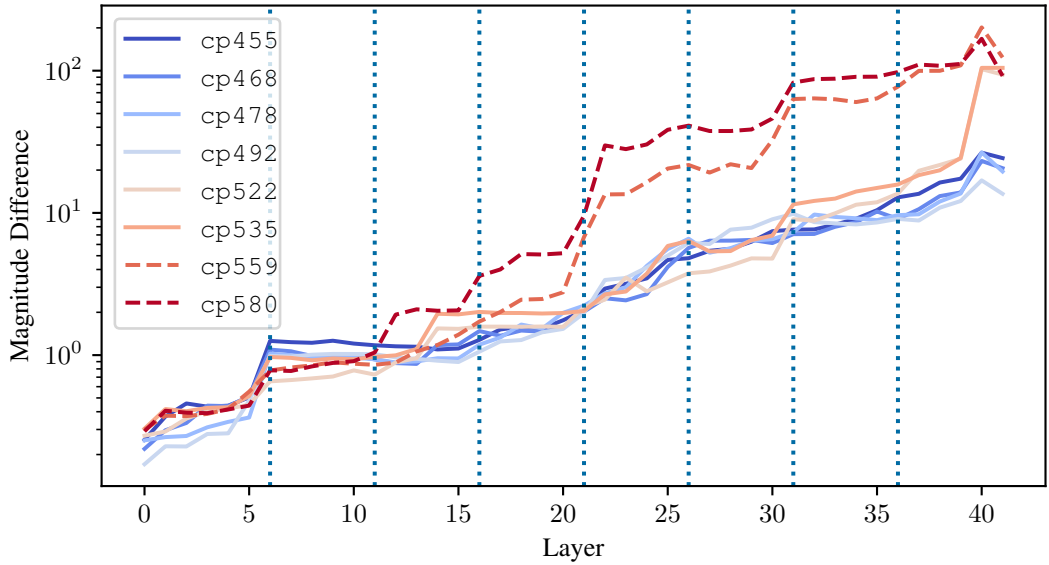}
    \caption{The magnitude of the maximum difference from \cpfivezerofive{} in activations per layer across the 8 models (excluding \cpfivezerofive{}). Again, the two checkpoints with adversarial training differ from the rest.}
    \label{fig:max-diff-activation}
\end{figure}

In Figure~\ref{fig:max-diff-activation} we instead look at the maximum per layer of the \emph{difference} between the activations of \cpfivezerofive{} and every other checkpoint. This highlights that the two most recent networks are substantially more different from \cpfivezerofive{} than previous models.
This suggests that adversarial training results in larger differences in the evaluation of these positions than a comparable number of time steps of usual training.

\begin{figure}
    \centering
    \input{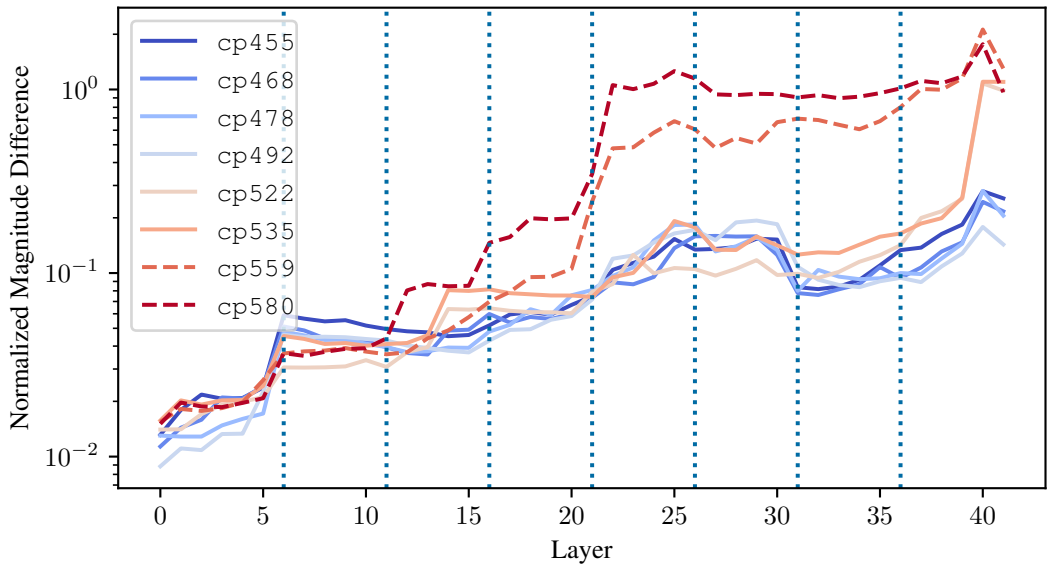}
    \caption{The magnitude of the maximum difference from \cpfivezerofive{} in activations per layer, normalized by dividing by the maximum activation per layer of \cpfivezerofive{}. We see an especially strong difference between the checkpoints with adversarial training and the rest starting around layer 21.}
    \label{fig:max-diff-normalized-activation}
\end{figure}

To more easily see which differences represent a substantial change, in Figure~\ref{fig:max-diff-normalized-activation} we normalize the differences by dividing by the maximum activation of \cpfivezerofive{} at that layer.
We see there is little difference between the models up to around layer 11.
After that, the most recent model gradually begins to break away, but this does not become pronounced until around layer 16 and especially 21--22. 
These results suggest that adversarial training may have little effect on early layers, implying that KataGo's misjudgment takes place in later layers, perhaps especially around layer 21.

\begin{figure}
    \centering
    \input{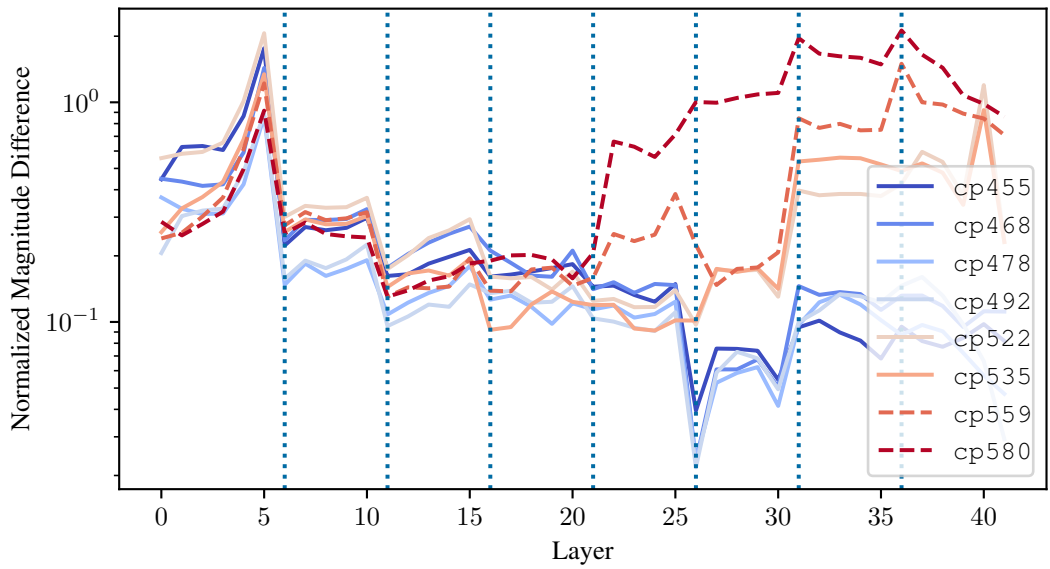}
    \caption{The magnitude of the maximum difference from \cpfivezerofive{} in \textit{channel-averaged} activations per layer, normalized by dividing by the maximum per layer of \cpfivezerofive{}. Similar to previous figures, sharp differences begin around layer 21.}
    \label{fig:diff-channelavg-max-normalized-activation}
\end{figure}

As a robustness check, in addition to analysing the activations of individual neurons we also analyse the activations of entire channels. We follow a similar process to the above, but after taking the difference in activations we then average the activations within each channel. Then we examine the max over all channels of the channel-mean activation, dividing by the maximum channel-mean activation of \cpfivezerofive{}. In Figure~\ref{fig:diff-channelavg-max-normalized-activation} we see that the checkpoints have a similar pattern up to layer 21, after which the two adversarially-trained checkpoints break away from the other models.

\begin{figure}
    \centering
    \includesvg[inkscapelatex=false, width=0.9\textwidth]{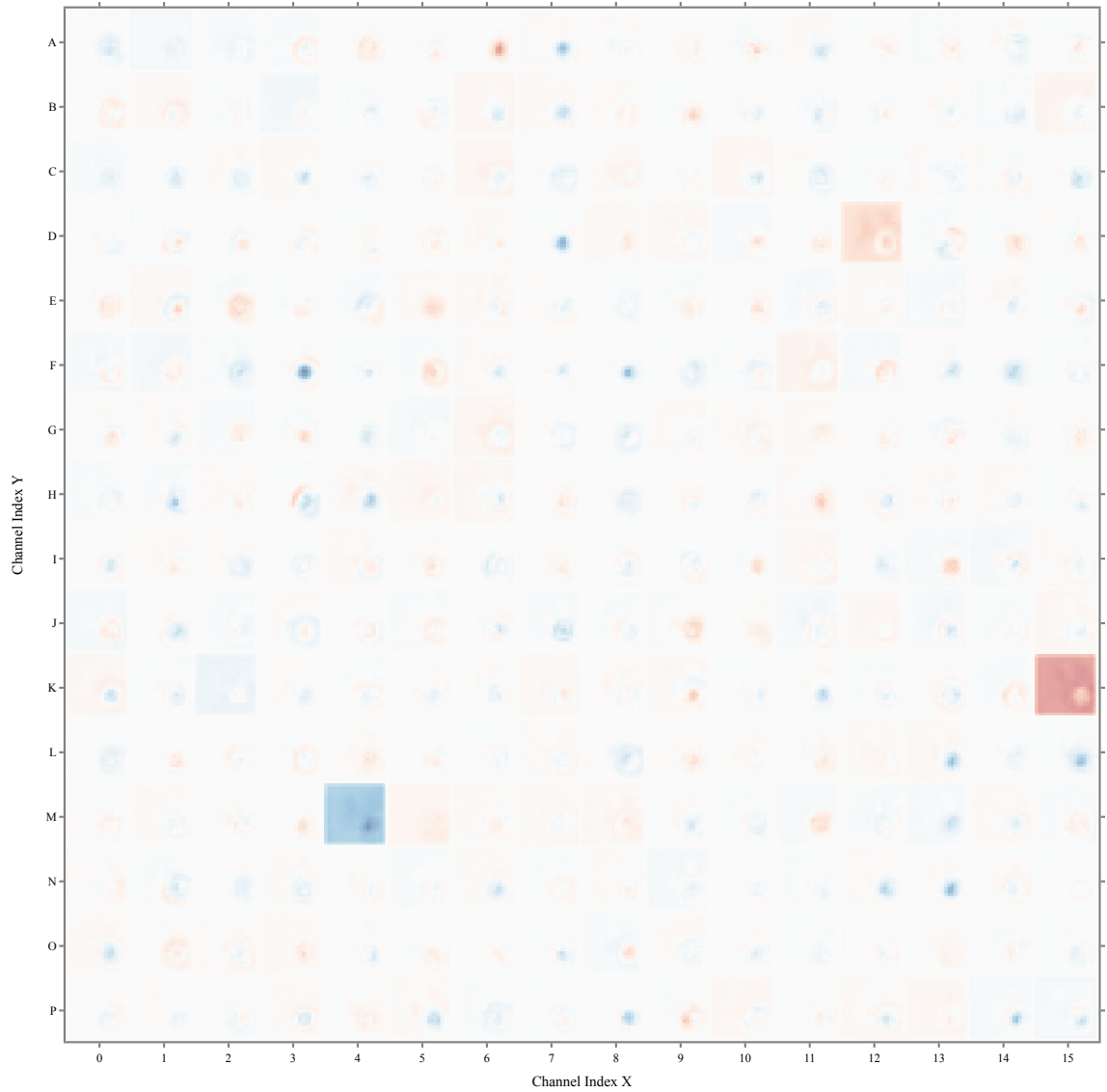}
    \caption{The difference in activations in layer 26 between \cpfivezerofive{} and \texttt{cp580} in a real game position (Figure~\ref{fig:activation-pos-realgame}).}
    \label{fig:cp505-realgame2-A-vs-b40-1286-realgame2-A-l26}
\end{figure}

So far our analysis has involved per-layer summary statistics. To gain a more granular understanding, we visualized the activations within each layer directly. In Figure~\ref{fig:cp505-realgame2-A-vs-b40-1286-realgame2-A-l26}, we plot the difference in activations between \cpfivezerofive{} and the most recent network (\texttt{cp580}) in layer 26. There are 256 channels each containing 19x19 activations, corresponding to the Go board. We pad to 20x20 to separate the channels slightly for clearer visualization. Brighter colors in the plot indicate a larger difference in activations.

We see 2-3 channels (especially the M4 and K15 channels) at layer 26 have a difference in activations across the entire board, shown in the visualization by the whole square representing that channel being brightly colored. This suggests these channels may be misjudging cyclic positions. Earlier layers do not exhibit comparable full-channel differences (there are some early layers with channels whose activations show whole board differences, but these are layers where the preceding analysis shows the overall magnitude difference is small).

\begin{figure}
    \centering
    \includesvg[inkscapelatex=false, width=0.9\textwidth]{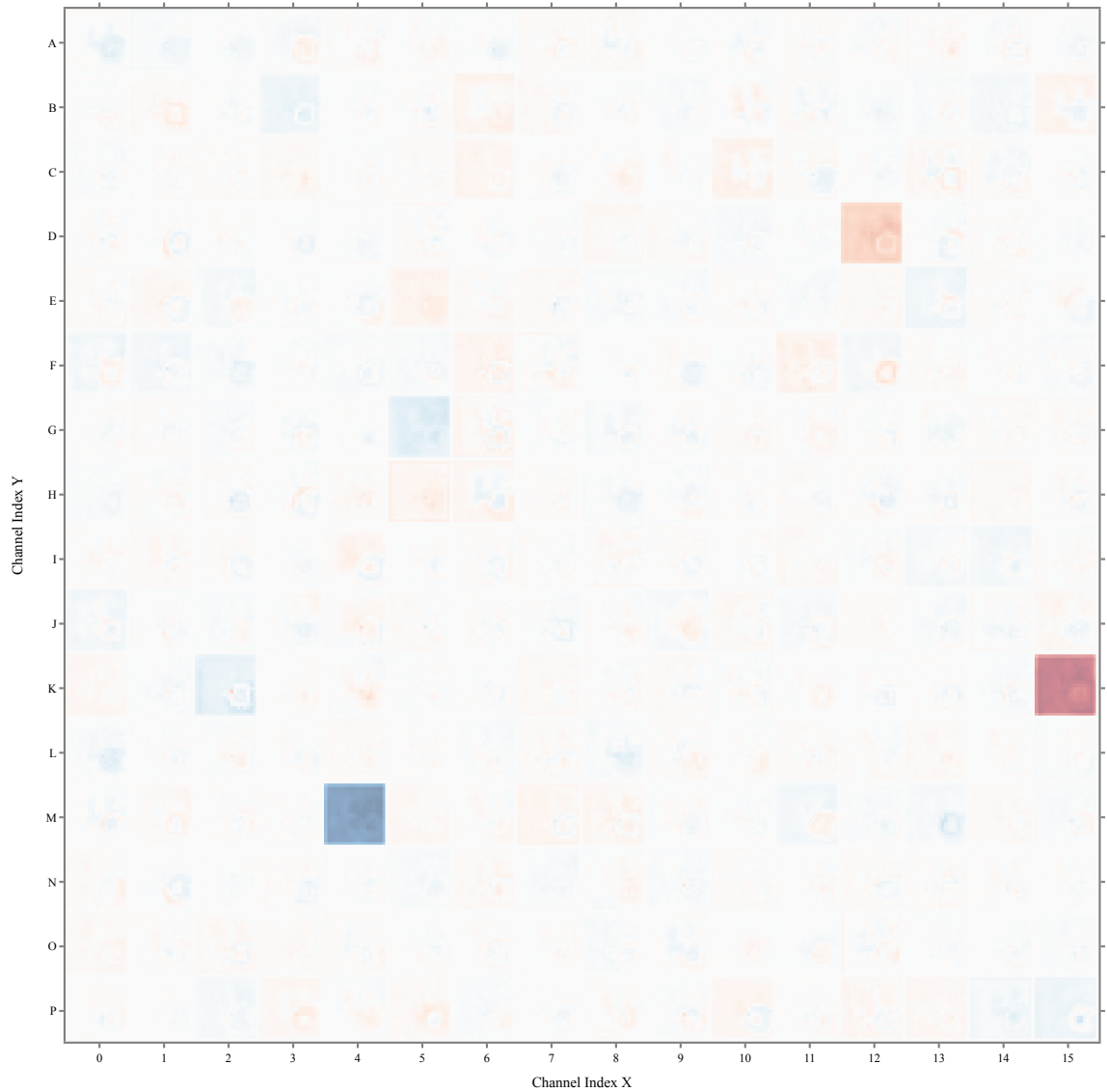}
    \caption{The difference in activations in layer 26 of \cpfivezerofive{} between a real-game cyclic position (Figure~\ref{fig:activation-pos-realgame}; the same position as the preceding Figure~\ref{fig:cp505-realgame2-A-vs-b40-1286-realgame2-A-l26}) and a minimally perturbed version of it that breaks the cycle.}
    \label{fig:cp505-realgame-AvB-l26}
\end{figure}

\begin{figure}
    \centering
    \includesvg[inkscapelatex=false, width=0.9\textwidth]{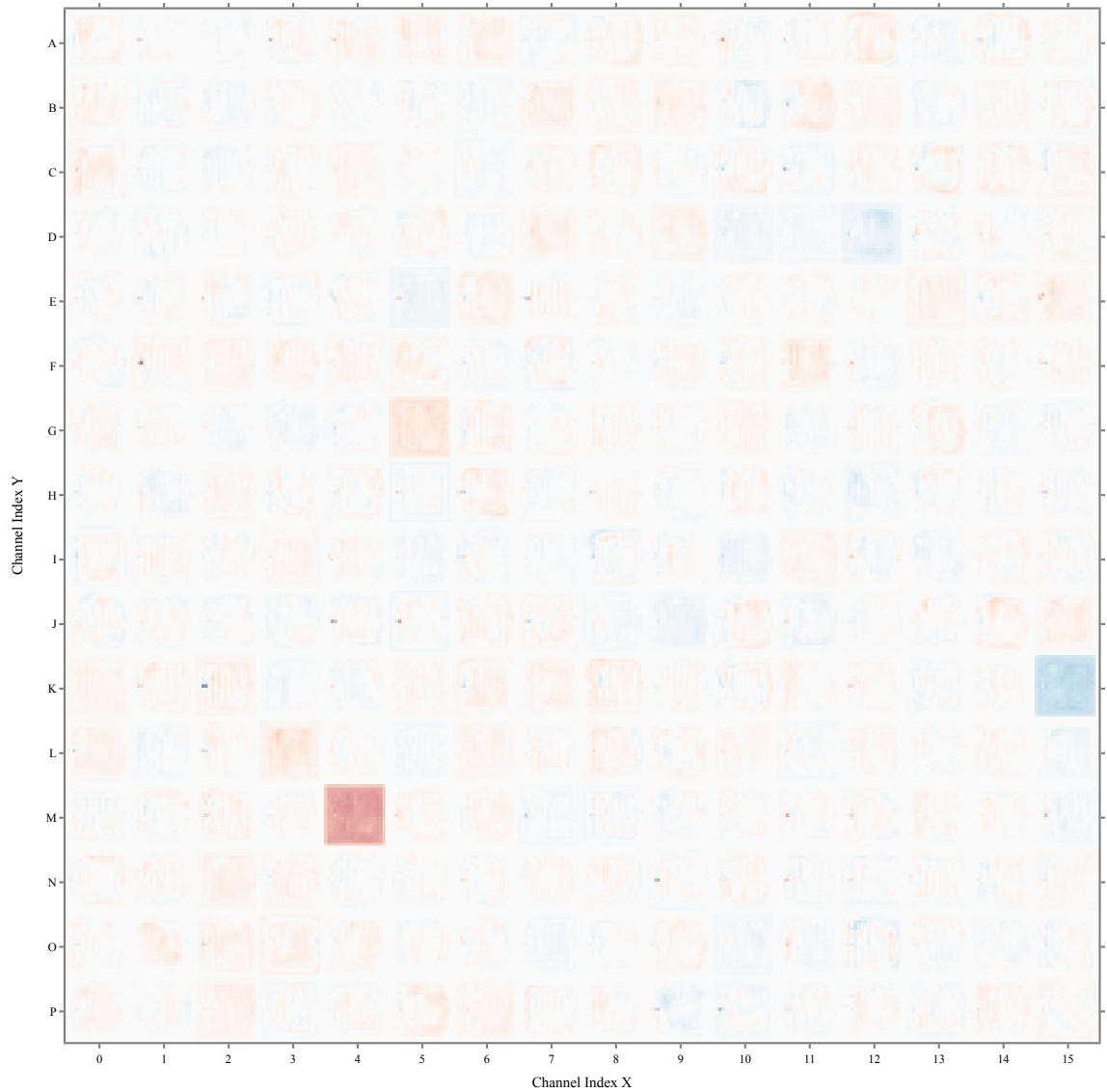}
    \caption{The difference in activations in layer 26 between \cpfivezerofive{} in manually constructed cyclic vs. non-cyclic positions (Figure~\ref{fig:activation-pos-manual}).}
    \label{fig:cp505-manual-AvB-l26}
\end{figure}

Moreover, we find a similar anomaly in layer 26 when holding a model fixed and comparing a cyclic position to the same position with the cycle broken. In Figure~\ref{fig:cp505-realgame-AvB-l26}, we compare \cpfivezerofive{}'s activations on the  the (a) and (b) positions from Figure~\ref{fig:activation-pos-realgame}, finding the same channels differ between positions as between networks. We see a similar pattern in Figure~\ref{fig:cp505-manual-AvB-l26} comparing the two manually constructed board states from Figure~\ref{fig:activation-pos-manual}. Note that the colors are reversed---in our testing, the sign of the activations corresponds to whose turn it is to play. However, the channels that differ the most remain the same.

\begin{figure}
    \centering
    \includesvg[inkscapelatex=false, width=0.9\textwidth]{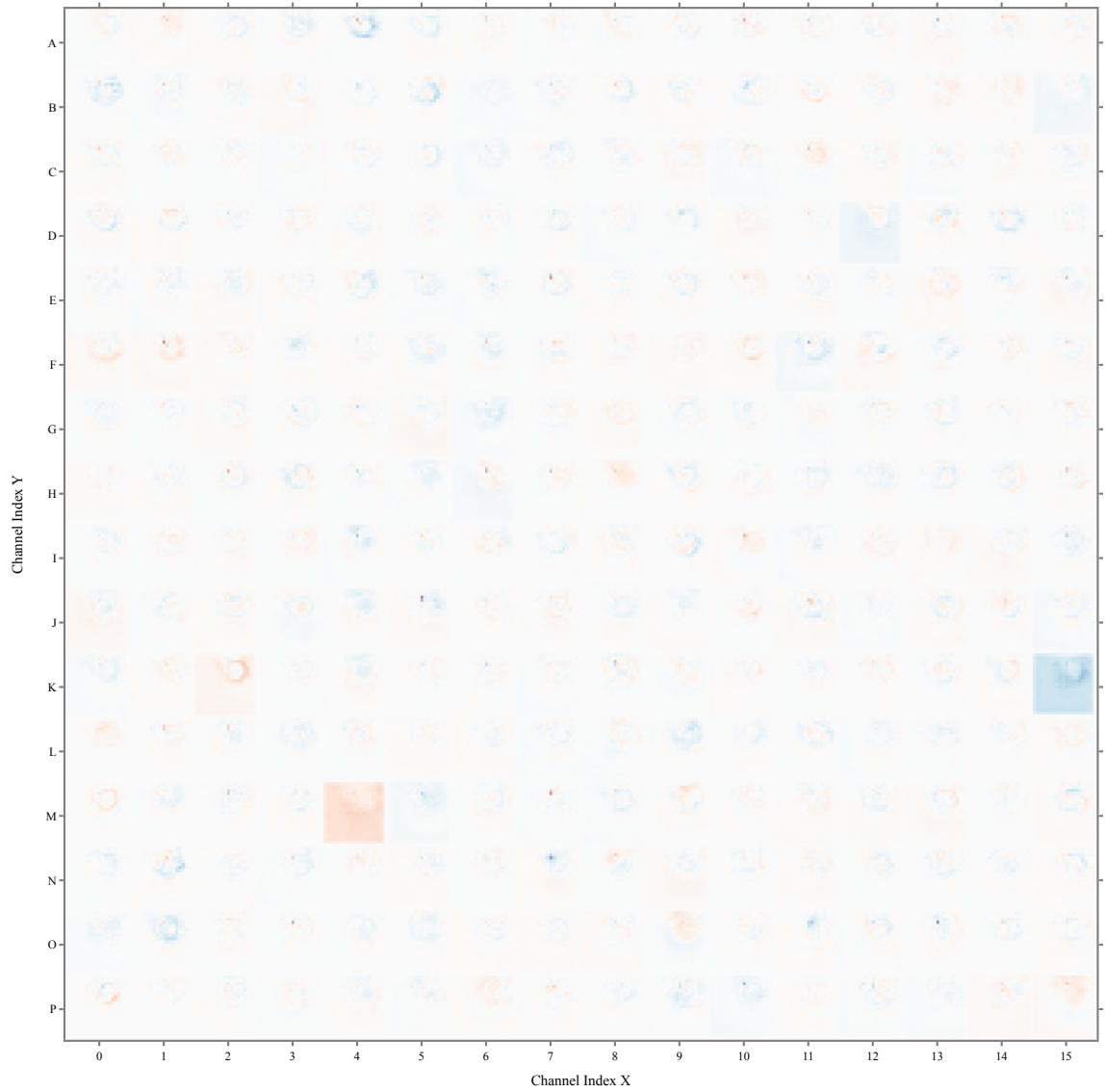}
    \caption{The difference in activations in layer 26 of \cpfivezerofive{} between a near real-game cyclic and non-cyclic position, where the cycle group is safe (Figure~\ref{fig:activation-pos-realgame3}).}
    \label{fig:cp505-realgame3-BvD-l26}
\end{figure}

\begin{figure}
    \centering
    \includesvg[inkscapelatex=false, width=0.9\textwidth]{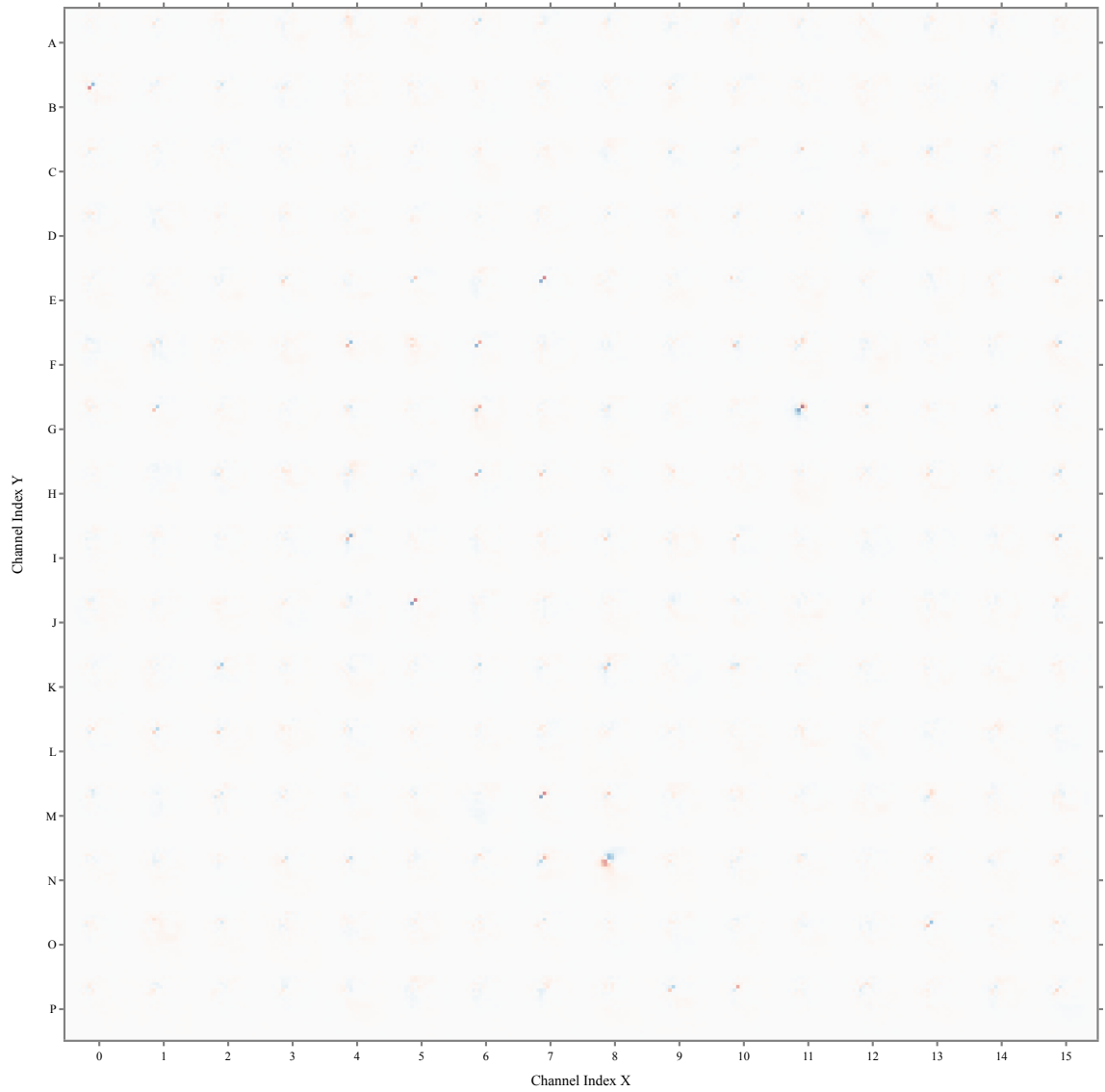}
    \caption{The difference in activations in layer 26 of \cpfivezerofive{} between two near real-game positions where the cycle is already broken and does not change (Figure~\ref{fig:activation-pos-realgame3-negative}).}
    \label{fig:cp505-realgame3-EvF-l26}
\end{figure}

We likewise see standout differences in M4 and K15 channels in the position where the cyclic group is alive (the activations shown in Figure~\ref{fig:cp505-realgame3-BvD-l26}, corresponding to the positions in Figure~\ref{fig:activation-pos-realgame3}). In this case, while still clearly visible, the channels are fainter indicating a smaller relative magnitude difference, which we hypothesize is because the group is alive independent of the illusion created by the cycle. Finally, as an additional validity check, we consider the case of Figure~\ref{fig:activation-pos-realgame3-negative}, where the cycle has been broken in both cases. In Figure~\ref{fig:cp505-realgame3-EvF-l26} we see that M4 and K15 no longer show unusual behavior, further confirming it was due to cyclic-ness and not arbitrary perturbations.

Layer 26, where we see the unusual behavior discussed above, is later than the first layers where activation magnitude substantially differs in the preceding aggregate analysis.
Accordingly, the channels at layer 26 may not be the root cause of the problem, but they do appear to be most strongly affected by it.
This suggests these stand-out channels at layer 26 are a good area in the network to examine further for a better understanding of why this vulnerability occurs, as well as potentially a place for targeted interventions aimed at improving robustness.

We include interactive plots for these positions and models, which show the activation differences of all the layers, on our \href{\demosite/activation-plots#contents}{website}.

\clearpage
\section{\textSwitch{KataGo adversarial training}{KataGo Adversarial Training}}
\label{app:adv-training}

Adversarial training against the cyclic exploit was incorporated into the official distributed training run of KataGo~\citep{katagotraining:2022} in mid-December 2022.\footnote{The latest net before the adversarial training was introduced was the b60c320 network at 6,729,327,872 steps of training, released on December 16 2022. Source: \url{https://discord.com/channels/417022162348802048/583775968804732928/1056607918545457252}.}
This adversarial training consists of starting a small fraction of self-play games in positions where the cyclic exploit is being executed, with the remainder of games being regular self-play games. 
These games were hand-selected by David Wu (the creator and primary developer of KataGo) from a \href{https://drive.google.com/drive/folders/1X3qHFbUJP7oZUHg\_Lfy-BoQTIs0hte\_m}{collection of training and evaluation games} taken from our cyclic-adversary training run (Figure~\ref{fig:evaluation:training-curve-cyclic}):

\begin{aquote}{David Wu, \href{https://discord.com/channels/417022162348802048/583775968804732928/1052951418685882408}{Computer Go Community Discord, December 15 2022}}
    I uploaded some sgfposes and hintposes about cyclic topology groups. I spent multiple hours hand-adding a bunch [of] contrastive examples, where you would have the same group with different numbers of liberties, or with different numbers of "false" (but actually real) eyes. Roughly 0.08\% of games should involve them now, we'll see if that tiny rate has any effect on learning them over the next weeks.
\end{aquote}

Figure~\ref{fig:adv-training:katago-training-win-rate} shows that over the course of 6 months, adversarial training caused the cyclic-adversary's win rate to decrease significantly 
across several KataGo architectures. Figure~\ref{fig:adv-training:katago-training-win-rate-datarows} shows that the different architectures all improved at a similar rate with respect to the total amount of data they were trained with.

In numbers:
\begin{itemize}
    \item Prior to adversarial training, our cyclic-adversary$^\text{600 visits}$ wins 1048 / 1048 games against \cpfivezerofive{}$^\text{1 visit}$, 973 / 1000 games against \cpfivezerofive{}$^\text{4096 visits}$, and 50 / 50 games against \texttt{b60-s6729m}$^\text{200 visits}$.

    \item After adversarial training, our cyclic-adversary$^\text{600 visits}$ wins 118 / 2000 games against \texttt{b60-s7702m}$^\text{1 visit}$ and 0 / 400 games against \texttt{b60-s7702m}$^\text{1600 visits}$.
    (\texttt{b60-s7702m} refers to KataGo network \texttt{b60c320-s7701878528-d3323518127}, released on \href{https://katagotraining.org/networks/}{May 17, 2023}, the latest 60-block KataGo network available at the time we ran this experiment. We round the number of training steps (7,701,878,528) to the nearest million to shorten the name.)
\end{itemize}

However, we are able to fine-tune our original adversary to defeat these updated networks (discussed below in Appendix~\ref{app:adv-training:adv-fine-tuning}). 
This suggests that it is non-trivial to defend against the cyclic exploit, unlike the pass exploit which we were able to manually patch.
Thus, developing techniques to train agents that are immune to the cyclic-exploit while maintaining high Go strength remains an interesting open problem.

\begin{figure}[t]
    \centering
    \includesvg[width=0.4\textwidth]{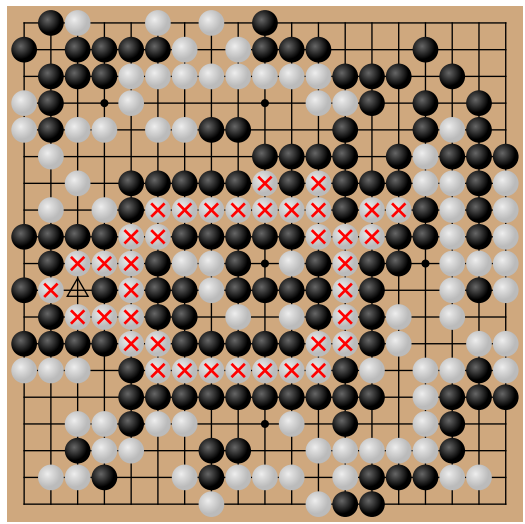}
	\caption{Our fine-tuned cyclic-adversary, playing as black,  still wins by capturing a cyclic group (\textcolor{red}{$\mathbf{\times}$}) that the victim (\texttt{b60-s770m}, $10^5$ visits, 10 search threads) leaves vulnerable. The adversary plays at the square marked $\Delta$ to capture the group. \href{\demosite/adversarial-training?row=0\#cyclic_fine_tune_vs_b60_s7702m_v100k-board}{Explore the game}.}
    \label{fig:app:cyclic-fine-tune-vs-b60-s7702m-v100k}
\end{figure}

Figure~\ref{fig:app:cyclic-fine-tune-vs-b60-s7702m-v100k} displays one game of the fine-tuned cyclic-adversary against $\texttt{b60-s7702m}^\text{100,000 visits}$.
The attack is still a cyclic attack, though the placement of the cyclic group has moved from the corner of the board to the center of one side of the board. 

\subsection{\textSwitch{Adversary fine-tuning}{Adversary Fine-Tuning}}
\label{app:adv-training:adv-fine-tuning}

While KataGo's adversarial training was effective at defending against our original cyclic-adversary,
we were able to fine-tune our cyclic-adversary to defeat the adversarially trained KataGo networks (Figures~\ref{fig:evaluation:training-curve-fine-tune} and \ref{fig:adv-fine-tune-gpu-days}).

In numbers:
\begin{itemize}
    \item Prior to fine-tuning, our original cyclic-adversary cyclic-adversary$^\text{600 visits}$ wins 118 / 2000 = 5.9\% of games  against \texttt{b60-s7702m}$^\text{1 visit}$ and 0 / 400 games against \texttt{b60-s7702m}$^\text{1600 visits}$.

    \item After fine-tuning, our improved cyclic-adversary$^\text{600 visits}$ wins 188 / 400 = 47\% of games against \texttt{b60-s7702m}$^\text{4096 visits}$ and 7 / 40 = 17.5\% of games against \texttt{b60-s7702m}$^\text{100000 visits}$.

    \item Our fine-tuning also transfers (albeit in a weaker form) to the \texttt{b18-s5832m} network, winning 51 / 400 = 12.75\% of games against \texttt{b18-s5832m}$^\text{4096 visits}$.

    \item Finally, fine-tuning causes our improved cyclic-adversary to do worse against \cpfivezerofive{}, which it only wins 212 / 380 = 55.79\% of games against (compared to 973 / 1000 = 97.3\% prior to fine-tuning).
\end{itemize}

In total, we spent 1154.9 V100 GPU-days on fine-tuning our cyclic-adversary,
which underwent $1.68 \times 10^8$ steps of gradient descent.
During training, we had two crashes that stalled the progress of training, so the amount of compute we used is somewhat higher than was necessary to achieve these win rates.

We stopped our fine-tuning run due to conference deadline time constraints, but trends in the training curve suggest that we could have improved our win rate against \texttt{b60-s7702m} with more compute (Figure~\ref{fig:adv-fine-tune-gpu-days}).
However, the win-rate-compute scaling against adversarially trained networks is much worse than against undefended networks (compare Figure~\ref{fig:adv-fine-tune-gpu-days} with Figure~\ref{fig:compute:win-rate-vs-gpu-days-cyclic}).

Our cyclic-adversary fine-tuning was performed using the following curriculum:
\begin{enumerate}[leftmargin=40pt]
    \item \texttt{b60c320-s7047906048-d3140270330}$^\text{32 visits}$
    \item \texttt{b60c320-s7047906048-d3140270330}$^\text{128 visits}$
    \item \texttt{b60c320-s7047906048-d3140270330}$^\text{512 visits}$
    \item \texttt{b60c320-s7047906048-d3140270330}$^\text{1600 visits}$
    \item \texttt{b60c320-s7701878528-d3323518127}$^\text{32 visits}$
    \item \texttt{b60c320-s7701878528-d3323518127}$^\text{64 visits}$
    \item[7--10.] ...
    \item[11.] \texttt{b60c320-s7701878528-d3323518127}$^\text{2048 visits}$
\end{enumerate}
For the final network in the curriculum, \texttt{b60-s7702m} (curriculum steps \#6--11),
its last $9.73 \times 10^8$ (12.63\% of its total $77.02 \times 10^8$) steps of training included cyclic positions discovered by our cyclic-adversary.

The curriculum was advanced whenever the cyclic-adversary's win rate reached 75\%,
except we trained for about two million extra steps on \texttt{b60c320-s7047906048-d3140270330}$^\text{1600 visits}$ (\texttt{b60-s7048m}$^\text{1600 visits}$) 
since that part of the fine-tuning run was performed at a time when \texttt{b60-s7048m} was the latest 60-block KataGo network.
We found that to attack \texttt{b60-s7702m}, we achieved a stronger win rate by continuing from that existing fine-tuning run 
rather than redoing fine-tuning with a curriculum that started at \texttt{b60c320-s7701878528-d3323518127}$^\text{1 visit}$.

\begin{figure}
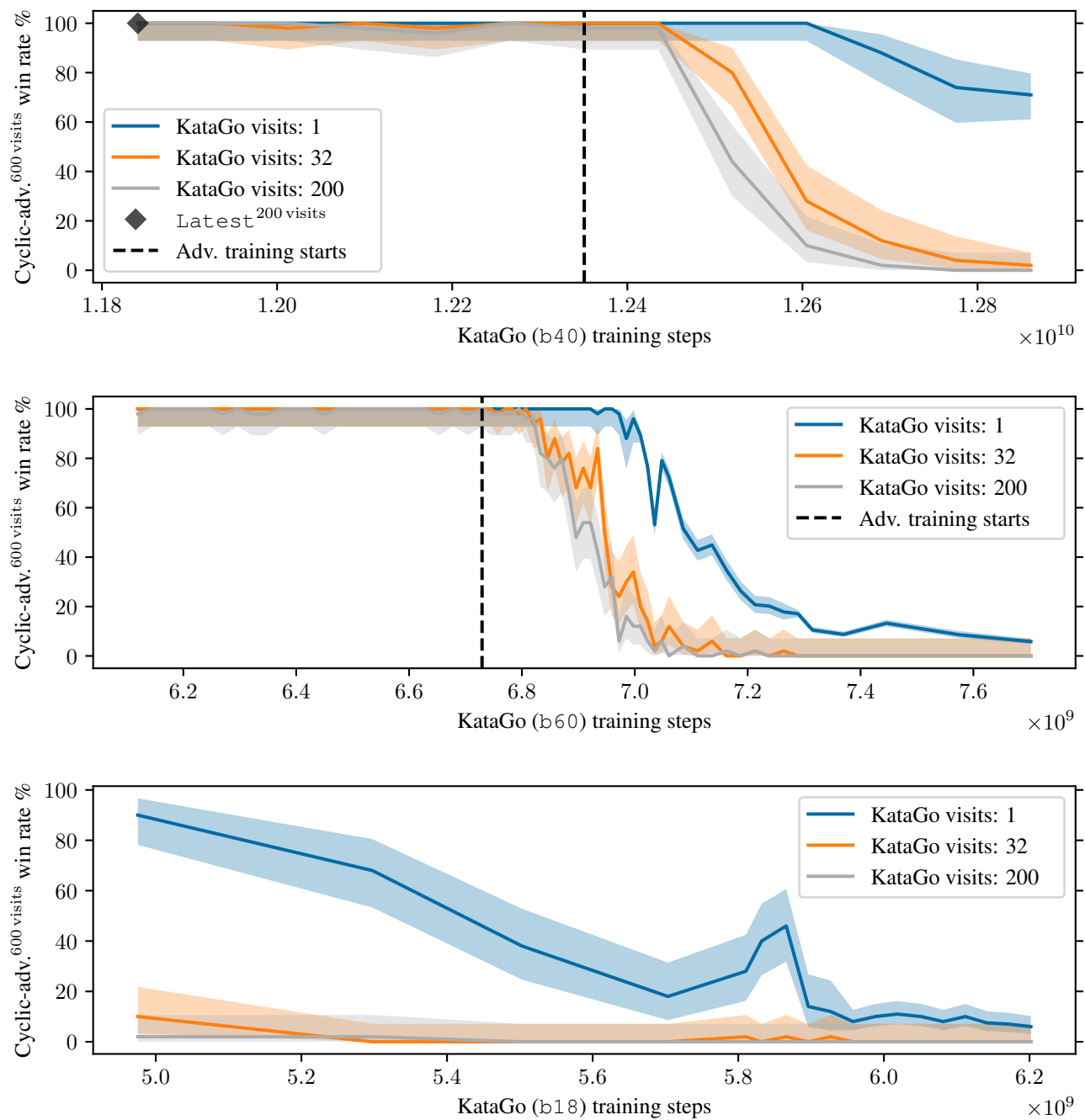

    \centering
    \input{figs/katagovisualizer/katago-adv-training-win-rate-b40c256.pgf}
    \\[5mm]
    \input{figs/katagovisualizer/katago-adv-training-win-rate-b60c320.pgf}
    \\[5mm]
    \input{figs/katagovisualizer/katago-adv-training-win-rate-b18c384nbt.pgf}
    \caption{%
        Win rate of the cyclic-adversary against
        \texttt{b40c256} (40-block, 256-channel),
        \texttt{b60c320} (60-block, 320-channel),
        and \texttt{b18c384nbt} (18-block, 384-channel) KataGo networks,
        which are convolutional+residual networks.
        The \texttt{b18c384nbt} networks have ``nested residual bottleneck blocks'',
        where
        ``each block has a linear bottleneck~[\citet{he2016deep}]
        from 384 trunk channels [to] 192 channels,
        followed by two regular $192\times192$ residual blocks,
        followed by a linear recovery from 192 channels [to] 384 trunk channels''~\citep{wu2022b18description}.
        Each subfigure displays win rate (vertical axis)
        against KataGo networks of a particular architecture
        with increasing amounts of self-play training (horizontal axis).
        The dotted black line marks when
        the adversarial cyclic positions were introduced into KataGo's distributed training run. The earliest \texttt{b18} network was released several months after the adversarial training positions were introduced.
    }
    \label{fig:adv-training:katago-training-win-rate}
\end{figure}

\begin{figure}
    \centering
    \input{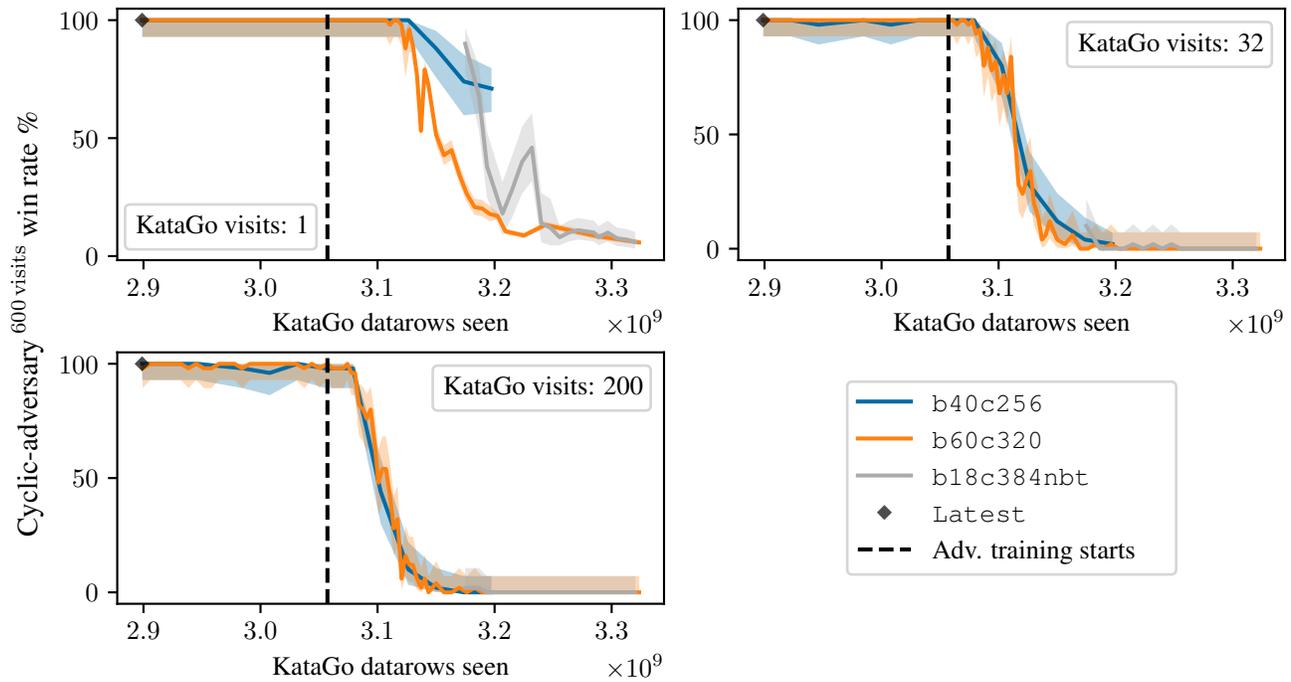}
    \caption{%
        Win rate of the cyclic-adversary against different KataGo networks.
        This is the same data as in Figure~\ref{fig:adv-training:katago-training-win-rate},
        except with the horizontal axis as number of datarows seen
        instead of number of SGD steps taken.
        In KataGo's distributed training run, all networks are trained
        on the same self-play data (which can be shared among different network types),
        so the training progress of different network architectures can be compared on the same $x$-axis scale by plotting against the number of datarows seen.
        We see that adversarial training improves performance against the cyclic-adversary
        across network architectures
        at roughly the same rate with respect to the number of datarows seen.
    }
    \label{fig:adv-training:katago-training-win-rate-datarows}
\end{figure}

\begin{figure}
    \centering
    \begin{subfigure}[b]{\textwidth}
        \centering
        \input{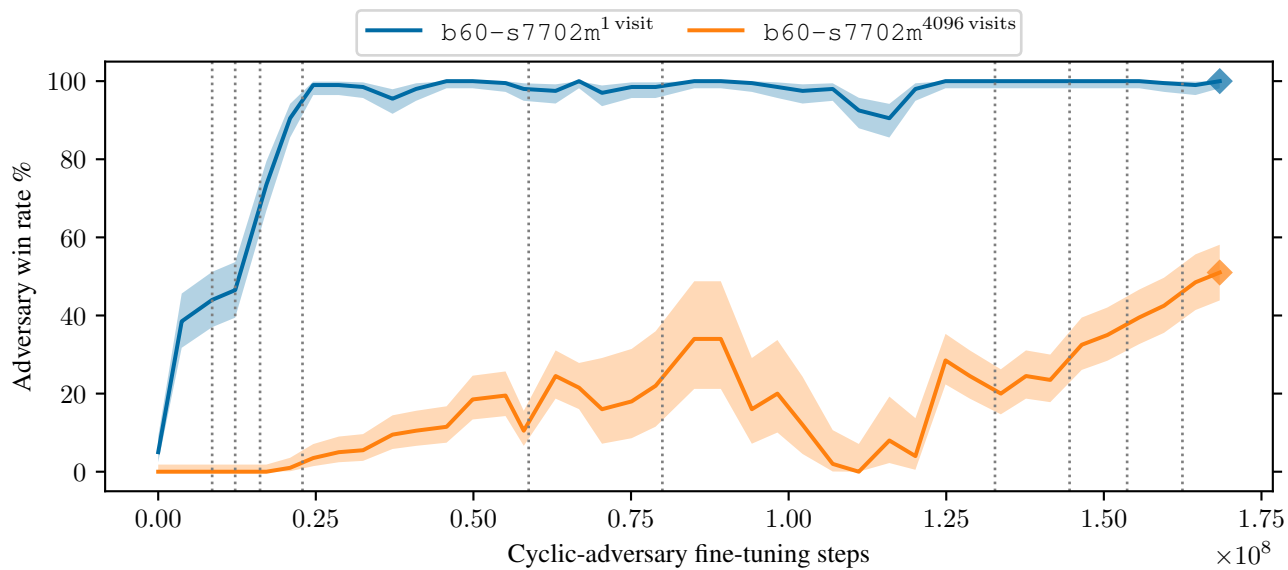}
        \caption{Win rate ($y$-axis) of the cyclic-adversary$^\text{600 visits}$ against an adversarially trained KataGo network (with varying visits) over the course of fine-tuning. The $x$-axis is the number of steps of gradient descent taken during fine-tuning.
        The shaded interval is a 95\% Clopper-Pearson interval over evaluation games (some checkpoints have more evaluations than others).}
        \label{fig:evaluation:training-curve-fine-tune:main}
    \end{subfigure}
    \\[5mm]
    \begin{subfigure}[b]{\textwidth}
        \centering
        \input{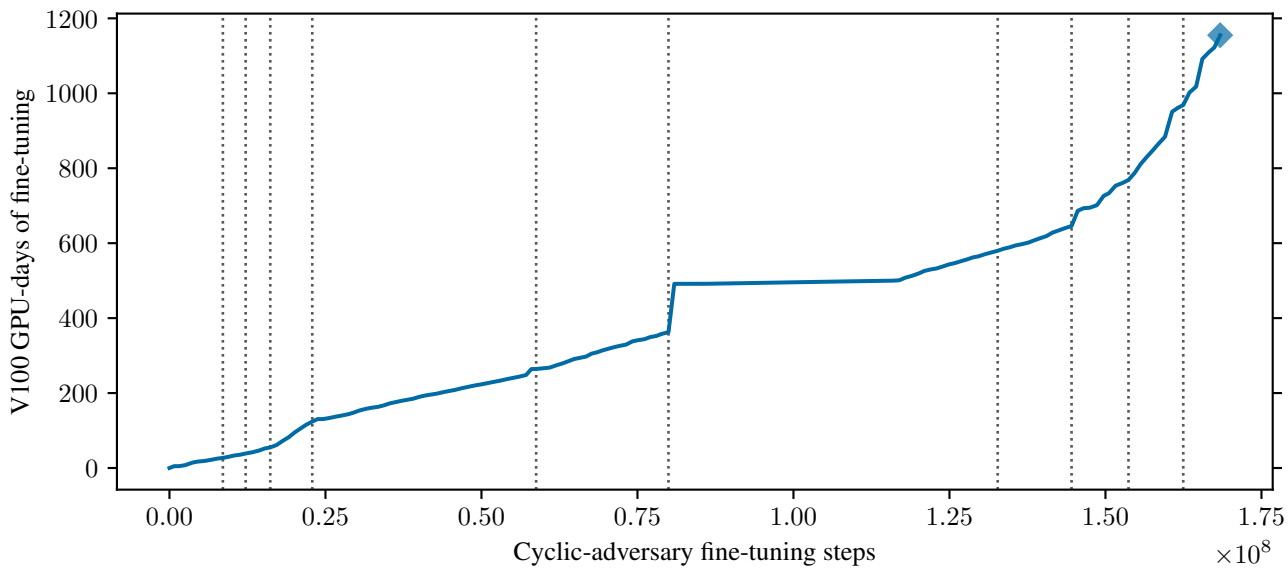}
        \caption{Fine-tuning steps ($x$-axis) vs. GPU days ($y$-axis) of the cyclic-adversary.}
        \label{fig:evaluation:training-curve-fine-tune:gpu-days}
    \end{subfigure}
    \caption{%
        Plots showing the fine-tuning progression of our cyclic-adversary against adversarially trained KataGo networks.
        Vertical dotted lines denote switches in the curriculum (as detailed in Appendix~\ref{app:adv-training:adv-fine-tuning}).
        The strongest cyclic-adversary checkpoint is marked with a diamond ($\blacklozenge$).
        Our data-shuffler crashed at $0.81 \times 10^8$ steps of fine-tuning,
        indicated by the sharp discontinuity in compute.
        This resulted in us subsequently taking many training steps with the same data, resulting in win-rate instability lasting until around $1.2 \times 10^8$ steps of fine-tuning.
        Similarly, the dip in win rate at $0.57 \times 10^8$ steps may be due to restarting the training run at that point due to a crash.
    }
    \label{fig:evaluation:training-curve-fine-tune}
\end{figure}

\begin{figure}
    \centering
    \begin{subfigure}[b]{\textwidth}
        \centering
        \input{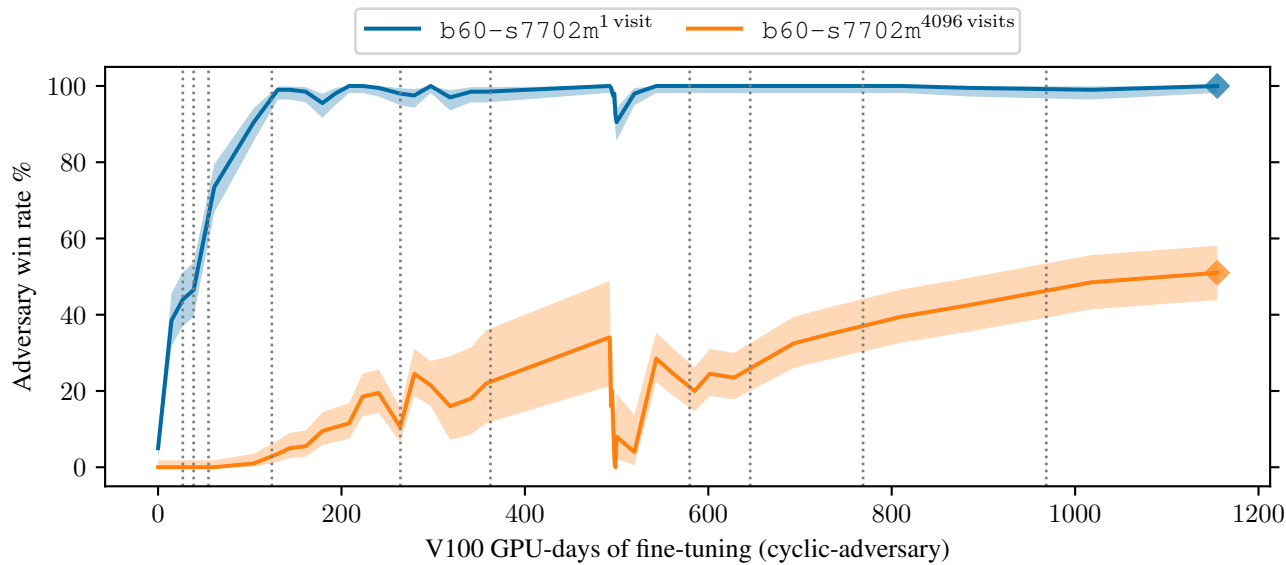}
        \caption{Win rate ($y$-axis) of the cyclic-adversary$^\text{600 visits}$ against an adversarially trained KataGo network (with varying visits) over the course of fine-tuning. The $x$-axis is the amount of compute spent fine-tuning.
        The shaded interval is a 95\% Clopper-Pearson interval over evaluation games (some checkpoints have more evaluations than others).}
        \label{fig:adv-fine-tune-gpu-days:eval}
    \end{subfigure}
    \\[5mm]
    \begin{subfigure}[b]{\textwidth}
        \centering
        \input{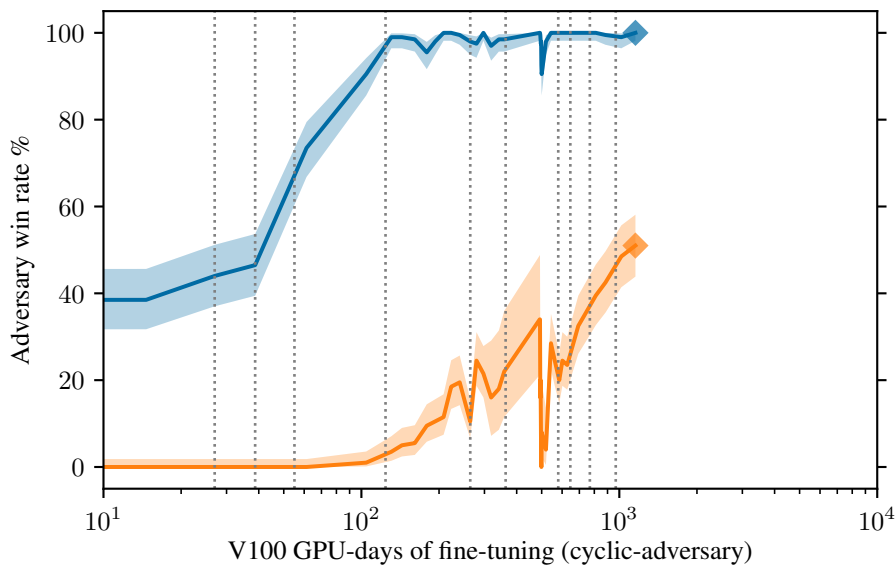}
        \caption{The same plot as above, but with a log-scaled $x$-axis.}
        \label{fig:adv-fine-tune-gpu-days:eval-log}
    \end{subfigure}
    \caption{%
        These plots are the same as as Figure~\ref{fig:evaluation:training-curve-fine-tune:main},
        but in terms of compute.
        The effects of the data-shuffler crash
        can be seen in 400-500 V100 GPU days region,
        where the win rate sharply drops and then recovers.
        On the right end of the plots,
        we see that the win-rate of the cyclic-adversary against \texttt{b60-s7702m}$^\text{4096 visits}$ steadily increases at roughly a log-linear rate.
        Rough eyeballing suggests that our cyclic-adversary may be able to reach a 90\%+ win rate against \texttt{b60-s7702m}$^\text{4096 visits}$ in about 10,000 V100 GPU days.
        This win-rate-compute scaling is much worse than the scaling of the original cyclic adversary (compare with Figure~\ref{fig:compute:win-rate-vs-gpu-days-cyclic}).
    }
    \label{fig:adv-fine-tune-gpu-days}
\end{figure}

\clearpage
\section{\textSwitch{Adversarial board state}{Adversarial Board State}}
\label{app:adversarial-state}

This paper focuses on training an \emph{agent} that can exploit Go-playing AI systems.
A related problem is to find an adversarial \emph{board state} which could be easily won by a human, but which Go-playing AI systems will lose from.
In many ways this is a simpler problem, as an adversarial board state need not be a state that the victim agent would allow us to reach in normal play.
Nonetheless, adversarial board states can be a useful tool to probe the blind spots that Go AI systems may have.

\begin{figure}[h]
    \centering
    \includesvg[width=0.48\textwidth]{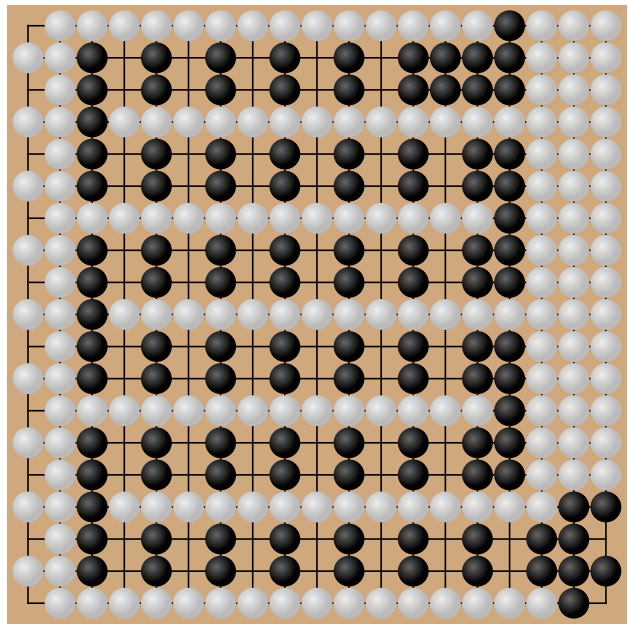}
    \caption{A hand-crafted adversarial example for KataGo and other Go-playing AI systems. It is black's turn to move. Black can guarantee a win by connecting its currently disconnected columns together and then capturing the large white group on the right.
    However, KataGo playing against itself from this position loses 40\% of the time as black.
    }
    \label{fig:adversarial-state}
\end{figure}

In Figure~\ref{fig:adversarial-state} we present a manually constructed adversarial board state.
Although quite unlike what would occur in a real game, it represents an interesting if trivial (for a human) problem.
The black player can always win by executing a simple strategy.
If white plays in between two of black's disconnected groups, then black should immediately respond by connecting those groups together.
Otherwise, the black player can connect any two of its other disconnected groups together.
Whatever the white player does, this strategy ensures that blacks' groups will eventually all be connected together.
At this point, black has surrounded the large white group on the right and can capture it, gaining substantial territory and winning.

Although this problem is simple for human players to solve, it proves quite challenging for otherwise sophisticated Go AI systems such as KataGo.
In fact, KataGo playing against a copy of itself \emph{loses} as black 40\% of the time.
We conjecture this is because black's winning strategy, although simple, must be executed flawlessly and over a long horizon.
Black will lose if at any point it fails to respond to white's challenge, allowing white to fill in both empty spaces between black's groups.
This problem is analogous to the classical cliff walking reinforcement learning task~\citep[Example~6.6]{sutton2018}.

\clearpage
\section{\textSwitch{Known failures of Go-playing agents}{Known Failures of Go-playing Agents}}
\label{app:bot_weaknesses}

The cyclic vulnerability our adversary finds is unique in the confluence of 3 key factors. First, it affects top Go-playing agents, even when they have a very large amount of search. Second, it consistently produces a game-winning advantage. Third, this consistency does not require exact sequences or board positions. Along with other factors like how it is non-trivial to defend against (see Appendix~\ref{app:adv-training}), that makes this vulnerability particularly significant. 

Nonetheless, there are other known vulnerabilities of Go AIs, some of which exhibit a subset of those significant characteristics and can be noteworthy of their own right. The following overview draws largely on discussion with David Wu, creator of KataGo.

\paragraph{Ladders}

\begin{figure}[h]
    \centering
        \centering
        \includesvg[inkscapelatex=false, width=0.48\textwidth]{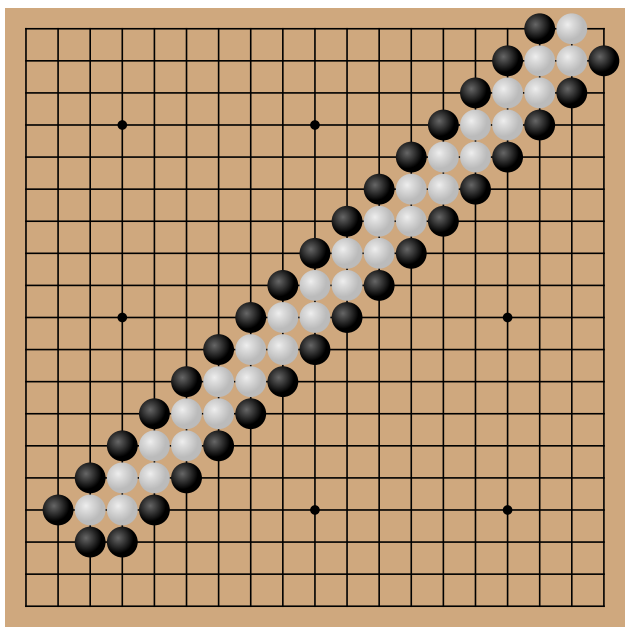}
	\caption{Illustration of a ladder. White ran across the board, but has hit the top edge and has nowhere left to run. Black can capture on the next move by playing in the top right corner.}
    \label{fig:ladder-example}
\end{figure}

A "ladder" is often the first tactic a beginner learns. An example is shown in Figure~\ref{fig:ladder-example}. In this pattern, the defending side only has a single move to attempt to escape, while the attacking side only has a single move to continue threatening the defender. After each move in the pattern, the same situation recurs, shifted one space over diagonally. The chain continues until either the defender runs into the edge of the board or more enemy stones, at which time there is no more room to escape, or conversely into the defender's allied stones, in which case the defender escapes and the attacker is usually left disastrously overextended. Consequently, it is a virtually unbranching pattern, but one that takes many moves (often dozens), depends on the position all the way on the other side of the board, and can decide the result of the game.

Bots struggle to understand when escape and capture is possible, especially fairly early in the game. This issue occurs across many different models. It is especially prevalent early in training and with less search, but even with thousands of playouts per move it still occurs.

This issue has been solved in KataGo by adding a separate, hardcoded ladder module as an input feature. Such an approach, however, would not work for flaws one is unaware of, or where hardcoded solutions are prohibitively difficult.

\paragraph{Liberty Counts}

Even without a long variation or consistent pattern, bots may occasionally fail to see that something can be captured on their move or their opponent's next move. Known examples of this occurred with very large groups in slightly unusual situations, but nonetheless where an intermediate human would easily make the correct judgment.

This is again mitigated in KataGo through a hardcoded auxiliary algorithm that provides input features (liberty counts) to the main network.

\paragraph{Complicated Openings}

There are some extremely complicated opening variations, especially variations of Mi Yuting's Flying Dagger joseki, which have crucial, unusual moves required to avoid a disadvantage. Once again, KataGo solved this with a manual intervention. Here it was through directly adding a large collection of variations to the training. Other bots still play certain variations poorly.

\paragraph{Cyclic Topology}

This is a position with a loop, such as the marked group in Figure~\ref{fig:cp505vis1-board:hardened}, and the weakness our cyclic-adversary exploits. Such situations are possible but very uncommon in normal play. David Wu's hypothesis is that information propagates through the neural network in a way analagous to going around the cycle, but it cannot tell when it has reached a point it has "seen" before. This leads to it counting each liberty multiple times and judging such groups to be very safe regardless of the actual situation.

We, the authors of this paper, were not aware of this weakness of Go bots until after we had trained the cyclic-adversary. It was also not known that this weakness could be consistently exploited.

\paragraph{Mirror Go}

This is where one player copies the other player's moves, mirroring them across the board diagonally. This is typically not part of training nor other aspects of agents' construction. However, even without specific counter strategies, there is a long time over the course of the game to stumble into a position where a generically good move also breaks the mirror. So this strategy is not a consistent weakness, but can occasionally win games if no such good mirror-breaking move happens to come up.

\paragraph{Other}

Finally, there are also other mistakes bots make that are more complex and more difficult to categorize. Even though the best bots are superhuman, they are certainly still a ways away from perfect play, and it is not uncommon for them to make mistakes. In some positions these mistakes can be substantial, but fixing them may be not so much about improving robustness as it is about building an overall stronger agent.

\paragraph{Summary}

There are a number of situations that are known to be challenging for computer Go players. Some can be countered through targeted modifications and additions to the model architecture or training, however, as we see with Cyclic Topology, it is difficult to design and implement solutions one-by-one to fix every possibility. Further, the weaknesses may be unknown or not clearly understood -- for instance, Cyclic Topology is normally rare, but through our work we now know it can be produced consistently. Thus, it is critical to develop algorithmic approaches for detecting weaknesses, and eventually for fixing them.

\clearpage
\ifisarxiv
\else

    \section{Reproducibility Statement}
    \label{sec:repro}
We take the following steps to promote and ensure reproducibility:
\begin{itemize}
    \item Our code is available at \githuburl{}. The code is containerized and includes instructions for running it.
    \item We make many game records available through our \href{\demosite/}{website}. We will make more game records available to researchers upon request and have already provided game records to David Wu, the creator and primary developer of KataGo, for use in KataGo's training process.
    \item We set up a bot running the most recent checkpoint of our cyclic-adversary on the KGS Go server, under the username \href{https://www.gokgs.com/graphPage.jsp?user=Adversary0}{\texttt{Adversary0}}. This bot was available for the public to play for a period of a month. See Appendix~\ref{app:experiments:human-vs-adversary} for more details.
\end{itemize}

A number of our key results have already been reproduced:
\begin{itemize}
    \item
    The vulnerability to the passing attack has been independently confirmed by David Wu.
    
    \item
    The vulnerability to the cyclic attack has been independently confirmed by David Wu, as well as many others in the computer Go community.
    
    \item
    The cyclic vulnerability and the adversary's ability to use it has been replicated through normal bot play against the KGS bot we made available, as has the result that novice human play beats the adversary.

    \item
    Human ability to use the cyclic attack has been independently reproduced against
    \href{https://online-go.com/game/53421988}{KataGo}, %
    as well as in transfer settings against
    \href{https://online-go.com/game/51321265}{ELF OpenGo},
    \href{https://h5.foxwq.com/txwqshare/index.html?chessid=1676910620010001365&boardsize=19}{FineArt},
    \href{https://online-go.com/game/51356405}{Leela Zero},
    and \href{https://online-go.com/game/51375020}{Sai}.
\end{itemize}

    \section{Acknowledgements}
    \label{app:ack}
    Thanks to David Wu and the Computer Go Community Discord for sharing their knowledge of computer Go with us and for their helpful advice on how to work with KataGo,
to Adri\`a Garriga-Alonso for feedback and assistance setting up activation analysis and infrastructure,
to Lawrence Chan, Euan McLean, and Niki Howe for their feedback on earlier drafts of the paper,
to ChengCheng Tan and Alyse Spiehler for assistance preparing illustrations,
to David Fontaine for help with debugging KataGo deadlocks, %
to Matthew Harwit for help with Chinese communication and feedback especially on Go analysis,
to Daniel Filan for Go game analysis and feedback on project direction,
and to Nir Shavit for his support of and high level feedback on the project. %

    Tony Wang was supported by funding from the Eric and Wendy Schmidt Center at the Broad Institute of MIT and Harvard. %

    \section{Author Contributions}
    \iffinalcopy
Tony Wang invented and implemented the \ourmctssampleabbrev{} algorithm,
made several other code contributions, and ran and analyzed many of the experiments.
Adam Gleave managed the project, wrote the core of the paper, suggested the curriculum approach, helped manage the cluster experiments were run on, and implemented some minor features.
Tom Tseng implemented and ran transfer experiments, trained and ran experiments with the pass-hardening defense enabled, and ran many of the evaluations.
Kellin Pelrine (our resident Go expert) provided analysis of our adversary's strategy, search vs. robustness, activation analysis, and manually reproduced the cyclic-attack against different Go AIs.
Nora Belrose implemented and ran the experiments for baseline adversarial policies, and our pass-hardening defense.
Joseph Miller developed the website showcasing the games, and an experimental dashboard for internal use.
Michael Dennis developed an adversarial board state for KataGo that inspired us to pursue this project, and contributed a variety of high-level ideas and guidance such as adaptations to MCTS.
Yawen Duan ran some of the initial experiments and investigated the adversarial board state.
Viktor Pogrebniak implemented the curriculum functionality and improved the KataGo configuration system.
Sergey Levine and Stuart Russell provided guidance and general feedback.
\else
Removed for double blind submission.
\fi

\fi

\iffinalcopy
    \ifisarxiv
        \section{Changelog} \label{app:changelog}
\href{https://arxiv.org/abs/2211.00241v3}{\texttt{arxiv.org/abs/2211.00241v3}} $\to$ \href{https://arxiv.org/abs/2211.00241v4}{\texttt{arxiv.org/abs/2211.00241v4}} changes:
\begin{itemize}
    \item Introduction rewritten.

    \item Added a new diagram (Figure~\ref{fig:amcts-diagram}) explaining \ourmctsabbrev{}.

    \item Added Section~\ref{sec:evaluation:defense} and Appendix~\ref{app:adv-training} detailing the KataGo team's official efforts to defend against our attack via adversarial training, and our response to those efforts.

    \item Added an exploration of how the internal activations of KataGo behave on cyclic and non-cyclic groups (Section~\ref{sec:evaluation:understanding-the-attack}, Figure~\ref{fig:activations-L25v26}, Appendix~\ref{app:activations}).

    \item Added a reproducibility statement (Section~\ref{sec:repro}).

    \item Improved figures and text across the whole paper.

    \item \demositecleanhref{} updated to reflect the \texttt{v4} paper changes.

    \item Paper format changed to (roughly) the ICML 2023 conference proceedings format,
    reflecting our acceptance to the conference.
    This arXiv version contains minor improvements on top of the conference proceedings version,
    in the form of formatting and layout changes,
    one additional figure (Figure~\ref{fig:defense:adv-finetuning}),
    and small textual edits.
\end{itemize}

\href{https://arxiv.org/abs/2211.00241v2}{\texttt{arxiv.org/abs/2211.00241v2}} $\to$ \href{https://arxiv.org/abs/2211.00241v3}{\texttt{arxiv.org/abs/2211.00241v3}} changes:
\begin{itemize}
    \item We update our evaluations to use a stronger cyclic-adversary that was trained against victims with higher visit counts. For example, Figure~\ref{fig:cp505vis1-board} now features a game between this stronger cyclic-adversary and \hardened{\cpfivezerofive{}} with $10^7$ visits.

    \item We evaluate the transferability of our attack in Section~\ref{sec:evaluation:transfer} and Appendix~\ref{app:transfer}. An author of the paper (our resident Go-expert) was able to learn how to manually perform the cyclic-attack (i.e. without computer assistance) and win convincingly against a range of superhuman Go AIs.

    \item We describe a vulnerability introduced by the pass-alive defense (Appendix~\ref{app:pass-alive-defense-vulnerability}).

    \item We give more precise estimates of compute required to train our adversaries and KataGo (Appendix~\ref{app:compute-estimates}).

    \item We discuss the role of search in robustness (Appendix~\ref{app:role-of-search}).

    \item We analyze the evolution of our cyclic-adversary's strategy over the course of training (Appendix~\ref{app:training-games-analysis}).

    \item Figure and table numbering changed to be section-relative.

    \item We make small improvements to figures and text across the whole paper.

    \item We update \demositecleanhref{} to reflect the \texttt{v3} paper changes.
\end{itemize}

\href{https://arxiv.org/abs/2211.00241v1}{\texttt{arxiv.org/abs/2211.00241v1}} $\to$ \href{https://arxiv.org/abs/2211.00241v2}{\texttt{arxiv.org/abs/2211.00241v2}} changes:
\begin{itemize}
    \item We train, evaluate, and analyze a new \textit{cyclic-adversary} (Figure~\ref{fig:cp505vis1-board:hardened}) which behaves in a qualitatively different manner from our \texttt{v1} adversary (now called the \textit{pass-adversary}). Our cyclic-adversary can beat KataGo playing with up to $10^7$ visits per move of search.

    \item We add a detailed description of the Go rules used in our evaluations (Appendix~\ref{app:rules}).

    \item We add an estimate of the strength of the AlphaZero agent that was successfully adversarially attacked by \citet{timbers2022} using methods similar to our own (Appendix~\ref{app:experiments:strength:alphazero}).

    \item We redo the evaluations of our \texttt{v1} pass-adversary with configurations more similar to those used in actual match-play (Appendix~\ref{app:unhardened-results}).

    \item We add a summary of known failures of Go AIs (Appendix~\ref{app:bot_weaknesses}).

    \item We make small improvements to figures and text across the whole paper.

    \item We update our paper website (\demositecleanhref) to reflect the \texttt{v2} paper changes.
\end{itemize}

    \fi
\fi

\end{document}